\documentclass{article}
\usepackage{microtype}
\usepackage{graphicx}
\usepackage{booktabs}
\usepackage{hyperref}
\usepackage[preprint]{icml2026}
\usepackage{amsmath,amssymb,amsthm}
\usepackage{needspace}
\usepackage{balance}
\usepackage{enumitem}
\usepackage{xurl}

\newtheorem{theorem}{Theorem}
\newtheorem{proposition}[theorem]{Proposition}
\newtheorem{corollary}[theorem]{Corollary}

\theoremstyle{definition}
\newtheorem{definition}{Definition}
\newcommand{\ess}{\mathrm{ESS}}
\newcommand{\E}{\mathbb{E}}
\newcommand{\R}{\mathbb{R}}

\icmltitlerunning{Is a Description Worth a Thousand Demonstrations?}

\begin{document}
\raggedbottom

\twocolumn[
\icmltitle{Is a Description Worth a Thousand Demonstrations?\texorpdfstring{\\}{ }Specificity and Reliability of Task Descriptions in In-Context Learning}
\begin{icmlauthorlist}
\icmlauthor{Bruce Nguyen}{dec,ua}
\icmlauthor{Bei Jiang}{ua}
\icmlauthor{Linglong Kong}{ua}
\end{icmlauthorlist}
\icmlaffiliation{dec}{Decaude}
\icmlaffiliation{ua}{Department of Mathematical and Statistical Sciences, University of Alberta, Edmonton, Canada}
\icmlcorrespondingauthor{Bruce Nguyen}{bruce.nguyen@decaude.com, trungqua@ualberta.ca}
\icmlkeywords{in-context learning, task descriptions, effective sample size, Bayesian inference}
\vskip 0.3in
]
\printAffiliationsAndNotice{}

\begin{abstract}
A prompt for in-context learning (ICL) can describe a task in addition to, or instead of, demonstrating it through examples. Yet how task descriptions work remains poorly understood theoretically. Through the Bayesian view of ICL, we model descriptions as informative priors characterized by their \emph{specificity} and \emph{reliability}. We adapt the effective sample size (ESS) from Bayesian statistics, quantifying a description's predictive value as an equivalent number of demonstrations. For in-context linear regression, we show that a reliable description's ESS grows with its specificity and with the task's dimension. A description that may be irrelevant, however, gains little from its specificity unless demonstrations are given. In practice, do in-context learners use descriptions as a Bayes-optimal predictor would? Our experiments show that meta-trained Transformers come close to the Bayes-optimal ESS when tested on their training distribution. Pretrained language models, however, make too little use of highly specific descriptions when answering without reasoning; most of them closely match the Bayes-optimal predictor only when reasoning is enabled. These results extend the Bayesian analysis of ICL from demonstrations to descriptions.
\end{abstract}

\section{Introduction}
\begin{figure*}[t]
\centering
\includegraphics[width=\textwidth]{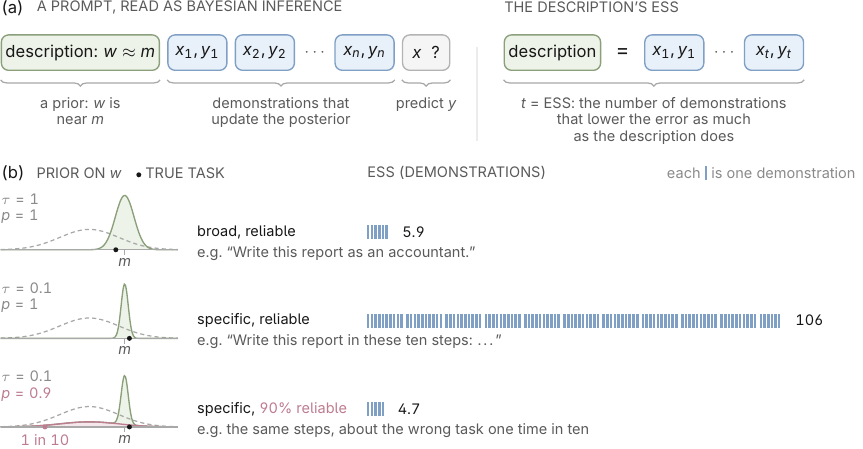}
\caption{\textbf{Measuring a task description in demonstrations.} (a) A description's ESS is the number of demonstrations that lower the error as much as it does. (b) The prior on the task $w$ under three descriptions, and each description's Bayes-optimal ESS at $n=0$, with $d=5$ and baseline prior variance $\tau_0^2=10$ (dashed); $\tau$ is the standard deviation of the task around the description's prior mean $m$. The curves are sketches and the quoted instructions are illustrations; dots mark the true task, and the pink curve is the prior when the description is about the wrong task.}
\label{fig:overview}
\end{figure*}

Large language models (LLMs) are capable of in-context learning (ICL): given a prompt that specifies a task through a natural-language description, demonstrations, or both, they can predict the output for a new query input \citep{brown2020}. In practice, prompts consist mostly of system prompts, user instructions, agent skill files, and earlier conversation turns, which typically describe the task rather than demonstrate it \citep{mao2025,patsakis2026}. The balance shifts further toward description as LLMs carry out increasingly long-horizon tasks \citep{kwa2025}, for which demonstrations are costly to provide.

Theories of ICL, however, have mostly focused on prompts consisting of demonstrations alone \citep{xie2022,garg2022,akyurek2023}. The few that include descriptions study only particular types, such as a task descriptor giving the mean of the inputs \citep{huangge2025}, a hypothesis class listed in the prompt \citep{lin2025}, or datasets from related tasks that inform the prior \citep{zhu2026}; none offers a systematic treatment of descriptions. We therefore ask: how can we characterize natural-language descriptions in general, so as to understand and analyze them formally? And how well do models, whether meta-trained Transformers or pretrained language models, use descriptions in practice?

We study these questions through the Bayesian view of ICL, where a model infers a latent task, a function mapping inputs to outputs, using the given demonstrations \citep{xie2022}. A description carries information about the task without being a demonstration of it, so it can be modeled as a prior over tasks. Descriptions differ in how much they say about a task, their \emph{specificity}: ``write this report as an accountant'' leaves most of the task open, whereas a step-by-step workflow leaves little. They also differ in whether they are about the right task: an instruction can be outdated or meant for a different task. Since any description may be about the wrong task, we model this property by the probability that it is about the right one, its \emph{reliability}.

Neither property can be measured directly in a natural-language instruction, so as a first step we study them in $d$-dimensional Gaussian linear regression, a standard controlled setting for ICL \citep{garg2022,panwar2024}, where both are known exactly. Here, a task is a latent regression vector, and a description gives a prior mean, the regression vector it points to. If the description is about the right task, which we call \emph{relevant}, the task is Gaussian around that mean with standard deviation $\tau$, and the description's specificity is the precision $c=\tau^{-2}$: the more it says, the higher $c$. Reliability is then the probability $p$ that the description is relevant. An \emph{irrelevant} description's mean is independent of the task (Figure~\ref{fig:overview}).

To compare descriptions with demonstrations, we adapt the effective sample size (ESS) from Bayesian statistics, which expresses a prior's information or impact as an equivalent number of observations \citep{wiesenfarth2020,reimherr2021,selby2025}. A description's \emph{predictive ESS} is the number of additional demonstrations that a predictor would need, in place of the description, to reach the same expected error (Section~\ref{sec:setting}). It requires only the predictor's errors, so it applies to any predictor: for the Bayes-optimal predictor, it gives a description's value under the statistical model; for a trained network or language model, it gives the value that the model realizes. To see how well models use descriptions, we test how close each model comes to the ESS of a Bayesian predictor given the same information.

Our contributions are as follows:
\begin{itemize}[leftmargin=*,itemsep=2pt,topsep=3pt]
\item \textbf{We show that specificity and dimension determine the value of reliable descriptions.} When little is known about the task without a description, one with precision $c$ is worth about $c+d+1$ demonstrations before any are seen. Demonstrations lower this value, to about $c$ after many; a description that constrains only part of the task can instead gain value from them (Section~\ref{sec:specificity}).
\item \textbf{We show that unreliability caps zero-shot ESS.} Any fixed positive probability of irrelevance caps the ESS without demonstrations, the zero-shot ESS, at a finite value, however specific the description. Demonstrations can then provide evidence of whether the description is relevant and raise its ESS above this ceiling (Section~\ref{sec:reliability}, Figure~\ref{fig:ess}b).
\item \textbf{We find that meta-trained Transformers approach the Bayes-optimal ESS when tested on their training distribution.} Without demonstrations, they weight a description by how often descriptions were relevant in training, whatever the test reliability (Section~\ref{sec:networks}).
\item \textbf{We find that language models use the precision and the reliability stated in the prompt mainly with reasoning enabled.} Without reasoning, every model tested makes far less use of a highly specific description than a Bayesian predictor given the same stated prior; with reasoning, most closely match that predictor's error. When one description in ten is in fact irrelevant, the three models tested with reasoning also rely less on a description that the demonstrations contradict (Section~\ref{sec:llm}).\end{itemize}

\section{Problem Setup}\label{sec:setting}

A description carries information about the task without being a demonstration of it, so we model it as a prior over tasks. Let $Q_0$ be the \emph{baseline prior}, the distribution of tasks when no description is given. A description comes with a prior of its own, $Q_{\mathrm{desc}}$. If the description is about the right task, which we call \emph{relevant}, the task is drawn from $Q_{\mathrm{desc}}$. Otherwise the description is \emph{irrelevant}, and the task is drawn from $Q_0$ whatever the description says. Given a description, the task therefore has the prior
\begin{equation}\label{eq:mixture}
p\,Q_{\mathrm{desc}}+(1-p)\,Q_0,
\end{equation}
where $p$, the description's \emph{reliability}, is the probability that it is relevant. We call the description \emph{reliable} when $p=1$ and \emph{unreliable} when $p<1$. This two-component form is the robust mixture prior that Bayesian clinical trials use to borrow external information that may not apply \citep{schmidli2014}.

A description's \emph{specificity} is how little uncertainty it leaves about the task when it is relevant. We measure this uncertainty by the expected loss of the best prediction of a query output made from the description alone \citep{degroot1962,grunwald2004}. We score predictions by squared error, as is standard for in-context regression \citep{garg2022}; under this loss, the uncertainty is the variance of $Q_{\mathrm{desc}}$, summed over the coordinates of the task (Appendix~\ref{app:specificity}). We therefore measure specificity by the precision of $Q_{\mathrm{desc}}$, the inverse of its variance in each coordinate: the higher the precision, the more specific the description. We study the prior~\eqref{eq:mixture} in Gaussian linear regression.

\paragraph{Tasks and demonstrations.}
A task is a vector $w\in\R^d$, and the baseline prior is $Q_0=\mathcal{N}(0,\tau_0^2I_d)$, where $\tau_0>0$. A prompt contains $n$ demonstrations $D_n=\{(x_i,y_i)\}_{i=1}^n$ and a query input $x$, with
\[
x_i,x\overset{\mathrm{iid}}{\sim}\mathcal{N}(0,I_d),\qquad
y_i=w^\top x_i+\epsilon_i,\quad
\epsilon_i\overset{\mathrm{iid}}{\sim}\mathcal{N}(0,1).
\]
Inputs and observation noise are independent of the task, of its description, and of one another. Numerical illustrations in Sections~\ref{sec:specificity}--\ref{sec:reliability} use $d=5$, the task dimension of the meta-trained Transformers in Section~\ref{sec:networks}, and $\tau_0^2=10$; other dimensions and baseline variances give the same pattern (Table~\ref{tab:grid} in Appendix~\ref{app:reliability}).

\paragraph{Descriptions.}
A description supplies a prior mean $m\in\R^d$, and its prior is $Q_{\mathrm{desc}}=\mathcal{N}(m,\tau^2I_d)$ with $0<\tau\leq\tau_0$: a relevant description places the task within about $\tau$ of $m$ in each direction. The variance of this prior is $\tau^2$ in each coordinate, so the description's specificity is the precision $c=\tau^{-2}$. We write $c_0=\tau_0^{-2}$ for the baseline prior's precision. A description changes what is known about the task but not how tasks are distributed, so the task must follow $Q_0$ once we average over descriptions. Under a relevant description, $w$ is $m$ plus independent noise of variance $\tau^2$, so $m$ must have variance $\tau_0^2-\tau^2$, the part of the task's variance that the description accounts for. With a latent relevance indicator $Z\sim\mathrm{Bernoulli}(p)$ independent of $m$, the joint model is
\begin{align*}
m&\sim\mathcal{N}(0,(\tau_0^2-\tau^2)I_d),\\
w\mid m,Z=1&\sim\mathcal{N}(m,\tau^2I_d),\\
w\mid m,Z=0&\sim\mathcal{N}(0,\tau_0^2I_d).
\end{align*}
The description is relevant when $Z=1$ and irrelevant when $Z=0$, which gives the prior in~\eqref{eq:mixture}. A description of the wrong task carries no information about this one, so when $Z=0$ its mean is independent of the task. A description of the right task that is only somewhat inaccurate is modeled instead as a relevant description with a larger $\tau$.

\paragraph{Prediction error.}
For a prediction $\hat y$, we measure expected squared error against the noiseless query output:
\begin{equation}\label{eq:regret}
R=\E\big[(\hat y-w^\top x)^2\big].
\end{equation}
We call this quantity \emph{prediction error}, or simply error; it excludes irreducible query noise. The expectation is over tasks, descriptions, demonstrations, and queries. For a fixed predictor, $R^{\mathrm{desc}}(n)$ and $R^{\mathrm{base}}(n)$ denote the error with and without a description. As a function of the demonstration count, $R^{\mathrm{base}}$ is the predictor's \emph{baseline curve}, its learning curve without the description.

\begin{definition}[Predictive effective sample size]\label{def:ess}
Given $n$ demonstrations, let $n^*$ be the smallest $t\geq0$ satisfying $R^{\mathrm{base}}(t)=R^{\mathrm{desc}}(n)$. When such a $t$ exists, the description's ESS is
\[
\ess_n=n^*-n.
\]
We interpolate the baseline curve between integer counts, so $n^*$ and the ESS need not be integers. When no such $t$ exists, as when the error with the description lies below the predictor's whole baseline curve, the ESS is undefined.
\end{definition}

For example, $\ess_n=10$ means that $n$ demonstrations plus the description match the error of $n+10$ demonstrations alone. We call $\ess_0$ the \emph{zero-shot ESS}.
Definition~\ref{def:ess} follows the prior sample size of \citet{reimherr2021}: the number of additional observations that a baseline prior needs to match the mean squared error of an estimate under the prior of interest. We match prediction error, which can be measured for any predictor, including a network or a language model, whose prior knowledge is only implicit in its parameters. Because ESS is measured on each predictor's own baseline curve, a large ESS need not mean a small error, so we also report the language models' absolute error (Table~\ref{tab:error}, Appendix~\ref{app:reasoning-runs}).

\section{The Value of a Reliable Description}\label{sec:specificity}

In this section and the next, error and ESS are those of the Bayes-optimal predictor. We first take a reliable description, $p=1$. Its ESS need not equal its specificity $c$: without demonstrations, it also depends on the task's dimension. We state this zero-shot result for a vague baseline prior, under which little is known about the task without the description: the limit $\tau_0\to\infty$ with $\tau$ fixed.

\begin{proposition}[Reliable descriptions]\label{prop:reliable}
Let $p=1$.
\begin{enumerate}[label=(\roman*),nosep,leftmargin=*]
\item With $\tau$ fixed and $c=\tau^{-2}$ a positive integer, $\ess_0\to c+d+1$ as $\tau_0\to\infty$.
\item With $\tau$ and $\tau_0$ fixed, $\ess_n\to c-c_0$ as $n\to\infty$.
\end{enumerate}
\end{proposition}

\begin{figure}[t]
\centering
\includegraphics[width=\linewidth]{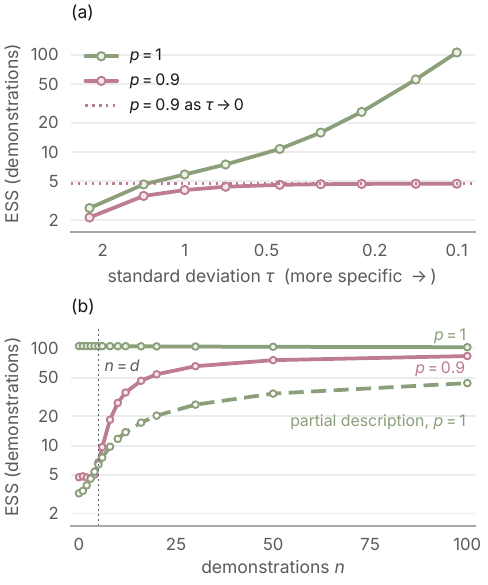}
\caption{\textbf{Without demonstrations, reliability limits the benefit of specificity; demonstrations recover much of it.} Bayes-optimal ESS with $d=5$ and $\tau_0^2=10$. (a) $\ess_0$ versus the description's standard deviation $\tau$, for $p=1$ and $p=0.9$; a smaller $\tau$ means a more specific description. The dotted line is the finite ceiling as $\tau\to0$ at $p=0.9$, about $4.7$ here; it tends to $p^2d=4.05$ as $\tau_0\to\infty$. (b) $\ess_n$ versus the number of demonstrations $n$ for three descriptions with $\tau=0.1$: reliable, unreliable ($p=0.9$), and a reliable partial description, which constrains $k=3$ of the $5$ coordinates of the task. Values are Monte Carlo estimates (Appendix~\ref{app:reliability}).}
\label{fig:ess}
\end{figure}

Part (i) follows by comparing the description's error, $d\tau^2=d/c$, with the baseline error after $t$ demonstrations, which in the limit $\tau_0\to\infty$ is $d/(t-d-1)$ for integer $t>d+1$ \citep{muirhead1982}: the two match at $t=c+d+1$ (Appendix~\ref{app:specificity}). If the demonstration inputs gave every direction of the task the same information, the baseline error would be $d/t$ and the description would be worth $c$ demonstrations. Random inputs constrain some directions less than that, so the error is $d/(t-d-1)$, and matching the description takes $d+1$ more demonstrations. At $d=5$ and $\tau_0^2=10$, a reliable description with $\tau=0.1$, so that $c=100$, is worth about $106$ demonstrations before any are seen (Figure~\ref{fig:ess}a). The limit is less accurate for a less specific description: at $\tau=1$, the ESS is $5.9$, against $c+d+1=7$ (Appendix~\ref{app:specificity}).

\paragraph{Demonstrations lower the value of a complete description.}
The descriptions modeled so far are \emph{complete}: they constrain every coordinate of $w$. By part (ii), after many demonstrations the dimension no longer adds to such a description's value, which tends to $c-c_0$. With unit noise and isotropic inputs, each demonstration adds expected precision $1$ in every direction. The limit $c-c_0$ is therefore the precision that the description adds to the baseline prior, counted in demonstrations: its conventional prior ESS \citep{morita2008} relative to that prior. Since $c-c_0$ is less than $c+d+1$, a complete description loses value as demonstrations accumulate: a $c=100$ description is worth $106$ demonstrations before any are seen and $103$ after $100$ (Figure~\ref{fig:ess}b).

\paragraph{Demonstrations can raise the value of a partial description.}
Descriptions often specify only part of a task. A \emph{partial} description constrains only $k<d$ of the $d$ coordinates of $w$: it has precision $c$ in those and leaves the others at the baseline prior (Proposition~\ref{prop:partial} in Appendix~\ref{app:partial}). Without demonstrations, its error comes almost entirely from the $d-k$ coordinates it leaves open. Each demonstration is informative about one direction of the task, so $k$ demonstrations without the description also leave $d-k$ directions at the baseline prior: the description is worth about $k$ demonstrations. After many demonstrations, it is worth the precision it adds, averaged over all $d$ coordinates: about $k(c-c_0)/d$ demonstrations, which exceeds $k$ when $c-c_0>d$. At $d=5$ and $\tau_0^2=10$, a $c=100$ partial description with $k=3$ is worth $3.2$ demonstrations before any are seen and $44.0$ after $100$ (Figure~\ref{fig:ess}b).

\section{The Cost of Unreliability}\label{sec:reliability}

\begin{figure*}[t]
\centering
\includegraphics[width=\textwidth]{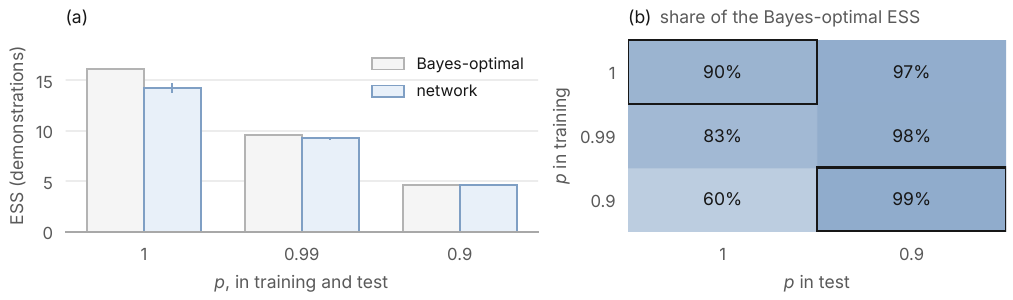}
\caption{\textbf{Meta-trained Transformers approach Bayes-optimal ESS and act on the reliability seen in training.} Results use $c=10$, $d=5$, $\tau_0^2=10$, and no demonstrations. (a) Training and test reliability match. Network ESS uses each network's own baseline curve; bars show the mean and one standard deviation across three training seeds. The Bayes-optimal bars use the same prompts as the networks, so that sampling noise is common to both; Table~\ref{tab:networks} gives larger-sample values. (b) Networks trained at each reliability, evaluated at each test reliability. Cells show network ESS as a share of the Bayes-optimal ESS under the test distribution, estimated on $200{,}000$ prompts; outlined cells have matching reliabilities. Table~\ref{tab:networks} includes $c=1$.}
\label{fig:networks}
\end{figure*}

We now allow the description to be irrelevant with probability $1-p$. Let $\pi_n=P(Z=1\mid m,D_n)$ be the \emph{posterior relevance probability}, the probability that the description is relevant given the demonstrations. The Bayes-optimal predictor averages the predictions of the two components of the prior~\eqref{eq:mixture} with weights $\pi_n$ and $1-\pi_n$ \citep{hoeting1999}.

\begin{proposition}[Unreliable descriptions]\label{prop:unreliable}
Let $0<p<1$.
\begin{enumerate}[label=(\roman*),nosep,leftmargin=*]
\item Without demonstrations, the error is
\begin{equation}\label{eq:zero}
R^{\mathrm{desc}}(0)=\big[(1-p^2)\,\tau_0^2+p^2\,\tau^2\big]\,d.
\end{equation}
\item With $\tau$ fixed, $\ess_0\to p^2d$ as $\tau_0\to\infty$.
\item With $\tau<\tau_0$ both fixed, as $n\to\infty$,
\begin{equation}\label{eq:many}
\ess_n\longrightarrow p\,(c-c_0)-\frac{C_d}{d},
\end{equation}
where $C_d=\E\big[\pi_\infty(1-\pi_\infty)\,\|cm-(c-c_0)w\|^2\big]$ and $\pi_\infty=P(Z=1\mid m,w)$. The limit lies strictly between $0$ and $p(c-c_0)$.
\end{enumerate}
\end{proposition}

\paragraph{Without demonstrations, unreliability imposes an error floor.}
The error in Equation~\eqref{eq:zero} consists of a floor, $(1-p^2)\tau_0^2d$, and $p^2$ times the error $d\tau^2$ of a reliable description. The floor arises because, without demonstrations, nothing shows whether the description is relevant: the posterior relevance probability is still the prior one, $\pi_0=p$, and the Bayes-optimal predictor shrinks the prior mean $m$ by the factor $p$. However specific the description, the error stays above the floor, so the zero-shot ESS stays below a finite ceiling: the number of demonstrations at which the baseline error equals the floor. For a vague baseline prior, $t<d$ demonstrations leave $d-t$ directions of the task at the baseline prior, with error about $(d-t)\tau_0^2$, which equals the floor at $t=p^2d$. The ceiling is therefore $p^2d$, the limit in part (ii), which does not depend on $\tau$. Figure~\ref{fig:ess}a shows the effect at $p=0.9$. Lowering $\tau$ from $1$ to $0.1$, which raises $c$ from $1$ to $100$, multiplies a reliable description's ESS by about eighteen but raises an unreliable one's only from $4.1$ to $4.7$, within $0.01$ of its ceiling.

\paragraph{Demonstrations distinguish low specificity from unreliability.}
By Proposition~\ref{prop:unreliable}, without demonstrations an unreliable description has the same error as a less specific reliable one. For example, at $d=5$ and $\tau_0^2=10$, a $c=100$ description with $p=0.9$ and a reliable one with $c\approx0.52$ both have ESS about $4.7$. After $100$ demonstrations, their ESS values are $83.5$ and $0.45$. Demonstrations provide evidence of whether the unreliable description is relevant, and when it is, its high specificity takes effect; the reliable description has low specificity to begin with, so demonstrations cannot raise its value in this way. The rise is not steady (Figure~\ref{fig:ess}b; Appendix~\ref{app:reliability} explains the early dip).

\paragraph{After many demonstrations, relevance stays uncertain.}
The ESS of an unreliable description levels off below the limit $c-c_0$ of a reliable one, for the two reasons that Equation~\eqref{eq:many} separates. First, a description adds its precision only when it is relevant, which leaves $p(c-c_0)$. Second, the posterior relevance probability never reaches $0$ or $1$: both components of the prior~\eqref{eq:mixture} give every task positive density, so even when the demonstrations have determined the task, it tends to $\pi_\infty$, strictly between $0$ and $1$. The predictor therefore keeps averaging the two components' predictions. These differ by a term of order $1/n$, and averaging them costs $C_d/d$ demonstrations. This cost is positive at every dimension; in our illustrations it is below $0.03$ demonstrations when $d\geq5$ and reaches $1.3$ at $d=1$ (Table~\ref{tab:gap}).

\section{Experiments}\label{sec:experiments}

Sections~\ref{sec:specificity} and~\ref{sec:reliability} give the Bayes-optimal ESS of a description (Table~\ref{tab:summary} in Appendix~\ref{app:proofs} collects their limits). We now ask how close two kinds of in-context learner come to it: Transformers meta-trained on our regression model, which must learn a description's specificity and reliability from their training data, and pretrained language models, whose prompts state both.

\subsection{Meta-Trained Transformers}\label{sec:networks}

Transformers meta-trained on demonstrations can approach Bayes-optimal prediction \citep{garg2022,genewein2025}. We test whether they also attain the Bayes-optimal ESS of a description. Following \citet{garg2022}, we train Transformers from scratch on our regression model, with $d=5$ as in \citet{huangge2025} and $\tau_0^2=10$; we call them \emph{the networks}. We train a separate network for each precision $c\in\{1,10\}$ and each reliability $p\in\{1,0.99,0.9\}$. Half of the training prompts omit the description, which gives each network its own baseline curve. A description token carries the prior mean $m$ but not $\tau$ or $p$, so a network must learn both from its training data (Appendix~\ref{app:networks}).

When training and test reliability match, the networks' zero-shot ESS is close to the Bayes-optimal ESS on the same prompts (Figure~\ref{fig:networks}a). The largest shortfall is at $c=10$ and $p=1$: $14.3$ against $16.1$, a Bayes-optimal value close to $c+d+1=16$. When the two reliabilities differ, the networks follow the training one. Without demonstrations, the Bayes-optimal predictor predicts $p\,m^\top x$, weighting the description by the test reliability $p$; a network instead weights it by about the reliability it was trained at, because the description alone gives no sign of how reliable the test descriptions are. Networks trained at $p=0.9$ thus weight descriptions by about $0.88$ (the coefficient from regressing their predictions on $m^\top x$) even when every test description is relevant, and reach only $60\%$ of the Bayes-optimal ESS (Figure~\ref{fig:networks}b, Table~\ref{tab:networks}). The opposite shift, from $p=1$ in training to $p=0.9$ in testing, costs little (Appendix~\ref{app:networks}).

\subsection{Pretrained Language Models}\label{sec:llm}

The networks learn a description's specificity and reliability from their training distribution. Pretrained language models are not trained on our tasks, so we state both properties in the prompt and ask whether the models use them as a Bayes-optimal predictor would. We use Gaussian mean estimation, the scalar case of our regression model with every input fixed at $x=1$, so that the results reflect how a model uses a description rather than how well it performs linear regression in context. To use a stated precision, a model must weight the description and the demonstrations by their precisions. Reasoning may help it carry out this computation, so we run each model in two modes, with reasoning enabled and with it disabled.

\begin{figure*}[t]
\centering
\includegraphics[width=\textwidth]{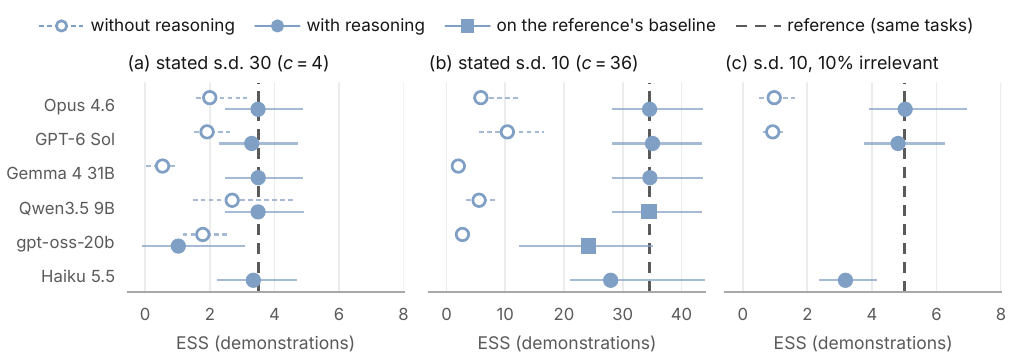}
\caption{\textbf{Language models use the precision and the reliability that a prompt states mainly with reasoning enabled.} Median ESS over $n=1$, $2$, and $4$, from error averaged over $200$ tasks at $\sigma=60$; $95\%$ bootstrap intervals resample tasks and are clipped to the display. The reference's ESS in (a) and (b) is close to $c$. (a) Descriptions generated and stated with standard deviation $30$, so $c=4$. (b) With standard deviation $10$, so $c=36$. (c) GPT-6 Sol, Opus~4.6, and Haiku~5.5 on the descriptions of (b) at reliability $p=0.9$, which the prompt states. Squares in (b) mark two models whose own baseline curves never reach their error with the description; their ESS is measured on the reference's baseline curve and is not comparable with the other points.}
\label{fig:llm}
\end{figure*}

\paragraph{Task and prompts.}
Each task has a latent mean $w$, which we draw from $\mathcal{N}(500,100^2)$. Its demonstrations are numbers drawn from $\mathcal{N}(w,\sigma^2)$, with $\sigma=60$, and rounded to whole numbers. The prompt states $\sigma$, lists $n$ demonstrations, and asks for the next number (Appendix~\ref{app:prompts}). As in Section~\ref{sec:setting}, we score the model's answer by its squared error against the noiseless output, which here is $w$.

A description adds one sentence to the prompt, such as “Its mean was drawn from a normal distribution with mean 510 and standard deviation 30.” The sentence states the description's prior, $\mathcal{N}(m,\tau^2)$, here with prior mean $m=510$ and standard deviation $\tau=30$. The description is relevant: we generate $m$ so that, given $m$, the mean $w$ follows this prior. In Section~\ref{sec:setting}, a demonstration has precision $1$, so a description of precision $c$ is as precise as $c$ demonstrations. Here a demonstration has precision $\sigma^{-2}$ and the description has precision $\tau^{-2}$, so the description is as precise as $\sigma^2/\tau^2$ demonstrations, and we take this number as $c$. We use $\tau=60$, $30$, and $10$, which give $c=1$, $4$, and $36$; Figure~\ref{fig:llm} shows the last two, and Table~\ref{tab:reason} all three.

\paragraph{Models.}
We evaluate six models on the same $200$ tasks. We run five of them in both modes: gpt-oss-20b \citep{openai2025}, Qwen3.5 9B \citep{qwen35}, Gemma~4 31B \citep{gemma4}, GPT-6 Sol \citep{openai2026sol}, and Claude Opus~4.6 \citep{anthropic2026opus}. We run the sixth, Claude Haiku~5.5 \citep{anthropic2026haiku}, only with reasoning, because our setup could not disable it.

\paragraph{Bayesian reference.}
We compare each model with the \emph{Bayesian reference}, the Bayes-optimal predictor given what the prompt states. With a description, the reference predicts $(cm+n\bar y)/(c+n)$: it weights the prior mean $m$ and the demonstrations' average $\bar y$ by their precisions. Without a description, the prompt says only that the mean is “unknown”, because stating a distribution for the mean would itself be a description. The reference then takes the mean to be uniform over the whole numbers $100$ to $999$, the range of the numbers in the prompts. The reference is thus optimal given what the model is told. It does not use $\mathcal{N}(500,100^2)$, the distribution that generates the tasks, because no prompt states it.

\paragraph{Evaluation.}
We measure each model's ESS on its own baseline curve, its error on prompts without a description at $1$ to $64$ demonstrations. The curve starts at one demonstration because models rarely answer with a number when the prompt has neither a description nor demonstrations. With a description, we compute the ESS at $n=1$, $2$, and $4$ and report the median of the three values, the \emph{median ESS}. Because every input is the same, the $d+1$ term of Proposition~\ref{prop:reliable} does not arise, and the reference's ESS is close to $c$: its median ESS is $1.0$, $3.5$, and $34.6$ at $c=1$, $4$, and $36$. Appendix~\ref{app:ess-est} explains why these values differ slightly from $c$ and shows that other ways of estimating the baseline curve change the results little.

\paragraph{Language models use a stated precision mainly with reasoning enabled.}\label{sec:once}\label{sec:reasoning}
Without reasoning, every model makes far less use of a highly specific description than the reference does. At $c=36$, no model reaches a third of the reference's ESS (Figure~\ref{fig:llm}b), and every model's error is well above the reference's (Table~\ref{tab:error}). The prompt states the precision in both modes, so these models do not lack information. They weight the description and the demonstrations differently from the reference (Appendix~\ref{app:rules}).

With reasoning, four of the six models predict nearly as the reference does: GPT-6 Sol, Opus~4.6, Gemma~4 31B, and Qwen3.5 9B. At $c=36$, the reference's root-mean-square error is about $9.4$, and theirs is within $0.3$ of it at every demonstration count (Table~\ref{tab:error}). Opus~4.6, GPT-6 Sol, and Gemma~4 31B also attain the reference's ESS at both precisions (Figure~\ref{fig:llm}a,b). Qwen3.5 9B's ESS at $c=36$ is undefined: its baseline curve stops falling after about $32$ demonstrations and never reaches its error with the description. The square in Figure~\ref{fig:llm}b measures its ESS on the reference's baseline curve instead. The other two models show that reasoning alone is not enough. Haiku~5.5's error is well above the reference's when it has a single demonstration (Table~\ref{tab:error}). gpt-oss-20b makes arithmetic errors, in about $2\%$ of its replies with a description and when it averages many demonstrations (Appendix~\ref{app:reasoning-runs}). At $c=4$, its ESS with reasoning is below its ESS without; at $c=36$, its ESS is undefined for the same reason as Qwen3.5 9B's (the other square).

For the three open-weight models, gpt-oss-20b, Qwen3.5 9B, and Gemma~4 31B, the two runs differ in more than reasoning: in serving software, in decoding, in how we read out the prediction, or in retry policy. Opus~4.6's two runs differ only in the reasoning setting, and they show the same shortfall without reasoning (Appendix~\ref{app:reasoning-runs}).

A model could do well with $c=36$ descriptions without using the stated precision, because a more specific description also has a more accurate prior mean. As a control, we give the same $c=36$ descriptions but state their precision as $c=4$. The reference's ESS then falls from $34.6$ to $13.7$; it stays above $4$ because the prior mean is as accurate as before. The ESS of Opus~4.6, GPT-6 Sol, and Gemma~4 31B with reasoning falls to between $11$ and $14$ (Table~\ref{tab:reason}), so these models use the stated precision itself.

\paragraph{When the prompt says what holds if a description is wrong, language models with reasoning rely less on a description that the demonstrations contradict.}\label{sec:llm-stated}
Section~\ref{sec:reliability} shows that demonstrations provide evidence of whether an unreliable description is relevant. To see whether language models use this evidence, we give the $c=36$ descriptions a reliability of $p=0.9$ and state it in the prompt: “There is a 10\% chance that this statement is wrong, in which case the mean was instead drawn uniformly at random from the whole numbers 100 to 999.” We present each task twice, once with its relevant description and once with an irrelevant one, whose prior mean we draw independently of the task, and we average the two errors with weights $0.9$ and $0.1$. We run this experiment on the three models that our subscriptions cover: GPT-6 Sol, Opus~4.6, and Haiku~5.5 (Appendix~\ref{app:llm-reliability}). At this reliability, the reference's ESS falls from $34.6$ to $5.0$ (Figure~\ref{fig:llm}c). Nearly all of the $5.0$ comes from updating the posterior relevance probability: a predictor that keeps it at $0.9$, whatever the demonstrations show, has an ESS of only $0.3$. With reasoning, GPT-6 Sol and Opus~4.6 attain the reference's ESS and Haiku~5.5 about two thirds of it, far above the predictor that does not update. Without reasoning, GPT-6 Sol and Opus~4.6 reach an ESS of about $1$, little above that predictor's $0.3$ (Table~\ref{tab:matched}).

\section{Related Work}

\paragraph{Theories of ICL.}
Prior work studies ICL as Bayesian inference over latent tasks \citep{xie2022,panwar2024}, as a learning algorithm implemented in a forward pass \citep{akyurek2023,vonoswald2023}, and as meta-learning across tasks \citep{garg2022,muller2022}. Agreement of meta-trained predictors with Bayesian ones can depend on training-task diversity and on the function class \citep{raventos2023,panwar2024}. \citet{lin2024} study a prior that mixes task groups, a mixture like the one an unreliable description induces here. In these accounts, the prior over tasks comes from pretraining or meta-training. The works closest to ours put prior information in the prompt, but each analyzes one particular kind \citep{huangge2025,lin2025,zhu2026}. We differ in characterizing a description by two properties that any description has, its specificity and its reliability, and in measuring its value in demonstrations.

\paragraph{Prior effective sample size.}
Prior ESS expresses the information or impact of a prior as an equivalent sample size, with different definitions matching different statistical quantities \citep{morita2008,neuenschwander2020}. Our measure is closest to those based on expected or predictive loss \citep{reimherr2021,wiesenfarth2020,selby2025}. We match prediction error for task descriptions, at each demonstration count and against each predictor's own baseline curve. Appendix~\ref{app:related} discusses these and further connections.
\paragraph{Descriptions and evidence in language models.}
\citet{gupta2025} give language models an explicitly stated coin bias together with observed flips, evaluate their predictions against Bayesian predictors, and find that in the models' predictions the observed flips outweigh the stated bias. The models we test show the same pattern without reasoning, but most do not with reasoning enabled. Our experiments also state the precision and the reliability of a description separately, and measure its value in demonstrations.

\section{Discussion}\label{sec:discussion}

We asked how descriptions can be characterized and how well models use them in practice. For the first question, we treated a description as a prior with a specificity and a reliability, and we measured it by its predictive ESS. A reliable description can replace many demonstrations; unreliability imposes a finite ESS ceiling before demonstrations are observed; and demonstrations can then raise a description's ESS by supplying evidence about its relevance. For the second, meta-trained Transformers approach the Bayes-optimal ESS and, without demonstrations, weight a description by how often descriptions in their training data were relevant. Language models make too little use of a specific description without reasoning; with reasoning, most predict nearly as the Bayesian reference does. When descriptions were irrelevant at a stated rate, the three models tested with reasoning also relied less on a description that the demonstrations contradicted. Treating a description as a prior thus extends the Bayesian analysis of ICL from demonstrations to descriptions.

Our controlled analysis suggests why a description can substitute for many demonstrations \citep{honda2025} or have limited influence \citep{webson2022,gupta2025,gloaguen2026}. A description has little influence when it says little about the task, when it may be about the wrong task, when the demonstrations already supply what it says, or when the model does not use it. A Bayesian predictor's ESS reflects the first three, and a model's shortfall from it reflects the fourth. For prompt design, the results suggest that an instruction that may be wrong should come with demonstrations, which let the model check it: without them, its value is capped however specific it is (Section~\ref{sec:reliability}).

\paragraph{Limitations and future work.}
We model a relevant description as a single Gaussian prior around one prior mean: a description that points to one task and is more or less specific about it. Future work could model other forms of description, such as one that names several candidate tasks, with a mixture of Gaussians, which can approximate a wide class of priors \citep{dalal1983}. Our model also treats relevance as all or nothing; a description that is right about some coordinates of the task and wrong about others would need a relevance indicator for each coordinate. Most of our language-model prompts state a description's precision and reliability explicitly, as numbers, so that every prediction can be compared with a Bayesian reference. In practice, these properties are implicit: a more specific instruction is typically a longer one that says more about the task, and an instruction rarely says how likely it is to be about the right task. How well models infer these properties from the wording of realistic instructions is the next question (Appendix~\ref{app:semantic} takes a first step), and predictive ESS applies to it directly, since it requires only a model's errors with and without the description. Our language-model experiments also use small synthetic tasks, and the test of stated reliability covers three of the models.

\section*{Impact Statement}
This work studies how models combine task descriptions with demonstrations. Its measurement framework can inform prompt evaluation and design when descriptions might be about the wrong task or leave much of it open. The experiments use synthetic tasks; their results do not establish dependable instruction following in consequential applications.
\bibliographystyle{icml2026}
\bibliography{refs}
\clearpage
\onecolumn
\appendix
\section{Proofs}\label{app:proofs}

Table~\ref{tab:summary} collects the limits proved in this appendix.

\begin{table}[!htb]
\centering\small
\caption{\textbf{The Bayes-optimal value of a description, in demonstrations.} Limits of the ESS of a description with precision $c=\tau^{-2}$: without demonstrations, for a vague baseline prior ($\tau_0\to\infty$ with $\tau$ fixed), and after many demonstrations ($n\to\infty$ with $\tau_0$ fixed, $c_0=\tau_0^{-2}$). The constant $C_d>0$ is defined in Proposition~\ref{prop:unreliable}(iii). The limits are proved in Propositions~\ref{prop:reliable}, \ref{prop:unreliable}, and~\ref{prop:partial}.}
\label{tab:summary}
\vspace{3pt}
\begin{tabular}{lcc}
\toprule
& \multicolumn{2}{c}{Demonstrations} \\
\cmidrule(lr){2-3}
Description & none & many \\
\midrule
Reliable & $c+d+1$ & $c-c_0$ \\
Unreliable, $p<1$ & $p^2d$ & $p\,(c-c_0)-C_d/d$ \\
Partial, $k$ of $d$ coordinates & $k$ & $k\,(c-c_0)/d$ \\
\bottomrule
\end{tabular}
\end{table}

\subsection{Reliable Descriptions}\label{app:specificity}

\paragraph{Notation and error.}
As in Section~\ref{sec:setting}, $c_0=1/\tau_0^2$ is the baseline prior precision and $c=1/\tau^2$ is the description's. Let $X_n\in\R^{n\times d}$ contain the demonstration inputs as rows, and set $W_n=X_n^\top X_n$. For a prior precision $\kappa>0$, define
\[
R^{\kappa}(n)=\E\operatorname{tr}(\kappa I_d+W_n)^{-1}.
\]
Thus $R^{\mathrm{base}}=R^{c_0}$ and, for a reliable description, $R^{\mathrm{desc}}=R^c$. Here and throughout, error is the expected squared prediction error against the noiseless query output $w^\top x$. Both identities hold because the Bayes-optimal prediction is the posterior mean. Given the prompt and a query $x$, its expected squared error is $x^\top\Sigma x$, where $\Sigma=(\kappa I_d+W_n)^{-1}$ is the posterior covariance; averaging over the independent query, with $\E[xx^\top]=I_d$, gives $\operatorname{tr}\Sigma$.

Throughout the proofs, $d$ is a positive integer, $0<\tau\le\tau_0$, so that $c\ge c_0$, $s=1-\tau^2/\tau_0^2\in[0,1)$, and $p\in[0,1]$. Specificity is the relevant component's precision $c=\tau^{-2}$; $s$ is the share of the baseline variance that the description removes. Reliability $p$ is the prior probability of relevance. Noninteger demonstration counts refer to linear interpolation between the errors at adjacent integers. We call the count $n^*$ of Definition~\ref{def:ess} the \emph{crossing}.

\paragraph{Wishart identities.}
We use four standard properties of $W_n$.
\begin{enumerate}[label=(W\arabic*),leftmargin=*]
    \item If $n<d$, then $W_n$ has rank $n$ almost surely. Consequently, $(\kappa I_d+W_n)^{-1}$ has $d-n$ eigenvalues equal to $1/\kappa$.
    \item If $n\ge d$ and $e$ is a fixed unit vector, then $(e^\top W_n^{-1}e)^{-1}\sim\chi^2_{n-d+1}$ \citep[Theorem~3.2.12]{muirhead1982}. Hence
    \[
    \E\operatorname{tr}W_n^{-1}=\frac{d}{n-d-1}\quad(n\ge d+2),
    \]
    whereas this expectation is infinite for $n=d$ and $n=d+1$.
    \item For fixed $d$, as $n\to\infty$,
    \[
    \E\operatorname{tr}W_n^{-2}=\frac{d}{n^2}+O(n^{-3}),
    \qquad
    \E\operatorname{tr}W_n^{-3}=O(n^{-3})
    \]
    \citep{vonrosen1988}.
    \item For fixed $d$ and every fixed integer $k\ge1$, $\E(\operatorname{tr}W_n^{-1})^k=O(n^{-k})$ as $n\to\infty$. Indeed, by (W2), $\operatorname{tr}W_n^{-1}$ is a sum of $d$ reciprocals of $\chi^2_{n-d+1}$ variables, and $\E(\chi^2_\nu)^{-k}=[(\nu-2)(\nu-4)\cdots(\nu-2k)]^{-1}$ for $\nu>2k$; Minkowski's inequality bounds the $k$th moment of the sum.
\end{enumerate}
The Sherman--Morrison identity implies that an additional demonstration strictly decreases the expected trace of the posterior covariance. Thus $R^\kappa(n)$ is strictly decreasing, and (W2) implies that it tends to zero. Its linear interpolation is therefore continuous and strictly decreasing. In the correctly specified joint model, conditioning on the description cannot increase expected posterior variance, so $R^{\mathrm{desc}}(n)\le R^{\mathrm{base}}(n)$ and the Bayes-optimal ESS is nonnegative.

\paragraph{Vague baseline prior.}
Several results use the baseline error measured in units of the baseline variance $\tau_0^2$. For every integer $n$,
\[
\frac{R^{c_0}(n)}{\tau_0^2}
=\E\operatorname{tr}(I_d+\tau_0^2W_n)^{-1}
\longrightarrow(d-n)^+
\qquad\text{as }\tau_0\to\infty.
\]
Indeed, an eigenvalue $1/(1+\tau_0^2\omega)$, where $\omega\ge0$ is an eigenvalue of $W_n$, is bounded by one and converges to the indicator of $\omega=0$, so dominated convergence and the almost-sure rank of $W_n$ apply. The same limit holds at every real $t\ge0$ by interpolation. Each finite-$\tau_0$ curve is strictly decreasing, and monotone functions that converge pointwise to a continuous limit converge locally uniformly. Hence, if an error level $\ell(\tau_0)$ satisfies $\ell(\tau_0)/\tau_0^2\to\lambda$ with $0<\lambda<d$, the count at which the baseline error equals $\ell(\tau_0)$ converges to $d-\lambda$, the only point where $(d-t)^+=\lambda$.

\begin{proof}[Proof of Proposition~\ref{prop:reliable}]
\emph{(i) No demonstrations.}
A reliable description without demonstrations has error $R^{\mathrm{desc}}(0)=d\tau^2$, the trace of its prior covariance. For fixed $\tau$, this error, $d/c$, does not depend on $\tau_0$. As $\tau_0\to\infty$, monotone convergence, (W1), and (W2) give
\[
R^{c_0}(n)\longrightarrow
\begin{cases}
+\infty,&n\le d+1,\\[2pt]
\displaystyle\frac{d}{n-d-1},&n\ge d+2.
\end{cases}
\]
The finite part of this limiting curve equals $d/c$ at the integer $n=c+d+1$. For every $t<c+d+1$ the limiting error exceeds $d/c$, and for every $t>c+d+1$ it is smaller, so the crossing of the interpolated finite-$\tau_0$ error converges to $c+d+1$. The baseline curve is interpolated linearly between integer counts, so the limit is exactly $c+d+1$ only for integer $c$, and the proposition is stated for that case.

\emph{(ii) Many demonstrations.}
This is the case $k=d$ of Proposition~\ref{prop:partial}(ii), proved in Appendix~\ref{app:partial}.
\end{proof}

At a finite $\tau_0$, the limit in part (i) is a good guide when the description is much more precise than the baseline prior, $c\gg c_0$. At $d=5$ and $\tau_0^2=10$, a $c=100$ description is worth $105.9$ demonstrations, against $c+d+1=106$, and a $c=1$ description is worth $5.9$, against $7$ (Table~\ref{tab:grid} gives other dimensions and baseline variances).

\paragraph{A fixed share of the baseline variance.}
The limit $\tau_0\to\infty$ can also be taken with the description removing a fixed share $s=1-\tau^2/\tau_0^2$ of the baseline variance, so that $\tau$ grows in proportion to $\tau_0$. A predictor with $t<d$ demonstrations and no description knows nothing about $d-t$ directions of $w$, and as $\tau_0\to\infty$ its error is dominated by what those directions contribute, $(d-t)\tau_0^2$. The description alone leaves an error of $d\tau^2=(1-s)d\tau_0^2$, which this predictor matches at $t=sd$: a description that removes a share of the uncertainty is worth that share of $d$ demonstrations.

\begin{proposition}[Fixed share]\label{prop:share}
Let $p=1$ and let the share $s\in[0,1)$ be fixed. Then $\ess_0\geq sd$, and $\ess_0\to sd$ as $\tau_0\to\infty$.
\end{proposition}

\begin{proof}
By (W1), $R^{c_0}(n)\ge(d-n)\tau_0^2$ for $n<d$; the inequality also holds trivially for $n\ge d$. Linear interpolation preserves this bound, so
\[
R^{\mathrm{base}}(t)\ge(d-t)\tau_0^2\qquad(t\ge0).
\]
The error with the description is $d\tau^2=(1-s)d\tau_0^2$. For $t<sd$, the right-hand side above exceeds it. The crossing $n^*$ of Definition~\ref{def:ess} therefore satisfies $n^*\ge sd$, so $\ess_0\ge sd$. In units of $\tau_0^2$, the error with the description is $(1-s)d$. For $s\in(0,1)$, this level lies strictly between $0$ and $d$, so by the vague-baseline limit the crossing converges to $d-(1-s)d=sd$. For $s=0$, the description is uninformative and $\ess_0=0$ for every $\tau_0$.
\end{proof}

\subsection{Unreliable Descriptions}\label{app:reliability}

The posterior relevance probability is $\pi_n=P(Z=1\mid m,D_n)$, whereas $p$ remains the prior relevance probability. Conditional on $Z=1$, the posterior parameters are
\[
\Sigma_1=(cI_d+W_n)^{-1},
\qquad
\mu_1=\Sigma_1(c m+X_n^\top y_{1:n}).
\]
Conditional on $Z=0$, the description's mean $m$ is independent of $w$, and
\[
\Sigma_0=(c_0I_d+W_n)^{-1},
\qquad
\mu_0=\Sigma_0X_n^\top y_{1:n}.
\]
Because $m$ has the same distribution under both relevance states, its density cancels from the posterior odds, and the posterior relevance probability follows from the two Gaussian marginal likelihoods of the demonstrations,
\begin{align*}
L_1&=\mathcal N\!\left(y_{1:n};X_nm,I_n+\tau^2X_nX_n^\top\right),\\
L_0&=\mathcal N\!\left(y_{1:n};0,I_n+\tau_0^2X_nX_n^\top\right),\\
\pi_n&=\frac{pL_1}{pL_1+(1-p)L_0}.
\end{align*}
The query input does not enter this update because it is independent of $(Z,m,D_n)$. Without demonstrations, both marginal likelihoods equal one, so $\pi_0=p$. The Bayes-optimal prediction is the posterior mean, $\big(\pi_n\mu_1+(1-\pi_n)\mu_0\big)^\top x$ \citep{hoeting1999}.

\begin{proof}[Proof of Proposition~\ref{prop:unreliable}]
\emph{(i) No demonstrations.}
At $n=0$, $\mu_1=m$, $\mu_0=0$, and $\pi_0=p$, so the prediction is $p\,m^\top x$. For each coordinate,
\[
\operatorname{Var}(w_i)=\tau_0^2,\qquad
\operatorname{Var}(m_i)=\tau_0^2-\tau^2=s\tau_0^2,\qquad
\operatorname{Cov}(w_i,m_i)=ps\tau_0^2.
\]
It follows that
\[
\E(w_i-pm_i)^2
=\tau_0^2-2p^2s\tau_0^2+p^2s\tau_0^2
=(1-p^2s)\tau_0^2
=(1-p^2)\tau_0^2+p^2\tau^2.
\]
Since $\E[xx^\top]=I_d$, the prediction error is $\E\|w-pm\|^2$. Summing over coordinates proves Equation~\eqref{eq:zero}. Because $p\,m$ is the mean of the mixture prior~\eqref{eq:mixture} given $m$, this error is the trace of that prior's covariance, averaged over $m$.

\emph{(ii) The ceiling and its limit.}
Since $R^{\mathrm{base}}(0)=d\tau_0^2$ and $s<1$, Equation~\eqref{eq:zero} gives, for every $\tau>0$,
\[
R^{\mathrm{desc}}(0)
=(1-p^2s)R^{\mathrm{base}}(0)
>(1-p^2)R^{\mathrm{base}}(0).
\]
The error on the left decreases to the right-hand side as $\tau\to0$, equivalently $s\to1$. The interpolated baseline error is continuous, strictly decreasing, and tends to zero. Let $n_p^{\sharp}$, the ceiling, be the count at which the baseline error equals the floor $(1-p^2)\tau_0^2d$; it is unique and finite. The crossing $n^*$ is strictly below $n_p^{\sharp}$ and converges to it as $\tau\to0$. Finally, the error level that defines $n_p^{\sharp}$ is $(1-p^2)d$ in units of $\tau_0^2$, which lies strictly between $0$ and $d$. By the vague-baseline limit of Appendix~\ref{app:specificity}, $n_p^{\sharp}\to d-(1-p^2)d=p^2d$ as $\tau_0\to\infty$. With $\tau$ fixed instead of $s$, $s\to1$ as $\tau_0\to\infty$, and the same limit gives $\ess_0\to p^2d$.

For comparison, a predictor that also observes $Z$ predicts $m^\top x$ when $Z=1$ and $0$ when $Z=0$, with zero-shot error
\[
p\,d\tau^2+(1-p)\,d\tau_0^2=(1-ps)\,d\tau_0^2.
\]
As $\tau\to0$ this error tends to $(1-p)d\tau_0^2$, so by the same limit its zero-shot ESS ceiling approaches $pd$ as $\tau_0\to\infty$, against $p^2d$ for the Bayes-optimal predictor, which does not observe $Z$. The difference, $p(1-p)d$, is the cost of not knowing whether the description is relevant.

\emph{(iii) Many demonstrations.}
Write $A=cm-(c-c_0)w$, so that $C_d=\E[\pi_\infty(1-\pi_\infty)\|A\|^2]$. The proof has six parts: the error, the difference between the two components' posterior means, the posterior relevance probability, the limit of the remainder, the crossing, and an equivalent form of the limit that shows it is strictly positive.

\emph{Part I: the error.} Given $(m,D_n)$, the task has the two-component posterior $\pi_n\,\mathcal N(\mu_1,\Sigma_1)+(1-\pi_n)\,\mathcal N(\mu_0,\Sigma_0)$, whose covariance is
\[
\pi_n\Sigma_1+(1-\pi_n)\Sigma_0+\pi_n(1-\pi_n)(\mu_1-\mu_0)(\mu_1-\mu_0)^\top.
\]
The error is the expected trace of this covariance. The matrices $\Sigma_1$ and $\Sigma_0$ depend only on the inputs, and $\E[\pi_n\mid X_n]=p$ because $Z$ is independent of the inputs. Hence
\[
R^{\mathrm{desc}}(n)=pR^{c}(n)+(1-p)R^{c_0}(n)+V_n,
\qquad
V_n=\E\big[\pi_n(1-\pi_n)\|\mu_1-\mu_0\|^2\big].
\]
The first two terms are the error of a predictor that also observes $Z$, which is the Bayes-optimal predictor for the component that $Z$ selects. The between-component term $V_n\ge0$ is what the Bayes-optimal predictor loses by not observing $Z$.

\emph{Part II: the difference between the component means.} The term $V_n$ depends only on the distribution of $(m,w,D_n)$, so we may take the demonstrations for every $n$ to be the first $n$ of one infinite sequence. Write $b_n=X_n^\top y_{1:n}$, so that $\mu_1=\Sigma_1(b_n+cm)$ and $b_n=(c_0I_d+W_n)\mu_0$. Then
\[
\mu_1-\mu_0=\Sigma_1\big\{cm-(c-c_0)\mu_0\big\}.
\]
With $y_{1:n}=X_nw+\epsilon_{1:n}$,
\[
\mu_0-w=-c_0\Sigma_0w+\Sigma_0X_n^\top\epsilon_{1:n}.
\]
By the strong law of large numbers, $W_n/n\to I_d$ and $X_n^\top\epsilon_{1:n}/n\to0$ almost surely, so $n\Sigma_1\to I_d$, $n\Sigma_0\to I_d$, and $\mu_0\to w$. Hence
\[
n(\mu_1-\mu_0)\longrightarrow A=cm-(c-c_0)w\qquad\text{almost surely.}
\]

\emph{Part III: the posterior relevance probability.} For $n\ge d$, $W_n$ is invertible almost surely; let $\hat w_n=W_n^{-1}b_n$ be the least-squares estimate. Given $(w,X_n)$, $\hat w_n\sim\mathcal N(w,W_n^{-1})$, and the part of $y_{1:n}$ orthogonal to the columns of $X_n$ is independent of $\hat w_n$ and has a distribution that does not depend on $(w,m,Z)$. This part therefore cancels from the ratio of the two components' marginal likelihoods. Given $(m,X_n)$, $\hat w_n$ is distributed as $\mathcal N(m,\tau^2I_d+W_n^{-1})$ when $Z=1$ and as $\mathcal N(0,\tau_0^2I_d+W_n^{-1})$ when $Z=0$, so
\[
\pi_n=\frac{p\,\mathcal N(\hat w_n;m,\tau^2I_d+W_n^{-1})}{p\,\mathcal N(\hat w_n;m,\tau^2I_d+W_n^{-1})+(1-p)\,\mathcal N(\hat w_n;0,\tau_0^2I_d+W_n^{-1})}.
\]
Since $\hat w_n-w=(W_n/n)^{-1}X_n^\top\epsilon_{1:n}/n\to0$ and $W_n^{-1}\to0$ almost surely, and Gaussian densities are continuous in their mean and covariance and positive,
\[
\pi_n\longrightarrow\pi_\infty=P(Z=1\mid m,w)
=\frac{p\,\mathcal N(w;m,\tau^2I_d)}{p\,\mathcal N(w;m,\tau^2I_d)+(1-p)\,\mathcal N(w;0,\tau_0^2I_d)}
\qquad\text{almost surely,}
\]
which lies strictly between $0$ and $1$. Many demonstrations thus identify the task but not whether the description is relevant.

\emph{Part IV: the limit of $n^2V_n$.} By Parts II and III, $n^2\pi_n(1-\pi_n)\|\mu_1-\mu_0\|^2\to\pi_\infty(1-\pi_\infty)\|A\|^2$ almost surely. To pass to expectations, we bound the fourth moment of $n\|\mu_1-\mu_0\|$. For $n\ge d$, both $\|n\Sigma_1\|_{\mathrm{op}}$ and $\|n\Sigma_0\|_{\mathrm{op}}$ are at most $M_n=n\operatorname{tr}W_n^{-1}$, and $\E M_n^k=O(1)$ for every fixed $k$ by (W4). Part II gives
\[
n(\mu_1-\mu_0)=n\Sigma_1A+(c-c_0)\,n\Sigma_1\,c_0\Sigma_0w-(c-c_0)\,n\Sigma_1\,\Sigma_0X_n^\top\epsilon_{1:n},
\]
and $\|c_0\Sigma_0w\|\le\|w\|$. The inputs are independent of $(m,w)$, so the fourth moments of the first two terms are at most $\E M_n^4\,\E\|A\|^4$ and $(c-c_0)^4\,\E M_n^4\,\E\|w\|^4$. Given $X_n$, the vector $\Sigma_0X_n^\top\epsilon_{1:n}$ is Gaussian with covariance $\Sigma_0W_n\Sigma_0\preceq W_n^{-1}$, so its conditional fourth moment is at most $3(\operatorname{tr}W_n^{-1})^2=3M_n^2/n^2$, and the fourth moment of the third term is at most $3(c-c_0)^4\,\E M_n^6/n^2$. Since $(a_1+a_2+a_3)^4\le27(a_1^4+a_2^4+a_3^4)$ for real numbers $a_1$, $a_2$, and $a_3$, $\sup_n\E\|n(\mu_1-\mu_0)\|^4<\infty$ over all large $n$. Since $\pi_n(1-\pi_n)\le1/4$, the variables $n^2\pi_n(1-\pi_n)\|\mu_1-\mu_0\|^2$ are uniformly integrable, and
\[
n^2V_n\longrightarrow C_d=\E\big[\pi_\infty(1-\pi_\infty)\|A\|^2\big].
\]
The uncertainty about relevance thus adds error at order $n^{-2}$, the order at which one more demonstration lowers the error.

\emph{Part V: the crossing.} The expansion in the proof of Proposition~\ref{prop:partial}(ii), applied with $\Lambda=cI_d$ and with $\Lambda=c_0I_d$, and Part IV give, uniformly over real $t$ in any bounded interval $[0,T]$,
\[
R^{\mathrm{base}}(n+t)-R^{\mathrm{desc}}(n)
=\frac{d}{n^2}\Big[p\,(c-c_0)-\frac{C_d}{d}-t+o(1)\Big].
\]
Let $L=p(c-c_0)-C_d/d$. Because $R^{\mathrm{desc}}(n)\le R^{\mathrm{base}}(n)$, the case $t=0$ shows that $L\ge0$. The crossing $n^*$ exists, because the interpolated baseline error is continuous and tends to zero (Appendix~\ref{app:specificity}). Fix $\varepsilon>0$. For all large $n$, the right-hand side is negative at $t=L+\varepsilon$, so $R^{\mathrm{base}}(n+L+\varepsilon)<R^{\mathrm{desc}}(n)$; the baseline error is strictly decreasing, so $n^*<n+L+\varepsilon$. If $L>\varepsilon$, the right-hand side is positive at $t=L-\varepsilon$, so $n^*>n+L-\varepsilon$; if $L\le\varepsilon$, the same bound holds because $\ess_n\ge0$. Hence $|\ess_n-L|\le\varepsilon$ for all large $n$, which proves Equation~\eqref{eq:many}.

\emph{Part VI: an equivalent form and strictness.} Because $\pi_\infty=P(Z=1\mid m,w)$, $\E[\pi_\infty\,g(m,w)]=p\,\E[g(m,w)\mid Z=1]$ for every integrable $g$. Given $Z=1$, $w=m+\xi$ with $\xi\sim\mathcal N(0,\tau^2I_d)$ independent of $m$, so $A=c_0m-(c-c_0)\xi$ and
\[
\E\big[\|A\|^2\mid Z=1\big]=d\big[c_0^2(\tau_0^2-\tau^2)+(c-c_0)^2\tau^2\big]=d\,(c-c_0).
\]
Hence $\E[\pi_\infty\|A\|^2]=pd\,(c-c_0)$, and since $\pi_\infty^2=\pi_\infty-\pi_\infty(1-\pi_\infty)$,
\begin{equation}\label{eq:positive}
L=p\,(c-c_0)-\frac{C_d}{d}=\frac1d\,\E\big[\pi_\infty^2\|A\|^2\big].
\end{equation}
Because $0<\pi_\infty<1$ and, since $\tau<\tau_0$, $A\neq0$ almost surely, both $C_d$ and the expectation in~\eqref{eq:positive} are positive. Therefore $0<L<p(c-c_0)$.
\end{proof}

\begin{corollary}[Score information]\label{cor:score}
Let $g_m(w)=p\,\mathcal N(w;m,\tau^2I_d)+(1-p)\,\mathcal N(w;0,\tau_0^2I_d)$ be the density of the prior~\eqref{eq:mixture} given the description, and let $J=\E\|\nabla_w\log g_m(w)\|^2$. Under the conditions of Proposition~\ref{prop:unreliable}(iii),
\[
\lim_{n\to\infty}\ess_n=\frac{J-dc_0}{d}.
\]
\end{corollary}

\begin{proof}
Differentiating the mixture gives $\nabla_w\log g_m(w)=-c_0w+\pi_\infty A$. Marginally $w\sim\mathcal N(0,\tau_0^2I_d)$, so $\E\|c_0w\|^2=dc_0$. The cross term vanishes: $\E[\pi_\infty A\mid w]=\E[\mathbf 1\{Z=1\}A\mid w]=p\,\E[A\mid w,Z=1]$, because $w$ has the same distribution under both values of $Z$, and given $Z=1$, $\E[m\mid w]=(1-c_0/c)\,w$, so $\E[A\mid w,Z=1]=0$. Hence $J=dc_0+\E[\pi_\infty^2\|A\|^2]=dc_0+dL$ by Equation~\eqref{eq:positive}.
\end{proof}

The quantity $J$ is the expected squared score of the description's prior, the prior-information term of the Bayesian Cram\'er--Rao bound \citep{gill1995}; it is not the Fisher information of the regression likelihood. The same expression gives the other limits after many demonstrations. A Gaussian prior with precision matrix $\Lambda$ has $J=\operatorname{tr}\Lambda$, so $(J-dc_0)/d$ is $c-c_0$ for a reliable complete description (Proposition~\ref{prop:reliable}(ii)) and $k(c-c_0)/d$ for a partial one (Proposition~\ref{prop:partial}(ii)). In each of these models, after many demonstrations a description is worth the score information that its prior adds to the baseline prior's, averaged over the $d$ directions.

Part (i) also connects an unreliable description to Proposition~\ref{prop:share}. Equation~\eqref{eq:zero} equals $(1-p^2s)\,d\tau_0^2$, the zero-shot error of a reliable description that removes the share $p^2s$ of the baseline variance. The proof of Proposition~\ref{prop:share} uses the description only through this error, so it applies with $p^2s$ in place of $s$: $\ess_0\ge p^2sd$, and $\ess_0\to p^2sd$ as $\tau_0\to\infty$ with $s$ fixed.

\paragraph{The two terms of the limit.}
The first term of Equation~\eqref{eq:many}, $p(c-c_0)$, is the limit for the predictor that observes $Z$, which gains the added precision $c-c_0$ with probability $p$. The second, $C_d/d$, is what the Bayes-optimal predictor loses by remaining uncertain whether the description is relevant. The two components' posterior means differ by about $A/n$, so the between-component term $V_n$ is of order $n^{-2}$. One more demonstration lowers the error by $d/n^2$ to leading order, the same order, so the term costs about $C_d/d$ demonstrations once many are seen. Table~\ref{tab:gap} estimates $C_d/d$ in two ways that agree: from its defining expectation, and by simulating $n^2V_n/d$ at $n=400$. By Part VI of the proof, this cost is strictly positive at every dimension. In these illustrations it is below $10^{-4}$ demonstrations at $d=16$ and largest at $d=1$. This pattern is numerical: the proposition does not establish that $C_d/d$ decreases monotonically with the dimension. One reason to expect small values in high dimension, when the description is much more specific than the baseline prior, is that the log odds of relevance given the task are a sum over the $d$ coordinates, so $\pi_\infty$ is then close to $0$ or $1$ for almost every task and $\pi_\infty(1-\pi_\infty)$ is close to zero; in the scalar case, many tasks are plausible under both components. At $d=5$, $\tau_0^2=10$, $c=100$, and $p=0.9$, where the limit is $89.91$, the ESS approaches it slowly: it is $83.5$ at $n=100$, $88.7$ at $n=400$, and $89.6$ at $n=1{,}000$.

\paragraph{The ceiling beyond the Gaussian model.}
The existence of a zero-shot ceiling does not depend on the Gaussian form of the two priors. For any $Q_{\mathrm{desc}}$ and $Q_0$ with finite second moments, the covariance of the mixture prior~\eqref{eq:mixture} is at least $1-p$ times that of $Q_0$ in the positive semidefinite order, since it is $p$ times the covariance of $Q_{\mathrm{desc}}$, plus $1-p$ times that of $Q_0$, plus a positive semidefinite between-component term. Under squared error with $\E[xx^\top]=I_d$, the zero-shot error is the trace of the prior covariance, so $R^{\mathrm{desc}}(0)\ge(1-p)R^{\mathrm{base}}(0)$ however specific the description. When demonstrations alone bring the baseline error down to $(1-p)R^{\mathrm{base}}(0)$, the zero-shot ESS is at most the number of demonstrations at which they do. The values of the ceiling, of its limit $p^2d$, and of the limits in Table~\ref{tab:summary} are specific to Gaussian linear regression with isotropic inputs.

\begin{table}[!htb]
\centering\small
\caption{\textbf{After many demonstrations, an unreliable description's ESS is within $0.03$ demonstrations of $p(c-c_0)$ at $d\geq5$ and further below it at $d=1$.} The two terms of the limit in Proposition~\ref{prop:unreliable}(iii) and the limit itself, at $\tau_0^2=10$. The term $C_d/d$ is a Monte Carlo estimate of its defining expectation from two million draws of $(m,w)$. The next column estimates the same quantity by simulating $n^2V_n/d$ at $n=400$, on $20{,}000$ prompts ($4{,}000$ at $d=16$). Standard errors are in parentheses.}
\label{tab:gap}
\vspace{3pt}
\begin{tabular}{rrrcccc}
\toprule
$d$ & $c$ & $p$ & $p(c-c_0)$ & $C_d/d$ & $n^2V_n/d$, $n=400$ & limit \\
\midrule
$16$ & $100$ & $0.9$ & $89.91$ & $<0.0001$ & $<0.0001$ & $89.91$ \\
$16$ & $1$ & $0.9$ & $0.81$ & $<0.0001$ & $<0.0001$ & $0.81$ \\
$5$ & $10$ & $0.9$ & $8.91$ & $0.0020$ $(0.0001)$ & $0.0024$ $(0.0007)$ & $8.91$ \\
$5$ & $1$ & $0.9$ & $0.81$ & $0.0099$ $(<0.0001)$ & $0.0104$ $(0.0005)$ & $0.80$ \\
$5$ & $1$ & $0.5$ & $0.45$ & $0.0232$ $(0.0001)$ & $0.0237$ $(0.0007)$ & $0.43$ \\
$1$ & $36$ & $0.9$ & $32.31$ & $1.302$ $(0.005)$ & $1.310$ $(0.049)$ & $31.01$ \\
$1$ & $1$ & $0.9$ & $0.81$ & $0.1052$ $(0.0002)$ & $0.1055$ $(0.0020)$ & $0.70$ \\
$1$ & $1$ & $0.5$ & $0.45$ & $0.1988$ $(0.0002)$ & $0.2003$ $(0.0020)$ & $0.25$ \\
\bottomrule
\end{tabular}
\end{table}

\paragraph{Other dimensions and baseline variances.}
The values in Figure~\ref{fig:ess} are Monte Carlo estimates, since the expected error has no closed form: eight batches of $4{,}000$ prompts ($2{,}000$ for the partial description), with standard errors of at most $0.09$ demonstrations. The numerical illustrations in the main text use $d=5$ and $\tau_0^2=10$, and $p=0.9$ for the unreliable description, a reliability that both sets of experiments in Section~\ref{sec:experiments} also use. Table~\ref{tab:grid} repeats the comparison of a reliable and an unreliable $c=100$ description at other values. In every case the reliable description's zero-shot ESS is within $1.6$ of $c+d+1$. At $p=0.9$, it falls to between $4.2$ and $44.4$, and it approaches the vague-prior limit $p^2d$ of Proposition~\ref{prop:unreliable} as $\tau_0^2$ grows.

\begin{table}[!htb]
\centering\small
\caption{\textbf{The cost of unreliability at other dimensions and baseline variances.} Bayes-optimal zero-shot ESS of a $c=100$ description at $p=1$ and $p=0.9$, with the limits of Propositions~\ref{prop:reliable} and~\ref{prop:unreliable}. The baseline curve is estimated from $4{,}000$ sets of demonstration inputs per count ($1{,}500$ at $d=50$). At $d=5$ and $\tau_0^2=10$, Figure~\ref{fig:ess} uses more prompts and gives $105.9$.}
\label{tab:grid}
\vspace{3pt}
\begin{tabular}{rrcccc}
\toprule
$d$ & $\tau_0^2$ & $\ess_0$, $p=1$ & $c+d+1$ & $\ess_0$, $p=0.9$ & $p^2d$ \\
\midrule
$5$ & $1$ & $105.0$ & $106$ & $8.7$ & $4.05$ \\
$5$ & $10$ & $105.8$ & $106$ & $4.7$ & $4.05$ \\
$5$ & $100$ & $105.9$ & $106$ & $4.2$ & $4.05$ \\
$16$ & $1$ & $115.8$ & $117$ & $17.3$ & $12.96$ \\
$16$ & $10$ & $116.9$ & $117$ & $13.5$ & $12.96$ \\
$16$ & $100$ & $117.0$ & $117$ & $13.0$ & $12.96$ \\
$50$ & $1$ & $149.5$ & $151$ & $44.4$ & $40.50$ \\
$50$ & $10$ & $150.9$ & $151$ & $40.9$ & $40.50$ \\
$50$ & $100$ & $151.0$ & $151$ & $40.5$ & $40.50$ \\
\bottomrule
\end{tabular}
\end{table}

\paragraph{ESS with demonstrations.}
The unreliable description's ESS in Figure~\ref{fig:ess}b does not rise steadily: after a small rise at $n=1$, from $4.73$ to $4.83$, it falls to $4.63$ at $n=3$, and it climbs steeply only as $n$ reaches $d$, to $6.7$ at $n=5$ and $18.5$ at $n=8$. With fewer demonstrations than $d$, its error is dominated by the cases in which it is irrelevant. The baseline curve falls steeply while directions of the task remain undetermined, so it matches that error within a few demonstrations.

\subsection{Partial Descriptions}\label{app:partial}

Descriptions often specify only part of a task. In our model, a reliable description may then constrain only $k$ of the $d$ coordinates of $w$; Proposition~\ref{prop:partial} gives its ESS, and the case $k=d$ is the complete description of Section~\ref{sec:specificity}. Formally, in the $k$ constrained coordinates, $m_i\sim\mathcal N(0,\tau_0^2-\tau^2)$ and $w_i\mid m\sim\mathcal N(m_i,\tau^2)$, as in Section~\ref{sec:setting}; in the remaining $d-k$ coordinates, $m_i=0$ and $w_i\sim\mathcal N(0,\tau_0^2)$. The prior precision given the description is $c=1/\tau^2$ in the constrained coordinates and $c_0=1/\tau_0^2$ elsewhere. Write
\[
\Lambda=\operatorname{diag}(cI_k,c_0I_{d-k}).
\]
The prediction error is $\E\operatorname{tr}(\Lambda+W_n)^{-1}$. Rotational invariance of $W_n$ means that the result is the same when the description constrains any $k$ orthogonal directions in place of $k$ coordinates.

\begin{proposition}[ESS of a partial description]\label{prop:partial}
Let $p=1$ and let $0\le k\le d$ be an integer.
\begin{enumerate}[nosep]
    \item[(i)] If $k<d$, then $\ess_0\to k$ as $\tau_0\to\infty$ with $\tau$ fixed.
    \item[(ii)] For fixed $d$, $\tau_0$, and $\tau$, as $n\to\infty$,
    \[
    \ess_n\longrightarrow\frac{k}{d}(c-c_0)
    =\frac{k}{d}\,sc.
    \]
    In particular, a complete description has $\ess_n\to c-c_0$.
\end{enumerate}
\end{proposition}

As the baseline prior becomes vague with $\tau$ fixed, the unconstrained coordinates dominate the zero-shot error, and ESS approaches the number of constrained coordinates, $k$. With even one unconstrained coordinate, the error with the description stays of order $\tau_0^2$, which a predictor without the description matches after about $k$ demonstrations; only a complete description brings the error down to order $\tau^2$, which takes $c+d+1$ demonstrations to match. After many demonstrations, ESS instead approaches the added precision averaged across all $d$ coordinates. When $c-c_0>d$, this large-sample ESS, $k(c-c_0)/d$, exceeds the vague-prior zero-shot ESS $k$, so for a vague enough baseline prior the description is worth more after many demonstrations than before any. Table~\ref{tab:partial} illustrates this: at $c=100$, a description of $3$ of the $d=5$ coordinates has ESS $3.2$ without demonstrations and $44.0$ with $100$ demonstrations.

\begin{proof}[Proof of Proposition~\ref{prop:partial}]
\emph{(i) No demonstrations.}
The error with the description is $\operatorname{tr}\Lambda^{-1}=(d-k)\tau_0^2+k\tau^2$. In units of $\tau_0^2$, it is
\[
d-k+\frac{k\tau^2}{\tau_0^2}\longrightarrow d-k
\qquad\text{as }\tau_0\to\infty.
\]
If $0<k<d$, this limit lies strictly between $0$ and $d$, so by the vague-baseline limit of Appendix~\ref{app:specificity} the crossing converges to $d-(d-k)=k$. If $k=0$, the description leaves the prior unchanged and its ESS is zero.

\emph{(ii) Many demonstrations.}
Applying the resolvent identity twice gives
\[
(\Lambda+W_n)^{-1}
=W_n^{-1}-W_n^{-1}\Lambda W_n^{-1}
+W_n^{-1}\Lambda W_n^{-1}\Lambda(\Lambda+W_n)^{-1}.
\]
Rotational invariance implies
\[
\E W_n^{-2}=\frac{\E\operatorname{tr}W_n^{-2}}{d}I_d.
\]
The expected trace of the middle term is therefore
$\operatorname{tr}(\Lambda)/n^2+O(n^{-3})$ by (W3). To bound the remainder, use $|\operatorname{tr}(MB)|\le\|M\|_{\mathrm{op}}\operatorname{tr}B$ for $B\succeq0$, together with $(\Lambda+W_n)^{-1}\preceq W_n^{-1}$. Its absolute trace is at most
\[
\|\Lambda\|_{\mathrm{op}}^2\|W_n^{-1}\|_{\mathrm{op}}^2\operatorname{tr}W_n^{-1}
\le d\|\Lambda\|_{\mathrm{op}}^2\operatorname{tr}W_n^{-3},
\]
whose expectation is $O(n^{-3})$. Consequently,
\[
\E\operatorname{tr}(\Lambda+W_n)^{-1}
=\frac{d}{n-d-1}-\frac{\operatorname{tr}\Lambda}{n^2}+O(n^{-3}).
\]
The baseline error is the case $\Lambda=c_0I_d$; if $k=0$ or $c=c_0$, the two errors coincide and $\ess_n=0$. Otherwise, fix an integer $T>k(c-c_0)/d$. Since $d/(n+t-d-1)=d/(n-d-1)-dt/n^2+O(n^{-3})$, uniformly over integers $0\le t\le T$,
\[
R^{\mathrm{base}}(n+t)-\E\operatorname{tr}(\Lambda+W_n)^{-1}
=\frac{d}{n^2}\left[\frac{\operatorname{tr}\Lambda-dc_0}{d}-t+O(n^{-1})\right].
\]
Linear interpolation gives the same expansion for real $t\in[0,T]$. The baseline error is strictly decreasing, so for large $n$ the left-hand side vanishes at exactly one $t$ in this range, which is $\ess_n$. Hence
\[
\ess_n=\frac{\operatorname{tr}\Lambda-dc_0}{d}+O(n^{-1})
\longrightarrow\frac{k(c-c_0)}{d}.
\]
One more demonstration and one more unit of prior precision in every direction both lower the error by $d/n^2$ to leading order, which is why a description's large-sample ESS is the precision it adds, averaged across directions.
\end{proof}

\begin{table*}[t]
\centering\small
\setlength{\tabcolsep}{3pt}
\caption{\textbf{Demonstrations can increase the value of a specific partial description.} Bayes-optimal ESS for a reliable description of $k$ of $d=5$ coordinates, with $\tau_0^2=10$ and precision $c=\tau^{-2}$. The case $k=5$ is a complete description. Each block reports ESS after $n$ demonstrations and its large-sample limit $(c-c_0)k/d$. Estimates use eight batches of $2{,}000$ prompts; standard errors are at most $0.14$.}
\label{tab:partial}
\vspace{3pt}
\begingroup
\fontsize{9}{11}\selectfont
\setlength{\tabcolsep}{4.5pt}
\renewcommand{\arraystretch}{1.08}
\begin{tabular}{@{}rrrrrrrrrrr@{}}
\toprule
& \multicolumn{5}{c}{$c=1$} & \multicolumn{5}{c}{$c=100$} \\
\cmidrule(lr){2-6}\cmidrule(lr){7-11}
$k$ & $n=0$ & $n=10$ & $n=100$ & $n=400$ & $n\to\infty$
& $n=0$ & $n=10$ & $n=100$ & $n=400$ & $n\to\infty$ \\
\midrule
1 & 0.9 & 0.4 & 0.2 & 0.2 & 0.2 & 1.0 & 2.2 & 11.4 & 16.9 & 20.0 \\
2 & 1.9 & 0.8 & 0.4 & 0.4 & 0.4 & 2.1 & 5.6 & 25.6 & 35.2 & 40.0 \\
3 & 2.9 & 1.2 & 0.6 & 0.5 & 0.5 & 3.2 & 11.8 & 44.0 & 55.1 & 59.9 \\
4 & 4.0 & 1.6 & 0.8 & 0.7 & 0.7 & 4.6 & 26.4 & 68.5 & 77.0 & 79.9 \\
5 & 5.9 & 1.9 & 1.0 & 0.9 & 0.9 & 106 & 105 & 103 & 101 & 99.9 \\
\bottomrule
\end{tabular}
\endgroup

\end{table*}

\section{Transformer Experiments}\label{app:networks}

\paragraph{Setup.}
Following \citet{garg2022}, we train from scratch on the model of Section~\ref{sec:setting}, with $\tau_0^2=10$ and $d=5$, the dimension used by \citet{huangge2025}. A prompt contains an optional description token carrying $m$, up to $20$ demonstration tokens, and a query token \citep{huangge2025}. The description token does not state $\tau$ or $p$; a network must learn these properties across training tasks. Half the prompts omit the description; a network's predictions on such prompts give its own baseline curve. The main networks are trained separately for $c\in\{1,10\}$ and $p\in\{1,0.99,0.9\}$; further networks, described later in this section, cover $c=0.2$ and $2$, $d=10$, and a partial description. The two main precisions are a factor of ten apart, as are the two nonzero probabilities of irrelevance, $1-p=0.01$ and $0.1$, so that a few networks span a wide range. We do not train at $c=100$, where the ESS would exceed the $20$ demonstrations a network sees.

\paragraph{Matched training and test reliability.}
When training and test reliability match, the networks come close to the Bayes-optimal ESS (Figure~\ref{fig:networks}a; Table~\ref{tab:networks}). The largest shortfall is at $c=10$ and $p=1$, where networks reach a mean ESS of $14.3$ against $16.1$ on the same prompts (Table~\ref{tab:networks} gives $15.9$, from a larger sample); in every other configuration, they are within $0.3$ of the Bayes-optimal ESS.

\paragraph{Reliability shifts.}
Each network learns how much to weight a description from how often descriptions were relevant in its training data, but descriptions at test time may be more or less reliable than that. We therefore evaluate each network at test reliabilities that differ from its training reliability, and compare it with two Bayesian predictors \citep{raventos2023}: one that uses the training reliability $p_{\mathrm{train}}$, and one that uses the test reliability $p_{\mathrm{test}}$ and is therefore Bayes-optimal. Because a Bayesian predictor with reliability $p$ predicts $p\,m^\top x$ at $n=0$, we summarize a network's predictions by the \emph{fitted description weight} $q$, the coefficient from regressing them on $m^\top x$.

For $c=10$, networks trained at $p=0.9$ keep description weights near $0.9$, as the $p_{\mathrm{train}}$ predictor does, even when every test description is relevant (Table~\ref{tab:networks}), and so fall short of the Bayes-optimal ESS (Figure~\ref{fig:networks}b). The opposite shift, from reliable descriptions in training to unreliable ones at test time, costs little without demonstrations (Figure~\ref{fig:networks}b). The two shifts differ because the Bayes-optimal error differs in size. When one description in ten is irrelevant, that error is already large, and a weight of $1$ instead of $0.9$ adds only about $5\%$ to it. When every description is relevant, the error is small, and a weight of $0.9$ instead of $1$ roughly doubles it. With six or more demonstrations, however, the description raises the error of networks trained at $p=1$ when they are tested at $p=0.9$, whereas networks trained at $p=0.99$ are not harmed (see \emph{Reliability shift after demonstrations are observed} below).

\paragraph{Evaluation.}
We evaluate each network on $20{,}000$ fresh prompts. The networks share these evaluation prompts, so standard deviations across training seeds exclude prompt-sampling variation. At $d = 5$, we evaluate the baseline curve at every integer from $0$ to $20$, except for $c = 0.2$ and $c = 2$, which use counts from $0$ to $15$. We do not train networks at $c=100$, the precision used in Sections~\ref{sec:specificity}--\ref{sec:reliability}: at $d=5$, its Bayes-optimal ESS, about $106$, lies far beyond the $20$ demonstrations over which a network's baseline curve is measured. We compute ESS at the first crossing of this curve with the prediction error obtained using the description, as in Definition~\ref{def:ess}. Table~\ref{tab:networks} reports results at $n = 0$, including test reliabilities that differ from the training reliability.

\begin{table}[!htb]
\centering\small
\caption{\textbf{Transformer ESS and fitted description weights under matched and shifted reliability.} Results use $d = 5$, $\tau_0^2 = 10$, and no demonstrations, with $20{,}000$ test prompts per network. Rows are grouped by training reliability. \emph{Network} reports mean ESS and standard deviation across the indicated training seeds; for a single seed, the uncertainty is the standard error over prompts. The two Bayesian predictors use $p_{\mathrm{train}}$ and $p_{\mathrm{test}}$, respectively; the latter is Bayes-optimal under the test distribution. We show one reference value when they coincide. At $n=0$, a network's prediction depends only on $m$ and $x$, whose distribution does not change with the test reliability, so its fitted weight is the same at every $p_{\mathrm{test}}$; the $p_{\mathrm{test}}=0.99$ rows for training reliabilities $1$ and $0.9$ differ only because they average one seed fewer. The values using $p_{\mathrm{test}}$, and those using $p_{\mathrm{train}}$ where the two reliabilities coincide, use $200{,}000$ prompts, with standard errors at most $0.1$. The other values using $p_{\mathrm{train}}$ use the networks' $20{,}000$ test prompts and can differ from a large-sample value by up to $0.6$, which is why some exceed the Bayes-optimal value beside them. \emph{Weight} is the fitted description weight $q$ in $\hat y \approx q\,m^\top x$, averaged across the available seeds. This coefficient is unrestricted and need not be a relevance probability.}
\label{tab:networks}
\vspace{3pt}
\begingroup
\fontsize{9}{11}\selectfont
\setlength{\tabcolsep}{4.5pt}
\renewcommand{\arraystretch}{1.08}
\begin{tabular}{@{}rrrrrrrr@{}}
\toprule
& & & & \multicolumn{3}{c}{ESS} & \\
\cmidrule(lr){5-7}
$c$ & $p_{\mathrm{train}}$ & $p_{\mathrm{test}}$ & Seeds
& Network & Bayes $p_{\mathrm{train}}$ & Bayes $p_{\mathrm{test}}$ & Weight $q$ \\
\midrule
1  & 1 & 1    & 2 & $5.85\pm0.03$  & 5.89  & ---   & 0.964 \\
1  & 1 & 0.99 & 1 & $5.65\pm0.06$  & 5.66  & 5.61  & 0.991 \\
1  & 1 & 0.9  & 2 & $4.02\pm0.03$  & 3.98  & 4.10  & 0.964 \\
10 & 1 & 1    & 3 & $14.26\pm0.48$ & 15.91 & ---   & 1.008 \\
10 & 1 & 0.99 & 2 & $9.38\pm0.16$  & 9.54  & 8.96  & 1.015 \\
10 & 1 & 0.9  & 3 & $4.58\pm0.03$  & 4.54  & 4.70  & 1.008 \\
\addlinespace[4pt]
1  & 0.99 & 1    & 2 & $5.87\pm0.01$  & 5.88  & 5.89  & 0.963 \\
1  & 0.99 & 0.99 & 2 & $5.65\pm0.01$  & 5.61  & ---   & 0.963 \\
1  & 0.99 & 0.9  & 2 & $4.02\pm0.03$  & 3.99  & 4.10  & 0.963 \\
10 & 0.99 & 1    & 3 & $13.28\pm0.57$ & 16.01 & 15.91 & 0.982 \\
10 & 0.99 & 0.99 & 3 & $9.25\pm0.15$  & 8.96  & ---   & 0.982 \\
10 & 0.99 & 0.9  & 3 & $4.60\pm0.04$  & 4.56  & 4.70  & 0.982 \\
\addlinespace[4pt]
1  & 0.9 & 1    & 2 & $5.67\pm0.06$ & 5.75  & 5.89  & 0.883 \\
1  & 0.9 & 0.99 & 1 & $5.51\pm0.06$ & 5.55  & 5.61  & 0.901 \\
1  & 0.9 & 0.9  & 2 & $4.03\pm0.01$ & 4.10  & ---   & 0.883 \\
10 & 0.9 & 1    & 3 & $9.49\pm0.44$ & 10.79 & 15.91 & 0.876 \\
10 & 0.9 & 0.99 & 2 & $8.08\pm0.28$ & 8.55  & 8.96  & 0.880 \\
10 & 0.9 & 0.9  & 3 & $4.66\pm0.01$ & 4.70  & ---   & 0.876 \\
\bottomrule
\end{tabular}
\endgroup

\end{table}

\paragraph{Architecture.}
The Transformer has $12$ layers, $8$ attention heads, and width $256$, matching the model size of \citet{garg2022}. Layers use pre-normalization, GELU activations, feed-forward width $1{,}024$, and no dropout. At $d = 5$, the model has $9.5$ million parameters.

\paragraph{Tokens.}
The description token carries $m$, each demonstration token carries an input and its output, and the prediction is read from the query token. Two indicator coordinates distinguish token types, and the model uses no positional encoding. When a description is absent, its token is masked out of attention. Outputs and description means are divided by $\tau_0$.

\paragraph{Optimization.}
Following \citet{zhu2026}, the model predicts a mean and a log variance for the query output and minimizes log loss. The optimal predicted mean is the posterior mean, which also minimizes squared error, and we evaluate the predicted mean using the loss in Equation~\eqref{eq:regret}. We train with Adam, using the PyTorch defaults $\beta_1 = 0.9$ and $\beta_2 = 0.999$ and stability constant $10^{-8}$, at a constant learning rate of $10^{-4}$ with no weight decay, for $40{,}000$ steps of $1{,}024$ newly sampled prompts and no curriculum. Gradients are clipped to norm $1$, and training uses bfloat16 autocast. The demonstration count is sampled uniformly from $0$ to $20$, except for the $c = 0.2$ and $c = 2$ experiments, which use $0$ to $15$, and the $d = 10$ experiments, which use $0$ to $40$. We train three seeds for $c = 10$, for $d = 10$, and for the partial description; two for $c = 1$; and one each for $c = 0.2$ and $c = 2$. A $d = 5$ network takes approximately $80$ minutes to train on one NVIDIA L4 GPU, and a $d = 10$ network takes approximately $2.8$ hours. Training all $25$ networks takes approximately $38$ GPU-hours.

Without a description, every $d = 5$ network's error is within $0.21$ of the Bayes-optimal error of approximately $50$ at $n = 0$, and within $0.32$ on average over its training demonstration counts. At $d = 10$, the gap is at most $0.50$ at $n=0$, where the Bayes-optimal error is approximately $100$, and $0.41$ on average over the training demonstration counts. With a description at $c = 10$, $p = 1$, and $n = 0$, network error ranges from $0.67$ to $0.71$, compared with $0.49$ for the Bayes-optimal predictor on the same prompts (its expected error is $d\tau^2=0.5$). The networks therefore remove $98.6\%$ of the baseline error, compared with $99.0\%$ for the Bayes-optimal predictor.

\paragraph{Additional specificity levels.}
Four additional networks cover $c = 0.2$ and $c = 2$, each at $p = 1$ and $p = 0.9$. Their zero-shot ESS is within $0.03$ of the Bayes-optimal value on the same prompts. Network and Bayes-optimal ESS are $2.64$ and $2.66$ at $c=0.2$ and $p=1$, $2.16$ and $2.15$ at $c=0.2$ and $p=0.9$, $7.32$ and $7.35$ at $c=2$ and $p=1$, and $4.38$ and $4.37$ at $c=2$ and $p=0.9$.

\paragraph{Higher dimension and partial descriptions.}
At $d = 10$, we train three seeds with $c = 10$ and $p=1$, and extend training and evaluation to $40$ demonstrations so that the baseline curve reaches the ESS crossing. Network ESS is $17.8$, compared with a Bayes-optimal value of $20.8$ and the limiting expression $c + d + 1 = 21$. As at $d = 5$ with $c=10$ and $p=1$, the networks capture most of the description's predictive value but fall short of the Bayes-optimal ESS.

We also train networks with a partial description, as defined in Appendix~\ref{app:partial}. The description token contains $m$ in the first $k = 3$ of the $d = 5$ coordinates and zeros elsewhere; the remaining task coordinates follow the baseline prior. For this description, at $c = 10$ and $p = 1$, mean ESS across three seeds increases from $3.27$ at $n = 0$ to $5.55$ at $n = 10$. The corresponding Bayes-optimal values are $3.22$ and $5.64$. This increase agrees with the behavior described by Proposition~\ref{prop:partial}.

\begin{figure*}[t]
\centering
\includegraphics[width=\textwidth]{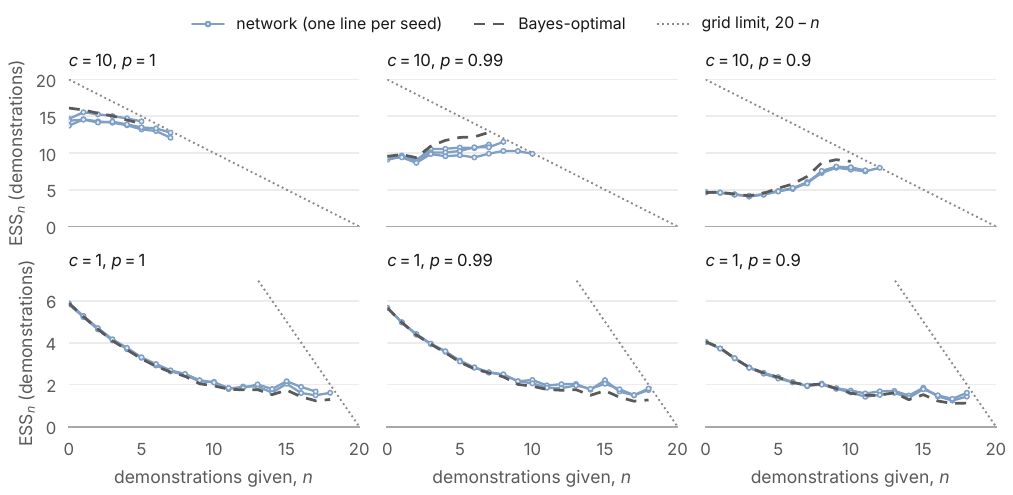}
\caption{\textbf{Transformer ESS as a function of demonstration count.} Each line represents a training seed; the Bayes-optimal predictor is evaluated on the same prompts. Training and test reliability are matched. Results cover $c = 10$ and $c = 1$ at $d = 5$, using $20{,}000$ prompts. A curve ends when its crossing exceeds the maximum of $20$ demonstrations. The dotted line, $20 - n$, marks the largest ESS identifiable on this grid.}
\label{fig:networks-ess-n}
\end{figure*}

\paragraph{ESS with demonstrations.}
Figure~\ref{fig:networks-ess-n} evaluates each network at its training reliability. For $c = 1$, network ESS remains within $0.6$ of the Bayes-optimal value at every demonstration count whose crossing lies within the evaluation grid. For $c = 10$, the largest shortfalls are $2.4$, $2.9$, and $1.4$ demonstrations at $p = 1$, $0.99$, and $0.9$, respectively; the largest excess is $0.3$. Demonstrations can raise the ESS of an unreliable description (Section~\ref{sec:reliability}): for the three $c = 10$ networks trained at $p = 0.9$, ESS rises from $4.66$ at $n=0$ to between $4.77$ and $4.83$ at $n = 5$, and the Bayes-optimal ESS on the same prompts rises from $4.63$ to $5.20$.

\paragraph{Reliability shift after demonstrations are observed.}
For $c = 10$, networks trained at $p = 1$ and $p = 0.99$ both have fitted description weights near one at $n = 0$. With demonstrations, their responses to a shift in test reliability differ. At test reliability $p = 0.9$, the description reduces error for networks trained at $p = 0.99$ at every evaluated $n$, as it does for the Bayes-optimal predictor under the test distribution. For networks trained at $p = 1$, the description instead increases error from six demonstrations onward in every seed. The excess error ranges from $0.77$ to $1.06$ at $n = 6$ and peaks between $2.74$ and $2.94$ at $n = 9$. A Bayesian predictor that assumes relevance ($p = 1$) also first has higher error with the description than without it at $n = 6$, with excess error $1.12$ there and $2.96$ at $n = 9$; this predictor returns the relevant component's prediction, $\mu_1^\top x$, whatever the demonstrations show. These networks behave as if every description were relevant.

\section{Additional Language-Model Results and Evaluation Details}\label{app:llm}

To measure behavior without reasoning more precisely and on more models, we run six open-weight models without reasoning on $2{,}000$ tasks; we refer to these experiments as the \emph{$2{,}000$-task run}. The models are gpt-oss-20b, Qwen3.5 9B, and Gemma~4 31B, which also appear in Section~\ref{sec:llm}, together with Phi-4 (14B; \citealp{abdin2024}), OLMo~2 32B Instruct \citep{walsh2025}, and Gemma~4 12B. The main numerical experiments in Section~\ref{sec:llm} use the first $200$ of these tasks. The message-sorting experiment uses a separate set of $200$ tasks, described in Appendix~\ref{app:semantic}. Appendices~\ref{app:prompts}--\ref{app:wording} and the complete tables at the end use the $2{,}000$-task run; the other subsections use the $200$-task runs of Section~\ref{sec:llm}.

\begin{figure*}[t]
\centering
\includegraphics[width=\textwidth]{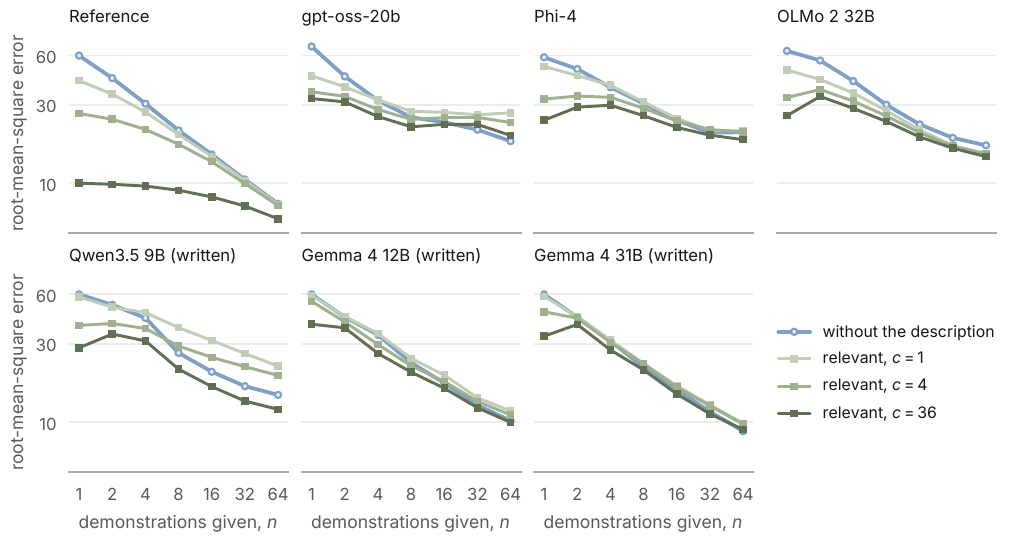}
\caption{\textbf{Learning curves used to estimate language-model ESS.} Root-mean-square error versus demonstration count for the reference and six models in the $2{,}000$-task run without reasoning, at $\sigma = 60$. Blue curves omit the description; green curves use relevant descriptions, with darker shades indicating higher $c$. Replies without a number are excluded. ESS is the horizontal distance from the point with the description at $n$ to the baseline curve at the same error. For Phi-4, OLMo~2 32B, Qwen3.5 9B, and Gemma~4 31B, error with $c = 36$ increases between one and two demonstrations because the models shift toward the demonstrations more than the reference does; Table~\ref{tab:llm-full} shows this for the first two.}
\label{fig:curves}
\end{figure*}

Table~\ref{tab:llm} reports median ESS over $n \in \{1,2,4\}$, with $95\%$ bootstrap intervals. Figure~\ref{fig:curves} shows the underlying learning curves. Tables~\ref{tab:llm-worth} and~\ref{tab:llm-worth-written} report ESS at every evaluated demonstration count; Table~\ref{tab:sensitivity} compares interpolation methods; and Table~\ref{tab:llm-full} reports description weights for the models with a distribution readout (Appendix~\ref{app:prompts}).

For the numerical experiments, the \emph{Bayesian reference} combines the probabilistic statements in the prompt with an assumed baseline prior when the prompt does not specify one. It is optimal under that assumed model, but need not minimize error under the actual task distribution when the two differ. We therefore distinguish this reference from the correctly specified Bayesian predictor in the theory. The message-sorting experiment specifies a finite task distribution and uses its exact Bayesian predictor.

In the numerical task, $c=\sigma^2/\tau^2$ is the prior precision expressed in units of one demonstration. In one condition, the precision used to generate a description differs from the one stated in the prompt: descriptions generated at $c=36$ are stated at $c=4$ (Section~\ref{sec:llm}); there we name both. Similarly, a stated reliability supplies a prior relevance probability to the model; it need not equal the proportion of relevant descriptions in the evaluation sample.

\subsection{Prompts, Tasks, and Answer Extraction}\label{app:prompts}

\paragraph{Model inclusion.}
We retain a model only if its numeric response rate, defined below for the readout we use, is at least three quarters overall and, averaged over $n=1$, $2$, and $4$ and both noise levels ($\sigma=30$ and $60$; see \emph{Task generation}), at least one half without a description, since those replies determine the baseline curve used for median ESS. This criterion excludes MiniCPM5~2B \citep{minicpm5}, OLMo~2 7B Instruct, and OLMo~2 13B Instruct, whose replies begin with a number in $62\%$, $20\%$, and $10\%$ of cases, respectively. These models often begin with explanations such as ``To predict\ldots'' or ``Given\ldots'' despite the numeric-only instruction.

The $2{,}000$-task run uses Hugging Face Transformers at each model's native precision, with checkpoints \path{openai/gpt-oss-20b}, \path{microsoft/phi-4}, \path{allenai/OLMo-2-0325-32B-Instruct}, \path{Qwen/Qwen3.5-9B}, \path{google/gemma-4-12B-it}, and \path{google/gemma-4-31B-it}. Tasks are generated with a fixed random seed, so every model sees the same tasks. Reasoning is disabled where the chat template supports this option. gpt-oss-20b writes its reasoning in a separate channel before its answer; we start generation directly in its final channel, which holds the answer.

\paragraph{Distribution and written readouts.}
For gpt-oss-20b, Phi-4, and OLMo~2 32B without reasoning, the prediction is the mean of the first-token distribution, renormalized over the $900$ three-digit number tokens. We call this the \emph{distribution readout}. It is possible for these three models because their tokenizers encode each number from $100$ to $999$ as a single token; the other models' tokenizers split numbers into digits. Other models, and all reasoning runs, use the number in the final answer, called the \emph{written readout}. We also evaluate written readouts for the three distribution-readout models as a robustness check (Table~\ref{tab:llm} for gpt-oss-20b, and Table~\ref{tab:paraphrase}). OLMo~2 32B assigns $41\%$ of first-token probability to numbers without a description at $n = 1$, and $75\%$ at $n = 2$. Its distribution readout at $n = 1$ therefore summarizes a minority of the first-token probability mass.

\begin{table}[!htb]
\centering\small
\setlength{\tabcolsep}{2.4pt}
\caption{\textbf{Description ESS and weights in the $2{,}000$-task run without reasoning.} Left: median ESS over $n = 1$, $2$, and $4$, for relevant descriptions with $c = 1$, $4$, and $36$, and for the $c = 36$ description when irrelevant or given a stated reliability of $p=0.9$ or $p=0.5$. The stated-reliability columns come from the open-alternative design of Appendix~\ref{app:llm-reliability}, in which every description is relevant and the prompt says only that the mean is otherwise unknown; they are not the mixed-reliability design of Section~\ref{sec:llm-stated}. Brackets give $95\%$ bootstrap intervals. The marker $\leq$ shows that a median is only an upper bound, because at one of its three counts a single demonstration already gives lower error than the description (Appendix~\ref{app:ess-est}); such a median is rounded up so that the printed bound holds, and so can be $0.1$ above the same value as rounded in Tables~\ref{tab:llm-worth} and~\ref{tab:llm-worth-written}. Right: the coefficient on the prior mean at $n = 1$; the Bayesian reference weight is $c/(c+n)$. All results use $\sigma = 60$. ``Written'' denotes the written readout.}
\label{tab:llm}
\vspace{3pt}
\begin{tabular}{lcccccccccc}
\toprule
& \multicolumn{6}{c}{ESS, median over $n = 1$, $2$, $4$} & & \multicolumn{3}{c}{weight, $n = 1$} \\
\cmidrule(lr){2-7} \cmidrule(lr){9-11}
& \multicolumn{3}{c}{relevant} & \multicolumn{3}{c}{$c = 36$} & & \multicolumn{3}{c}{relevant} \\
\cmidrule(lr){2-4} \cmidrule(lr){5-7} \cmidrule(lr){9-11}
Model & $c = 1$ & $c = 4$ & $c = 36$ & irrelevant & stated $p = 0.9$ & stated $p = 0.5$ & & $c = 1$ & $c = 4$ & $c = 36$ \\
\midrule
Reference & 1.3 & 4.4 & 36 & $\leq$ $-$1.0 & 22 & 5.8 & & 0.50 & 0.80 & 0.97 \\
\midrule
gpt-oss-20b & 0.9 {\scriptsize[0.7, 1.0]} & 2.0 {\scriptsize[1.7, 2.3]} & 2.8 {\scriptsize[2.6, 3.1]} & $\leq$ $-$0.4 & 2.4 {\scriptsize[1.9, 2.6]} & 2.3 {\scriptsize[1.9, 2.7]} & & 0.53 & 0.48 & 0.49 \\
Phi-4 & 0.7 {\scriptsize[0.4, 0.9]} & 4.1 {\scriptsize[3.5, 4.6]} & 7.6 {\scriptsize[6.5, 8.6]} & $\leq$ $-$1.0 & 8.7 {\scriptsize[7.8, 9.8]} & 9.7 {\scriptsize[8.7, 11]} & & 0.82 & 0.76 & 0.74 \\
OLMo 2 32B & 1.9 {\scriptsize[1.7, 2.1]} & 3.4 {\scriptsize[3.2, 3.9]} & 5.3 {\scriptsize[4.5, 6.3]} & $\leq$ $-$0.6 & 6.0 {\scriptsize[5.6, 7.1]} & 5.4 {\scriptsize[5.0, 6.0]} & & 0.72 & 0.70 & 0.70 \\
gpt-oss-20b (written) & 0.2 {\scriptsize[0.1, 0.4]} & 2.0 {\scriptsize[1.9, 2.6]} & 3.7 {\scriptsize[3.0, 4.4]} & $\leq$ $-$1.0 & 2.4 {\scriptsize[2.1, 2.6]} & 2.3 {\scriptsize[2.1, 2.6]} & & 0.64 & 0.52 & 0.52 \\
Qwen3.5 9B (written) & 0.2 {\scriptsize[$-$0.1, 0.4]} & 2.7 {\scriptsize[2.1, 3.2]} & 4.0 {\scriptsize[3.5, 4.5]} & $\leq$ $-$1.0 & 0.6 {\scriptsize[0.3, 1.0]} & 1.0 {\scriptsize[0.6, 1.5]} & & 0.81 & 0.76 & 0.80 \\
Gemma 4 12B (written) & 0.0 {\scriptsize[$-$0.1, 0.1]} & 0.5 {\scriptsize[0.3, 0.7]} & 1.8 {\scriptsize[1.4, 2.2]} & $\leq$ $-$0.7 & 1.3 {\scriptsize[1.1, 1.6]} & 2.7 {\scriptsize[2.5, 3.1]} & & 0.34 & 0.24 & 0.59 \\
Gemma 4 31B (written) & 0.0 {\scriptsize[0.0, 0.0]} & 0.2 {\scriptsize[0.1, 0.3]} & 1.4 {\scriptsize[1.2, 1.6]} & $\leq$ $-$0.7 & 0.6 {\scriptsize[0.5, 0.7]} & 0.5 {\scriptsize[0.4, 0.6]} & & 0.48 & 0.49 & 0.72 \\
\bottomrule
\end{tabular}

\end{table}

\paragraph{Prompt.}
The following text is supplied as the user turn in the model's chat template:
\begin{quote}\scriptsize\ttfamily\raggedright
The numbers below are drawn from a normal distribution with standard deviation 60.\\
Its mean was drawn from a normal distribution with mean 510 and standard deviation 30.\\[3pt]
548\\
431\\[3pt]
Predict the next number. Reply with the number only.
\end{quote}

Without a description, the second sentence is ``Its mean is unknown.'' To state unreliability, a third sentence is added. In the mixed-reliability design of Section~\ref{sec:llm-stated} (Appendix~\ref{app:llm-reliability}), it reads ``There is a 10\% chance that this statement is wrong, in which case the mean was instead drawn uniformly at random from the whole numbers 100 to 999.'' In the open-alternative design of Appendix~\ref{app:llm-reliability}, its final clause is ``in which case the mean is unknown.'' With no demonstrations, the final question is ``Predict the first number.''

\paragraph{Task generation.}
A relevant description's prior mean is sampled conditional on the task so that the joint distribution satisfies $m \sim \mathcal{N}(500,100^2-\tau^2)$ and $w\mid m \sim \mathcal{N}(m,\tau^2)$. Demonstrations are rounded to integers, and the prompt shows them without inputs, like the coin flips that \citet{gupta2025} use as in-context examples. Language models predict short unsigned numbers best \citep{gruver2023,requeima2024}, so we keep every number to three digits: a task is redrawn if a displayed number or prior mean falls outside $100$ to $999$, which occurs for less than $1\%$ of tasks. The $2{,}000$-task run also includes $\sigma = 30$ conditions. Both standard deviations stated in the prompt are then halved, preserving $c \in \{1,4,36\}$. The qualitative pattern is unchanged (Table~\ref{tab:llm-worth}).

\paragraph{Numeric response rates.}
For the distribution readout, the numeric response rate is the probability mass assigned to number tokens at the first position. For the written readout, it is the fraction of replies containing a three-digit number. Tables~\ref{tab:paraphrase}, \ref{tab:llm-worth}, \ref{tab:llm-worth-written}, and~\ref{tab:llm-full} report this quantity as ``number''. Tables~\ref{tab:paraphrase} and~\ref{tab:llm-full} italicize cells whose rate is below three quarters.

\paragraph{Written-readout implementation and checks.}
The written readout extracts the number in the last \verb|\boxed{}| expression in the final answer, if present. Otherwise, it takes the last number with three digits before an optional decimal part, retaining the decimals. A digit group after a thousands separator, as in $14{,}945$, does not count as a qualifying number. In the $2{,}000$-task run, decoding is greedy with a limit of eight new tokens, so the reply is the number itself; for these replies, the readout takes the first qualifying number and keeps only its integer part. Replies without a qualifying number are excluded from both error curves and weight regressions; the reported response rates identify the extent of this exclusion.

For gpt-oss-20b, written-readout weights differ from distribution-readout weights by at most $0.21$ across all conditions and demonstration counts. For $c = 1$, $4$, and $36$, written-readout median ESS is $0.2$, $2.0$, and $3.7$, compared with $0.9$, $2.0$, and $2.8$ from the distribution readout. A single greedy reply is a noisier summary than the distribution mean. Phi-4 produces a number in only $58\%$ of written replies across all conditions, so Table~\ref{tab:llm} reports only its distribution readout. In the wording experiment of Appendix~\ref{app:wording}, however, its numeric response rate exceeds three quarters for the original wording, so Table~\ref{tab:paraphrase} includes its written replies. OLMo~2 32B produces a number in only $5\%$, $36\%$, and $48\%$ of written replies without a description at $n = 1$, $2$, and $4$, respectively, with $\sigma = 60$; we therefore do not report its written-readout ESS. Its written-readout weights for the three relevant descriptions at $n = 1$ are $0.83$, $0.81$, and $0.83$, compared with distribution-readout weights of $0.72$, $0.70$, and $0.70$.

\subsection{ESS Estimation and the Bayesian Reference}\label{app:ess-est}

\paragraph{Measurement.}
We apply Definition~\ref{def:ess} to squared error against the latent mean. In the $2{,}000$-task run, the baseline curve is evaluated at $n = 1$, $2$, $4$, $\ldots$, $64$. Because this grid doubles $n$, we interpolate squared error linearly in $\log n$, which is less biased here than interpolation in $n$ (see \emph{Sensitivity to interpolation} below). The ESS at $n$ is the crossing $n^*$, the first interpolated demonstration count whose baseline error matches the error with the description, minus $n$. For a decreasing baseline curve, a negative ESS indicates higher error with the description than without it at the same demonstration count. For the Bayes-optimal predictor under the correctly specified model, conditioning on a description cannot increase expected error, so its ESS is nonnegative.

If the error with the description lies below the entire baseline curve, which ends at $t_{\max}=64$, we report $\ess_n > t_{\max}-n$. Because the curve starts at one demonstration, we use the marker $\ess_n \leq 1-n$ when one demonstration alone already gives lower error. These markers record censoring at the evaluation boundaries; they do not define an extrapolated matching count or an exact ESS when the crossing does not exist. Median ESS uses $n = 1$, $2$, and $4$. We exclude $n = 0$ because models rarely return a number without a description or demonstrations. At large $n$, ESS is strongly affected by slow baseline learning and excessive weight on the $c = 1$ and $c = 4$ descriptions. At $\sigma = 60$ and $n = 64$, model root-mean-square error without a description ranges from $9$ to $20$, compared with $7.5$ for the reference. A model with a flatter baseline curve can assign a larger ESS to the same error reduction. Description-weight estimates complement ESS because they do not use that curve. We obtain them by regressing predictions on $m$, $\bar y$, and an intercept: the coefficient on $m$ is the \emph{fitted description weight}, and the reference weight is $c/(c+n)$, which ranges from $0.2$ to $0.97$ over the conditions we evaluate.

Bootstrap replicates resample tasks jointly for the curves with and without the description and recompute the complete ESS statistic. The intervals therefore measure variation across tasks; they do not include interpolation bias, quantified below. We use $500$ replicates with a fixed seed and the same resampled tasks for every model.

\paragraph{The reference's baseline prior and its exact ESS.}
Because every input is the same, the $d+1$ term of Section~\ref{sec:specificity}, which arises from random inputs, does not arise here: in the idealized continuous Gaussian model with a flat baseline prior, a correctly specified description has exact ESS $c$, its conventional prior ESS \citep{neuenschwander2020}. The Bayesian reference takes ``unknown'' to mean uniform over $100$ to $999$. Without a description, its prediction is therefore approximately the sample mean, with negligible boundary truncation for the evaluated prompts. For a relevant Gaussian description with correctly stated precision, the idealized continuous model with a flat baseline prior gives exact ESS $c$. Integer rounding, finite-range truncation, and the task-rejection rule are not part of that exact identity; we compute the reported reference values on the same generated tasks as the models. When the prompt states a reliability $p$, the reference uses it as the prior relevance probability and computes the posterior relevance probability from the demonstrations. Otherwise, it treats the description as relevant.

Applying the empirical ESS procedure to the reference on the $2{,}000$-task run gives $1.3$, $4.4$, and $36$ for $c = 1$, $4$, and $36$ at $\sigma = 60$, and $1.2$, $4.7$, and $43$ at $\sigma = 30$. Interpolation in $\log n$ induces an upward bias, while finite-task sampling can shift the estimate in either direction. We use the same estimator for the reference and the models. On the $200$ tasks of Section~\ref{sec:llm}, the grid additionally includes $n = 40$ and $48$ near the $c = 36$ crossing, and the reference values are $1.0$, $3.5$, and $34.6$. In the open-alternative design of Appendix~\ref{app:llm-reliability}, in which every description is relevant and the prompt says only that the mean is otherwise unknown, the reference's median ESS varies more across task samples: on $20{,}000$ tasks it is $7.3$ at a stated reliability of $p=0.5$ and $23.4$ at $p=0.9$, compared with $5.8$ and $22$ on all $2{,}000$ tasks and $5.7$ and $27.5$ on the first $200$. Each model is compared with the reference on the same tasks.

The actual task prior, $\mathcal{N}(500,100^2)$, is not stated in the prompt. A reference using that prior would have baseline precision equivalent to $0.36$ demonstrations at $\sigma = 60$ and $0.09$ at $\sigma = 30$. At $\sigma = 60$, the corresponding idealized continuous Gaussian model would give a relevant description exact ESS $c-0.36$, namely $0.64$, $3.64$, and $35.64$. This is the scalar analogue of $c-c_0$ in Section~\ref{sec:specificity}. We use the Bayesian reference because the models are not given the actual task prior. This distinction matters under misspecification: unlike the correctly specified Bayesian predictor in Sections~\ref{sec:specificity}--\ref{sec:reliability}, a reference that incorrectly assumes an irrelevant description to be relevant can have negative ESS, indicating increased prediction error \citep{reimherr2021}.

\paragraph{Sensitivity to interpolation.}
The grid of the $2{,}000$-task run doubles $n$, so interpolation can affect ESS. On the reference's idealized errors, $\sigma^2/n$ without a description and $\sigma^2/(c+n)$ with one, interpolation in $\log n$ overestimates ESS by $0.2$, $0.3$, and $1.8$ for $c = 1$, $4$, and $36$. Linear interpolation in $n$ gives overestimates of $0.3$, $0.6$, and $4.1$.

For the models without reasoning, every crossing is unique, because every baseline curve decreases over the counts where crossings occur. Among the twelve baseline curves across the two noise levels, the only increase is Phi-4's at $\sigma = 60$, whose root-mean-square error rises slightly from $20.3$ at $n = 32$ to $20.4$ at $n = 64$. Table~\ref{tab:sensitivity} compares four estimators: interpolation in $\log n$; linear interpolation in $n$; a least-squares nonincreasing fit using the pool-adjacent-violators algorithm; and a relative-least-squares fit of $A/(n+n_0)+B$ to the baseline errors. This last form is a Bayesian learning curve plus an error floor and recovers exact ESS when fit to the reference's idealized errors.

Across these choices, no model's median ESS changes by more than $1.4$ demonstrations. For the $c = 36$ description, no model exceeds $8.1$, while the reference ranges from $35$ to $38$. Differences become larger at higher demonstration counts, where a flat learning curve turns a small error difference into a large change in ESS. For example, gpt-oss-20b at $n = 8$ and $c = 36$ has ESS $17$ under the primary log-$n$ estimator (Table~\ref{tab:llm-worth}), but $9.2$ under the fitted curve.

\begin{table*}[t]
\centering\small
\setlength{\tabcolsep}{4pt}
\caption{\textbf{Sensitivity of median ESS to baseline-curve estimation.} Median ESS over $n = 1$, $2$, and $4$ for relevant descriptions, using the $2{,}000$-task run without reasoning at $\sigma = 60$. ``Log $n$'' is the primary estimator; ``linear'' interpolates in $n$; ``monotone'' applies a nonincreasing fit; and ``fit'' uses $A/(n+n_0)+B$. The reference's idealized ESS values are $1$, $4$, and $36$.}
\label{tab:sensitivity}
\vspace{3pt}
\begin{tabular}{lrrrrrrrrrrrr}
\toprule
 & \multicolumn{4}{c}{$c = 1$} & \multicolumn{4}{c}{$c = 4$} & \multicolumn{4}{c}{$c = 36$} \\
\cmidrule(lr){2-5} \cmidrule(lr){6-9} \cmidrule(lr){10-13}
Predictor & log $n$ & linear & monotone & fit & log $n$ & linear & monotone & fit & log $n$ & linear & monotone & fit \\
\midrule
Reference & 1.3 & 1.4 & 1.3 & 1.0 & 4.4 & 4.7 & 4.4 & 4.0 & 36 & 38 & 36 & 35 \\
\midrule
gpt-oss-20b & 0.9 & 1.0 & 0.9 & 0.9 & 2.0 & 2.4 & 2.0 & 2.4 & 2.8 & 2.9 & 2.8 & 3.4 \\
Phi-4 & 0.7 & 0.8 & 0.7 & 0.4 & 4.1 & 4.4 & 4.1 & 3.6 & 7.6 & 8.1 & 7.6 & 7.1 \\
OLMo 2 32B & 1.9 & 2.1 & 1.9 & 1.4 & 3.4 & 3.7 & 3.4 & 3.3 & 5.3 & 5.8 & 5.3 & 5.5 \\
Qwen3.5 9B (written) & 0.2 & 0.3 & 0.2 & $-$0.1 & 2.7 & 2.9 & 2.7 & 1.3 & 4.0 & 4.3 & 4.0 & 2.8 \\
Gemma 4 12B (written) & 0.0 & 0.0 & 0.0 & 0.0 & 0.5 & 0.6 & 0.5 & 0.4 & 1.8 & 1.9 & 1.8 & 1.6 \\
Gemma 4 31B (written) & 0.0 & 0.0 & 0.0 & $-$0.1 & 0.2 & 0.3 & 0.2 & 0.0 & 1.4 & 1.7 & 1.4 & 1.0 \\
\bottomrule
\end{tabular}

\end{table*}

\subsection{Irrelevant Descriptions and Wording}\label{app:wording}

Apart from the stated-reliability experiments of Section~\ref{sec:llm-stated}, the experiments in the main text keep every description relevant, and all of them state its precision in one wording. This section relaxes each of these choices in the $2{,}000$-task run.

\paragraph{Irrelevant descriptions.}
An irrelevant description states a prior mean drawn independently of the task, from the same distribution as relevant prior means, with the same stated precision. At $c = 36$, these descriptions have negative ESS for every model and for the Bayesian reference, which treats them as relevant. The models assign less weight to the description and consequently incur lower error in this condition: at $n = 4$, their root-mean-square errors range from $35$ to $64$, compared with $124$ for the reference.

\paragraph{Wording without reasoning.}
Descriptions in practice express uncertainty in many ways. To test whether the results depend on the original wording, we repeat the three relevant-description conditions and the no-description baseline at $n = 1$, $2$, and $4$ using three alternative description sentences: ``Its mean is about $m$, give or take $\tau$.'', ``Mean: $m \pm \tau$'', and ``An earlier estimate put its mean at $m$, with a standard error of $\tau$.'' Table~\ref{tab:paraphrase} uses the tasks of the $2{,}000$-task run and written readouts.

Wording substantially changes the weight assigned to the description. For Gemma~4 12B with $c = 1$ and $n = 1$, this weight is $0.34$ under the original wording and $1.00$ under each alternative. For gpt-oss-20b, it ranges from $0.17$ to $0.67$. Nevertheless, none of the tested wordings yields the reference's sensitivity to prior precision: across models and wordings, the weight difference between $c = 36$ and $c = 1$ is at most $0.25$ at $n = 1$ and $0.44$ at $n = 4$, compared with reference differences of $0.47$ and $0.70$. The informal descriptions do not uniquely specify a probability distribution, so this comparison changes how explicitly the prompt specifies the prior, not only its wording.

\begin{table}[!htb]
\centering\footnotesize
\caption{\textbf{Description weights under four wordings without reasoning.} Weights are reported for the three relevant descriptions at $n = 1$ and $4$, using written readouts on the tasks of the $2{,}000$-task run. ``Number'' denotes the numeric response rate averaged over demonstration counts; a weight is italicized when the response rate at that count is below three quarters. The reference weights are $0.50$, $0.80$, and $0.97$ at $n = 1$, and $0.20$, $0.50$, and $0.90$ at $n = 4$. A fitted weight above $1$ means that, holding the sample mean fixed in the regression, a unit change in the prior mean corresponds to more than a unit change in the prediction. Wording changes overall reliance on the description, but the observed change with stated precision remains smaller than the reference change.}
\label{tab:paraphrase}
\vspace{3pt}
\begin{tabular}{llccccccc}
\toprule
& & & \multicolumn{3}{c}{weight, $n = 1$} & \multicolumn{3}{c}{weight, $n = 4$} \\
\cmidrule(lr){4-6} \cmidrule(lr){7-9}
Model & Wording of the description & number & $c = 1$ & $c = 4$ & $c = 36$ & $c = 1$ & $c = 4$ & $c = 36$ \\
\midrule
Gemma 4 12B & normal, mean $m$, sd $\tau$ (ours) & 1.00 & 0.34 & 0.24 & 0.59 & 0.19 & 0.24 & 0.31 \\
 & about $m$, give or take $\tau$ & 1.00 & 1.00 & 1.00 & 1.00 & 0.87 & 0.76 & 0.82 \\
 & Mean: $m \pm \tau$ & 1.00 & 1.00 & 1.00 & 1.00 & 0.60 & 0.66 & 0.74 \\
 & earlier estimate $m$, s.e.\ $\tau$ & 1.00 & 1.00 & 1.00 & 1.00 & 0.45 & 0.54 & 0.67 \\
\midrule
Gemma 4 31B & normal, mean $m$, sd $\tau$ (ours) & 1.00 & 0.48 & 0.49 & 0.72 & 0.02 & 0.04 & 0.18 \\
 & about $m$, give or take $\tau$ & 1.00 & 1.00 & 1.00 & 1.00 & 1.01 & 1.22 & 1.45 \\
 & Mean: $m \pm \tau$ & 1.00 & 1.00 & 1.00 & 1.00 & 0.90 & 1.03 & 1.15 \\
 & earlier estimate $m$, s.e.\ $\tau$ & 1.00 & 0.94 & 0.94 & 0.96 & 0.09 & 0.23 & 0.41 \\
\midrule
gpt-oss-20b & normal, mean $m$, sd $\tau$ (ours) & 1.00 & 0.64 & 0.52 & 0.52 & 0.23 & 0.29 & 0.34 \\
 & about $m$, give or take $\tau$ & 1.00 & 0.24 & 0.17 & 0.38 & 0.28 & 0.29 & 0.38 \\
 & Mean: $m \pm \tau$ & 1.00 & 0.17 & 0.26 & 0.40 & 0.23 & 0.24 & 0.30 \\
 & earlier estimate $m$, s.e.\ $\tau$ & 1.00 & 0.67 & 0.63 & 0.67 & 0.28 & 0.37 & 0.42 \\
\midrule
OLMo 2 32B & normal, mean $m$, sd $\tau$ (ours) & 0.61 & 0.83 & 0.81 & 0.83 & \textit{0.20} & \textit{0.29} & 0.38 \\
 & about $m$, give or take $\tau$ & 0.82 & 0.93 & 0.92 & 0.96 & 0.60 & 0.57 & 0.65 \\
 & Mean: $m \pm \tau$ & 0.82 & 0.88 & 0.83 & 0.88 & 0.44 & 0.46 & 0.55 \\
 & earlier estimate $m$, s.e.\ $\tau$ & 0.82 & 0.95 & 0.94 & 0.95 & 0.49 & 0.57 & 0.65 \\
\midrule
Phi-4 & normal, mean $m$, sd $\tau$ (ours) & 0.77 & 0.96 & 0.78 & 0.76 & \textit{0.38} & \textit{0.37} & \textit{0.38} \\
 & about $m$, give or take $\tau$ & 0.97 & 1.02 & 1.01 & 1.02 & 0.65 & 0.57 & 0.62 \\
 & Mean: $m \pm \tau$ & 0.83 & \textit{0.99} & \textit{0.87} & \textit{0.87} & 0.45 & 0.42 & 0.50 \\
 & earlier estimate $m$, s.e.\ $\tau$ & 0.81 & 0.88 & 0.76 & 0.73 & \textit{0.47} & \textit{0.49} & 0.54 \\
\midrule
Qwen3.5 9B & normal, mean $m$, sd $\tau$ (ours) & 1.00 & 0.81 & 0.76 & 0.80 & 0.26 & 0.27 & 0.44 \\
 & about $m$, give or take $\tau$ & 1.00 & 0.99 & 1.00 & 0.99 & 0.35 & 0.43 & 0.57 \\
 & Mean: $m \pm \tau$ & 1.00 & 1.08 & 1.25 & 1.14 & 0.19 & 0.27 & 0.38 \\
 & earlier estimate $m$, s.e.\ $\tau$ & 1.00 & 0.87 & 0.86 & 0.86 & 0.61 & 0.75 & 0.82 \\
\bottomrule
\end{tabular}

\end{table}

\subsection{Reasoning Settings and Inference Details}\label{app:reasoning-runs}

\paragraph{Comparisons between modes.}
The labels ``with reasoning'' and ``without reasoning'' refer to generation settings. When we say that a reply contains reasoning, we mean that the provider recorded reasoning output for it; with reasoning enabled, some replies have none. Paired runs use the same model identifier and prompts, but reasoning settings, serving software, readout, and retry policy vary across the runs. These comparisons concern the evaluated configurations and do not identify a unique internal mechanism.

\paragraph{Shared evaluation prompts.}
Most configurations in Figure~\ref{fig:llm} are evaluated on the same $7{,}800$ prompts from the first $200$ tasks. These include no-description prompts at $n=1$, $2$, $4$, $\ldots$, $64$, with additional points at $n=40$ and $48$ near the $c=36$ crossing, and every description condition at $n=1$, $2$, and $4$. The gpt-oss-20b and Olmo~3 7B Think \citep{olmo2025three} runs omit the two additional baseline points and contain $7{,}400$ prompts. Results without reasoning for the open-weight models are taken from the $2{,}000$-task run restricted to these tasks. Paired mode comparisons use the same prompts, although serving environments differ as detailed below. At least $99.7\%$ of replies in each run contain a number.

\paragraph{Open-weight reasoning models.}
With reasoning, gpt-oss-20b runs in vLLM using greedy decoding, low reasoning effort, and at most $4{,}096$ new tokens. It reaches the final channel in $99.9\%$ of replies and uses $322$ tokens on average. Qwen3.5 9B runs through OpenRouter at temperature $0$, with reasoning enabled and a limit of $24{,}576$ new tokens. We pin requests to DeepInfra, which serves the model in bfloat16, because quantization varies across providers. Of its replies, $2{,}305$ repeat themselves until they reach the token limit. Each is rerun once with the model card's recommended reasoning sampling settings: temperature $1.0$, top-$p$ $0.95$, top-$k$ $20$, and presence penalty $1.5$. Eleven remain unfinished and one request fails.

Gemma~4 31B (\texttt{gemma-4-31b-it}) runs through the Gemini API at temperature $0$. Reasoning is enabled by default, and the model uses $2{,}613$ reasoning tokens on average, with the same $24{,}576$-token limit. Of its replies, $435$ reach the limit; all finish after one rerun with the model card's recommended sampling settings: temperature $1.0$, top-$p$ $0.95$, and top-$k$ $64$.

We also run Olmo~3 7B Think (\path{allenai/Olmo-3-7B-Think}), a small open-weight model that always reasons and so has no run without reasoning. It runs in vLLM with greedy decoding and at most $32{,}768$ new tokens. We rerun the $169$ replies that reach this limit once with temperature $0.6$ and top-$p$ $0.95$; $17$ remain unfinished. The model generates $9{,}215$ tokens on average and writes a worked solution before most answers. Its long greedy replies depend on the hardware: an earlier run of the same conditions on a different GPU gave the same final number on only half the prompts. We report a single run and treat it as one sample from this variation. It averages the demonstrations about as well as the reference but places more weight on the description at $c=1$ and $4$. At $c=36$, its replies split between the prior mean alone and other prediction rules, as detailed below.

\paragraph{Models served through command-line agents.}
GPT-6 Sol (\texttt{gpt-6-sol}, the model identifier served by Codex during the runs) runs through the Codex command-line agent in a read-only sandbox, with an empty configuration directory so that no user instructions or settings load, and with every tool that can run code disabled. We evaluate the model with reasoning disabled (no reasoning tokens) and enabled at low effort, at which, in a pilot, its answers were already within $1$ of the reference's predictions.

Claude Opus~4.6 (\texttt{claude-opus-4-6}) and Haiku~5.5 (\texttt{claude-haiku-5-5}) run through the Claude Code command line, with one prompt per call, no tools, no project or user instructions, and the system prompt ``You are a helpful assistant.'' Opus~4.6 uses high effort in both evaluated modes, with reasoning either enabled or disabled, so that the two runs differ only in reasoning; the disabled mode emits no reasoning tokens. With reasoning enabled, at least $99\%$ of replies in each description condition include reasoning output, with an average of $599$ reasoning tokens across all replies. Without a description, only $1\%$, $24\%$, and $79\%$ of replies include reasoning output at $n=1$, $2$, and $4$, respectively, so its baseline curve at small $n$ mostly reflects replies without reasoning; using the reference's baseline curve instead changes its ESS by at most $1$ (see below). Haiku~5.5 is evaluated mainly at medium effort, because at low effort it often answers without reasoning; we also report low effort for comparison. At low effort and $n=1$, it often repeats the single demonstration value without reasoning output, so its description weight at $n=1$ is low even in conditions whose ESS is close to the reference value. At medium effort and $c=36$, its ESS is $3.4$ at $n=1$, $35$ at $n=2$, matching the reference, and $28$ at $n=4$, against the reference's $33$; its median ESS is therefore $28$, against $34.6$, though its bootstrap interval in Figure~\ref{fig:llm}b includes the reference's value. At $n=1$, $28\%$ of its replies contain no recorded reasoning output, against $19\%$ at $c=1$ and $20\%$ at $c=4$. Both Claude models use the command line's default sampling settings, which the command line does not report, and we collect one reply per prompt. Without reasoning, Opus~4.6 still writes out intermediate steps in $64$ of its $7{,}800$ replies; excluding these replies changes no median ESS by more than $0.1$.

\paragraph{Run dates.}
Because hosted models can change over time, we record when each was evaluated. The $2{,}000$-task run took place on 2--3 October 2026, and its conditions in which the stated precision differs from the generating one, and gpt-oss-20b's reasoning run, took place on 20 October. Haiku~5.5 was evaluated on 20 October for the runs of Section~\ref{sec:llm} and 4 November for wording, the stated baseline prior, and the message-sorting experiment. Opus~4.6, Qwen3.5 9B, Gemma~4 31B, and Olmo~3 7B Think were evaluated on 3--4 November. GPT-6 Sol was evaluated on 4 November. The irrelevant-description runs of the mixed-reliability design took place on 9--10 November, when Opus~4.6's runs with the baseline prior stated were also completed. The GPT-6 Sol logs do not record a model version beyond the identifier above.

\paragraph{Without reasoning.}
Counted in demonstrations, the shortfall from the reference is much smaller at $c=4$ and $c=1$ than at $c=36$, because the reference's own ESS is small there, but most models still reach less than three fifths of the reference's ESS (Table~\ref{tab:reason}). Most models also respond too little to the stated precision itself. When the prompt states $c=4$ for descriptions generated at $c=36$, the reference's ESS falls from $34.6$ to $13.7$ (Table~\ref{tab:reason}). It stays above its value for descriptions generated at $c=4$, because the prior mean is as accurate as before. GPT-6 Sol's ESS falls by a larger fraction than the reference's, and that of Opus~4.6 and Gemma~4 31B by about half, somewhat less than the reference's. The ESS of Qwen3.5 9B and gpt-oss-20b falls by a quarter or less, against three fifths for the reference, so these two models respond little to the stated precision.

\paragraph{ESS when a model's baseline curve stops falling.}
With reasoning, gpt-oss-20b and Qwen3.5 9B nearly match the reference description weight: at $n=4$ and $c=36$, their fitted weights are $0.89$ and $0.90$, compared with $0.90$ for the reference. Their errors without descriptions stop falling at around $32$ demonstrations, so at two or more of the three counts we evaluate, their error with a $c=36$ description lies below their entire baseline curve, and their median ESS on their own curves is not defined ($>60$ in Table~\ref{tab:reason}). gpt-oss-20b makes errors when averaging long sequences: its no-description root-mean-square error is $19.0$ at $n=32$ and $24.7$ at $n=64$, compared with reference errors $10.0$ and $7.4$. At these counts, $37$ and $62$ replies, respectively, differ from the sample mean by more than $20$. Qwen3.5 9B levels off near $11$ from $n=32$ onward ($11.2$ at $n=64$), with up to $11$ replies per cell that far from the sample mean.

For these models, the error with a description sometimes lies below the whole baseline curve, so the evaluated grid contains no crossing. Table~\ref{tab:reason} therefore also reports, in parentheses, a comparison using the reference's baseline. Specifically, it finds the reference demonstration count $t_{\mathrm{ref}}$ that matches the model's error with the description and reports $t_{\mathrm{ref}}-n$. This comparison differs from ESS on the model's own baseline curve because it uses another predictor's baseline curve. gpt-oss-20b also makes arithmetic errors with descriptions: about $2\%$ of its replies are more than $40$ from the reference prediction, from errors such as exchanging the description's and the demonstrations' precisions. These errors reduce its ESS; for example, its median ESS is $1.3$ when the prompt states a reliability of $p=0.9$ in the open-alternative design of Table~\ref{tab:reason}. At a stated $p=0.5$, by contrast, it largely ignores the stated reliability (Table~\ref{tab:hedge}), and because every description in those runs is relevant, its ESS of $8.8$ exceeds the reference's $5.7$.

Opus~4.6 and GPT-6 Sol do not exhibit these arithmetic failures. Opus~4.6 has root-mean-square error $7.4$ at $n=64$, matching the reference on these tasks. For these two models, changing between the model and reference baselines changes ESS by at most $1$. Table~\ref{tab:error} reports each run's absolute error with a description.

\begin{table*}[t]
\centering\small
\setlength{\tabcolsep}{2.5pt}
\caption{\textbf{Description ESS across reasoning settings.} Results use the first $200$ tasks at $\sigma=60$. Median ESS uses each run's own baseline curve; parentheses report the value computed using the reference baseline where it differs. ``Stated reliability'' gives the reliability that the prompt states for a description generated at $c=36$, in the open-alternative design of Appendix~\ref{app:llm-reliability}: every description is in fact relevant, and the prompt says only that the mean is otherwise unknown. These columns therefore differ from the mixed-reliability design of Section~\ref{sec:llm-stated}, whose results are in Table~\ref{tab:matched}. The last column fixes the generating precision at $36$ and states it as $4$. Each run names the mode and, in parentheses, the provider's effort setting. An upper bound, $\leq$, indicates that one demonstration alone gives lower error; $>60$ means that the error with the description lies below the model's entire baseline curve, through $n=64$, so no crossing exists on the grid. Reference values use the same tasks and estimator. The open-weight rows without reasoning take these $200$ tasks from the $2{,}000$-task run, whose results on all its tasks are in Table~\ref{tab:llm}. The runs also include three further conditions in which the stated precision differs from the generating one, not shown.}
\label{tab:reason}
\vspace{3pt}
\begingroup
\fontsize{9}{11}\selectfont
\setlength{\tabcolsep}{3pt}
\renewcommand{\arraystretch}{1.08}
\resizebox{\textwidth}{!}{%
\begin{tabular}{@{}lrrrrrrr@{}}
\toprule
& & & & & \multicolumn{2}{c}{Stated reliability} & \\
\cmidrule(lr){6-7}
Run & $c=1$ & $c=4$ & $c=36$ & Irrelevant
& $p=0.9$ & $p=0.5$ & \shortstack{Stated $c=4$\\Generated $c=36$} \\
\midrule
Reference & $1.0$ & $3.5$ & $34.6$ & $\le-1.0$ & $27.5$ & $5.7$ & $13.7$ \\
GPT-6 Sol, without reasoning & $0.9$ & $1.9$ & $10$ & $\le-1.0$ & $3.3$ & $0.9$ & $2.4$ \\
GPT-6 Sol, with reasoning (low) & $0.9$ & $3.3$ & $35\ (34)$ & $\le-1.0$ & $30\ (29)$ & $4.2$ & $11$ \\
Opus 4.6, without reasoning & $0.9$ & $2.0$ & $5.8\ (5.9)$ & $\le-1.0$ & $6.1\ (6.0)$ & $3.8$ & $3.0$ \\
Opus 4.6, with reasoning (high) & $1.0$ & $3.5$ & $35$ & $\le-1.0$ & $29$ & $3.8$ & $13$ \\
Gemma 4 31B, without reasoning & $0.0\ (-0.2)$ & $0.5\ (0.2)$ & $2.0\ (1.6)$ & $\le-0.3$ & $1.0\ (0.7)$ & $0.8\ (0.4)$ & $1.0\ (0.6)$ \\
Gemma 4 31B, with reasoning & $1.0$ & $3.5$ & $35$ & $\le-1.0$ & $21$ & $4.0$ & $14$ \\
Qwen3.5 9B, without reasoning & $-0.2\ (-0.8)$ & $2.7\ (0.4)$ & $5.5\ (1.1)$ & $\le-1.0$ & $1.6\ (-0.7)$ & $2.3\ (-0.6)$ & $4.1\ (1.1)$ \\
Qwen3.5 9B, with reasoning & $0.9$ & $3.5$ & $>60\ (34)$ & $\le-1.0$ & $29\ (27)$ & $3.9$ & $15\ (14)$ \\
gpt-oss-20b, without reasoning & $0.5\ (0.3)$ & $1.8\ (1.3)$ & $2.7\ (1.7)$ & $\le-0.5$ & $1.8\ (1.5)$ & $1.7\ (1.5)$ & $2.5\ (1.6)$ \\
gpt-oss-20b, with reasoning (low) & $0.3$ & $1.0$ & $>60\ (24)$ & $\le-1.0$ & $1.3\ (1.2)$ & $8.8$ & $8.6\ (8.4)$ \\
Haiku 5.5, with reasoning (medium) & $1.1$ & $3.3\ (3.5)$ & $28$ & $\le-1.0$ & $33$ & $3.8\ (3.9)$ & $8.5\ (8.8)$ \\
Haiku 5.5, with reasoning (low) & $0.8\ (0.7)$ & $3.2\ (3.3)$ & $24$ & $\le-1.0$ & $27\ (26)$ & $5.5\ (5.6)$ & $8.6\ (8.4)$ \\
Olmo 3 7B Think, with reasoning & $-0.5$ & $0.0$ & $6.2\ (6.0)$ & $\le-1.0$ & $5.9\ (5.6)$ & $5.5\ (5.3)$ & $15\ (13)$ \\
\bottomrule
\end{tabular}}
\endgroup

\end{table*}

\begin{table}[!htb]
\centering\small
\setlength{\tabcolsep}{4pt}
\caption{\textbf{Absolute error with a description.} Root-mean-square error against the latent mean, for descriptions generated and stated at $c=4$ and $c=36$, on the first $200$ tasks at $\sigma=60$. Rows are the runs of Table~\ref{tab:reason}, whose ESS compares each run with its own baseline curve; this table reports the error itself. Replies without a number are excluded.}
\label{tab:error}
\vspace{3pt}
\begin{tabular}{@{}lrrrrrr@{}}
\toprule
& \multicolumn{3}{c}{$c=4$} & \multicolumn{3}{c}{$c=36$} \\
\cmidrule(lr){2-4} \cmidrule(lr){5-7}
Run & $n=1$ & $2$ & $4$ & $1$ & $2$ & $4$ \\
\midrule
Reference & $26.7$ & $24.0$ & $19.9$ & $9.5$ & $9.3$ & $9.3$ \\
GPT-6 Sol, without reasoning & $32.1$ & $28.6$ & $24.2$ & $15.2$ & $16.3$ & $18.4$ \\
GPT-6 Sol, with reasoning (low) & $26.7$ & $26.7$ & $20.1$ & $9.5$ & $9.4$ & $9.6$ \\
Opus 4.6, without reasoning & $29.4$ & $28.1$ & $23.7$ & $13.6$ & $18.9$ & $18.1$ \\
Opus 4.6, with reasoning (high) & $26.7$ & $24.0$ & $19.8$ & $9.5$ & $9.3$ & $9.3$ \\
Gemma 4 31B, without reasoning & $43.4$ & $41.6$ & $27.6$ & $32.6$ & $39.1$ & $24.0$ \\
Gemma 4 31B, with reasoning & $26.7$ & $24.0$ & $19.9$ & $9.5$ & $9.3$ & $9.3$ \\
Qwen3.5 9B, without reasoning & $38.0$ & $38.7$ & $34.5$ & $28.5$ & $33.6$ & $26.0$ \\
Qwen3.5 9B, with reasoning & $26.7$ & $24.0$ & $19.9$ & $9.6$ & $9.3$ & $9.3$ \\
gpt-oss-20b, without reasoning & $34.5$ & $32.6$ & $26.5$ & $31.5$ & $30.5$ & $23.7$ \\
gpt-oss-20b, with reasoning (low) & $28.0$ & $34.5$ & $52.4$ & $11.9$ & $32.2$ & $11.0$ \\
Haiku 5.5, with reasoning (medium) & $33.7$ & $24.0$ & $19.8$ & $27.0$ & $9.3$ & $10.1$ \\
Haiku 5.5, with reasoning (low) & $46.1$ & $24.0$ & $20.2$ & $41.9$ & $9.7$ & $11.1$ \\
Olmo 3 7B Think, with reasoning & $28.7$ & $42.1$ & $32.1$ & $20.8$ & $16.5$ & $28.8$ \\
\bottomrule
\end{tabular}

\end{table}

\subsection{Prediction Rules in Individual Replies}\label{app:rules}

\paragraph{Per-reply description weights.}
At $n=4$, we compute the per-reply description weight $(\hat w-\bar y)/(m-\bar y)$. This quantity describes one reply and differs from the fitted coefficient obtained by regressing predictions across tasks. We use the tasks with $|m-\bar y|\geq30$, where the weight is well determined: $61$ tasks at $c=36$, whose more accurate prior means lie closer to the sample mean, and $124$ at $c=1$. We exclude replies without a number. A reply is classified as consistent with a rule when its per-reply weight is within $0.05$ of that rule's weight. The candidate weights are the Bayesian reference ($0.90$ for $c=36$ and $0.20$ for $c=1$); the reference with the description's and the demonstrations' precisions exchanged, $n/(c+n)$ ($0.10$ and $0.80$); the description counted as one additional demonstration, $1/(n+1)$; the description alone, with weight $1$; and the demonstrations alone, with weight $0$. The one-additional-demonstration rule coincides with the reference at $c=1$.

With reasoning, Opus~4.6, GPT-6 Sol, Gemma~4 31B, and Haiku~5.5 match the reference rule on $89$ to $100\%$ of eligible replies. Without reasoning, $77\%$ of Opus~4.6 replies at $c=36$ match none of these candidate rules, as do more than half of GPT-6 Sol replies. Enabling reasoning does not ensure agreement with the reference in every model. Olmo~3 7B Think reaches a median ESS of only $6.2$ at $c=36$, although it averages demonstrations well: without a description, its root-mean-square error falls from $56$ at $n=1$ to $10.8$ at $n=32$, compared with $10.0$ for the reference. Its shortfall lies in combining the description with the demonstrations. At $n=1$, it repeats the prior mean in $57$ to $80\%$ of replies, more often at higher stated precision, and at $n=4$, about half of its eligible replies return the prior mean alone and a third match the reference.

\subsection{Stated Reliability}\label{app:llm-reliability}

We test stated reliability in two designs. In both, the prompt adds a sentence, the \emph{warning}, that states the description's reliability $p$. In the \emph{mixed-reliability design} of Section~\ref{sec:llm-stated}, descriptions are irrelevant with probability $1-p$ and the warning states what holds in that case. In the \emph{open-alternative design}, the warning leaves the irrelevant case open, saying only that the mean is then unknown; in the results of Tables~\ref{tab:llm} and~\ref{tab:reason}, every description in this design is relevant; we report it second, because it shows what models do when the prompt does not determine how to update.

\paragraph{Mixed-reliability design.}
The prompt adds the sentence ``There is a 10\% chance that this statement is wrong, in which case the mean was instead drawn uniformly at random from the whole numbers 100 to 999,'' or the same sentence with 50\%. For cost reasons, we run this design on the three models that our subscriptions cover, GPT-6 Sol, Opus~4.6, and Haiku~5.5. Each of the $200$ tasks is evaluated with its relevant $c=36$ description and with an irrelevant one, whose mean is drawn independently of the task from the same marginal distribution. At stated reliability $p$, the error with the description is $p$ times the mean squared error with the relevant description plus $1-p$ times that with the irrelevant one, so that the stated reliability is the true one. The stated alternative is the reference's baseline prior, not the distribution that generates the tasks; the prompt states it because without it the posterior relevance probability is not determined. ESS is then computed as in Section~\ref{sec:llm}, against each model's own baseline curve, with intervals that resample tasks. \emph{Fixed-weight averaging} conditions each component on the demonstrations and averages the two posterior means with the prior weights $p$ and $1-p$; a predictor that ignores the warning treats every description as relevant.

Table~\ref{tab:matched} reports the results. At $p=0.9$, the reference's median ESS is $5.0$, against $34.6$ without the warning, and the two simpler predictors have ESS $0.3$ and $-0.1$: the description's remaining value depends on recognizing, from the demonstrations, the tasks on which it is irrelevant. At $p=0.5$, the reference's ESS is $0.9$, and both simpler predictors do worse with the description than without it. GPT-6 Sol with reasoning is within $0.2$ of the reference at both reliabilities, and about nine in ten of its replies are within $2$ of the reference's prediction with either kind of description. Opus~4.6 with reasoning is within $0.1$ of the reference at both reliabilities, with $96$ to $100\%$ of its replies within $2$ of the reference's prediction. Haiku~5.5 falls short, at $p=0.9$ mainly with irrelevant descriptions and at $p=0.5$ with both kinds. Without reasoning, GPT-6 Sol and Opus~4.6 gain little or nothing from the description.

\begin{table}[!htb]
\centering\small
\setlength{\tabcolsep}{3.5pt}
\caption{\textbf{Unreliable descriptions with stated reliability.} Median ESS in the mixed-reliability design, with $95\%$ bootstrap intervals over tasks, and the share of replies within $2$ of the reference's prediction when the description is relevant (rel.) and irrelevant (irr.), pooled over $n=1$, $2$, and $4$. A value of $\le-1.0$ is censored: at $n=2$ the description does worse than one demonstration alone.}
\label{tab:matched}
\vspace{3pt}
\begin{tabular}{lcccccc}
\toprule
& \multicolumn{3}{c}{$p=0.9$} & \multicolumn{3}{c}{$p=0.5$} \\
\cmidrule(lr){2-4} \cmidrule(lr){5-7}
& ESS & rel. & irr. & ESS & rel. & irr. \\
\midrule
Reference & $5.0\ [3.9,6.2]$ & & & $0.9\ [0.5,1.2]$ & & \\
Fixed weights & $0.3\ [-0.2,1.0]$ & & & $-0.7\ [-1.0,-0.4]$ & & \\
Ignores the warning & $-0.1\ [-0.6,0.6]$ & & & $\le-1.0$ & & \\
\midrule
GPT-6 Sol, low effort & $4.8\ [3.8,6.3]$ & $89\%$ & $92\%$ & $0.8\ [0.5,1.2]$ & $90\%$ & $92\%$ \\
Opus~4.6, high effort & $5.0\ [3.9,6.9]$ & $100\%$ & $96\%$ & $0.9\ [0.5,1.3]$ & $98\%$ & $96\%$ \\
Haiku~5.5, medium effort & $3.2\ [2.4,4.1]$ & $85\%$ & $60\%$ & $0.6\ [0.3,1.0]$ & $58\%$ & $52\%$ \\
GPT-6 Sol, without reasoning & $0.9\ [0.6,1.2]$ & $19\%$ & $12\%$ & $-0.1\ [-0.2,0.1]$ & $10\%$ & $12\%$ \\
Opus~4.6, without reasoning & $1.0\ [0.5,1.6]$ & $15\%$ & $11\%$ & $0.0\ [0.0,0.0]$ & $14\%$ & $13\%$ \\
\bottomrule
\end{tabular}
\end{table}

\paragraph{When the prompt leaves the irrelevant case open.}
In the open-alternative design, the warning stated a reliability of $p=0.9$ or $p=0.5$ and said that otherwise ``the mean is unknown,'' and the descriptions themselves remained relevant. A stated reliability of $p=0.5$ then reduces reference ESS from $34.6$ to $5.7$ under the uniform baseline prior (Table~\ref{tab:reason}). Without reasoning, the ESS of Opus~4.6, gpt-oss-20b, Qwen3.5 9B, and Gemma~4 31B falls by about a third to three fifths, against $84\%$ for the reference, and GPT-6 Sol's falls from $10.3$ to $0.9$.

\paragraph{Fixed-weight averaging.}
The Bayesian reference averages its two component predictions with the posterior relevance probability $\pi_n$, which the demonstrations move away from the stated $p$. For $c=36$, we compare it with fixed-weight averaging, defined above, which keeps the prior weights $p$ and $1-p$. Treating the uniform baseline prior as flat, fixed-weight averaging gives weight $p\,36/(36+n)$ to $m$. Table~\ref{tab:hedge} reports fitted description weights under the open-alternative design's warning, which leaves the baseline prior unspecified. At the stated $p=0.5$, GPT-6 Sol, Gemma~4 31B, and Qwen3.5 9B with reasoning are close to fixed-weight averaging. Opus~4.6 starts near the reference at $n=1$ but falls below fixed-weight averaging by $n=4$, and Haiku~5.5 at medium effort lies between the two at $n=1$ and $2$. Haiku~5.5 at low effort comes closest to the reference, although its replies often contain no reasoning output at $n=1$. The individual replies show the same pattern. Pooling $n=1$, $2$, and $4$ at the stated $p=0.5$, the share of replies within $2$ of the fixed-weight prediction is $100\%$ for Gemma~4 31B, $97\%$ for Qwen3.5 9B, $89\%$ for GPT-6 Sol, $80\%$ for Opus~4.6, $58\%$ for Haiku~5.5 at medium effort, and $21\%$ for gpt-oss-20b. The share within $2$ of the reference prediction is at most $24\%$ for every model. Regressing the replies' normalized description weights (per-reply weights divided by $c/(c+n)$, defined below) on the reference's $\pi_n$ gives slopes of $0.01$ for GPT-6 Sol, $0.02$ for Gemma~4 31B, $0.06$ for Qwen3.5 9B, $0.22$ for Haiku~5.5, $0.30$ for Opus~4.6, and $0.00$ for gpt-oss-20b, where updating as the reference does gives $1$.

\begin{table}[!htb]
\centering\small
\setlength{\tabcolsep}{3pt}
\caption{\textbf{Description weights under stated unreliability.} Fitted coefficients on the $c=36$ prior mean for the first $200$ tasks, with stated reliability $p$. All model rows use reasoning. ``Fixed weights'' uses weight $p\,36/(36+n)$; the reference updates the relevance probability using a uniform baseline prior over $100$ to $999$. Bootstrap standard errors are at most $0.03$ at $n=4$ at the stated $p=0.5$ and at most $0.06$ overall, except for gpt-oss-20b, whose maximum is $0.09$. At the stated $p=0.5$, several reasoning configurations closely track fixed-weight averaging.}
\label{tab:hedge}
\vspace{3pt}
\begin{tabular}{lcccccc}
\toprule
& \multicolumn{3}{c}{Stated $p=0.5$} & \multicolumn{3}{c}{Stated $p=0.9$} \\
\cmidrule(lr){2-4} \cmidrule(lr){5-7}
& $n=1$ & $2$ & $4$ & $1$ & $2$ & $4$ \\
\midrule
Reference & $0.58$ & $0.61$ & $0.64$ & $0.88$ & $0.87$ & $0.84$ \\
Fixed weights &$0.49$ & $0.47$ & $0.45$ & $0.88$ & $0.85$ & $0.81$ \\
\midrule
GPT-6 Sol & $0.49$ & $0.46$ & $0.40$ & $0.90$ & $0.86$ & $0.71$ \\
Opus~4.6 & $0.56$ & $0.43$ & $0.40$ & $0.93$ & $0.86$ & $0.77$ \\
Gemma~4 31B & $0.49$ & $0.46$ & $0.45$ & $0.89$ & $0.68$ & $0.76$ \\
Qwen3.5 9B & $0.49$ & $0.46$ & $0.43$ & $0.88$ & $0.84$ & $0.80$ \\
Haiku~5.5, medium & $0.55$ & $0.56$ & $0.46$ & $0.79$ & $0.93$ & $0.89$ \\
Haiku~5.5, low & $0.49$ & $0.62$ & $0.58$ & $0.42$ & $0.94$ & $0.88$ \\
gpt-oss-20b & $0.89$ & $0.87$ & $0.82$ & $0.95$ & $0.87$ & $0.90$ \\
\bottomrule
\end{tabular}
\end{table}

\paragraph{Stating the baseline prior.}
Computing $\pi_n$ requires the marginal likelihood of the demonstrations under each component, and hence the baseline prior under irrelevance. The open-alternative design's warning says only that the mean is otherwise ``unknown,'' so the prompt does not determine a unique posterior relevance probability: a model must supply the prior from its pretrained knowledge, and the reference must assume one. The assumption matters: with its baseline curve held fixed, the reference's median ESS at the stated $p=0.5$ is $5.7$ under our uniform prior over $100$ to $999$, $4.6$ under the generating distribution $\mathcal{N}(500,100^2)$, and $13.4$ under a uniform prior over $0$ to $2000$. To remove this ambiguity, we rerun the stated-reliability condition on the same $200$ tasks, at $c=36$, stated reliabilities of $p=0.9$ and $p=0.5$, and $n=1$, $2$, and $4$. The prompt now states the reference's uniform baseline prior over $100$ to $999$. We use GPT-6 Sol at low effort, Haiku~5.5 at medium effort, and Opus~4.6 at high effort, with $1{,}200$ prompts each; these runs also supply the relevant-description half of the mixed-reliability design above. Opus~4.6 without reasoning answers the same prompts, which supply the relevant half of its results in the mixed-reliability design. Table~\ref{tab:fallback} compares their fitted description weights with those from the open-alternative design.

At the stated $p=0.5$, the reference and fixed-weight predictions differ by at least $5$ on $123$ to $155$ tasks per cell. On these tasks, when the baseline prior is stated, the share of replies closer to the reference than to the fixed-weight prediction rises from $0$--$5\%$ to $92$--$97\%$ for GPT-6 Sol, from $21$--$41\%$ to $87$--$95\%$ for Haiku~5.5, and from $4$--$32\%$ to $100\%$ for Opus~4.6. At the stated $p=0.9$, at most $7$ tasks per cell separate the two rules by this amount, so the comparison is uninformative there. Pooling the three values of $n$ at the stated $p=0.5$, the share of replies within $2$ of the reference prediction rises from $10\%$ to $90\%$ for GPT-6 Sol, from $23\%$ to $58\%$ for Haiku~5.5, and from $15\%$ to $98\%$ for Opus~4.6.

A constant increase in the description weight could also make replies closer to the reference, because its posterior relevance probability exceeds the stated $p=0.5$ on most tasks. To test whether the weights vary with the evidence, we use each reply's \emph{normalized description weight}: its per-reply weight divided by $c/(c+n)$, restricting to tasks with $|m-\bar y|\geq30$. For the scalar reference, this equals its posterior relevance probability $\pi_n$, up to the negligible boundary effects discussed above. For model predictions, the ratio can lie outside $[0,1]$ and does not identify an internal probability. Here $\pi_n$ ranges from $0.08$ to $0.88$ as the demonstrations provide different evidence for relevance; fixed-weight averaging keeps the normalized weight constant.

We regress the models' normalized weights on $\pi_n$, pooling $n=1$, $2$, and $4$ at the stated $p=0.5$ ($269$ replies per model). With the baseline prior stated, the slopes are $0.96$ [$0.84$, $1.04$] for GPT-6 Sol, $0.98$ [$0.82$, $1.15$] for Haiku~5.5, and $0.99$ [$0.97$, $1.00$] for Opus~4.6, compared with $0.01$ [$-0.09$, $0.11$], $0.22$ [$-0.01$, $0.45$], and $0.30$ [$0.14$, $0.43$] under the open-alternative design's warning. The $95\%$ intervals resample the $153$ tasks. The corresponding intercepts with the prior stated are $0.01$, $0.03$, and $0.01$, and $R^2$ is $0.57$, $0.33$, and $0.99$.

The agreement also holds when the demonstrations count against the description. Among the $23$ replies with $\pi_n$ below $0.5$ (mean $0.33$), the mean normalized weights are $0.30$, $0.38$, and $0.34$ for the three models with the baseline prior stated, compared with $0.48$, $0.50$, and $0.38$ under the open-alternative design's warning. The three models therefore vary their description weights across tasks in approximately the amount predicted by Bayesian model averaging.

Median ESS uses each model's baseline curve from the runs of Section~\ref{sec:reasoning}. At the stated $p=0.5$, fixed-weight averaging gives $4.4$ and the Bayesian reference gives $5.7$. Stating the baseline prior raises ESS from $4.2$ to $5.4$ for GPT-6 Sol, from $3.8$ to $4.2$ for Haiku~5.5, and from $3.8$ to $5.7$ for Opus~4.6. At the stated $p=0.9$, it changes from $29.7$ to $25.0$ for GPT-6 Sol, from $32.6$ to $15.1$ for Haiku~5.5, and from $28.8$ to $27.4$ for Opus~4.6, compared with $27.5$ for the reference. GPT-6 Sol's and Haiku~5.5's fitted weights agree within $0.01$ in this condition (Table~\ref{tab:fallback}). Haiku's lower ESS therefore does not reflect lower average reliance on the description; it comes from a minority of replies far from the reference, some of which repeat a demonstration value. We also ran two further configurations with the baseline prior stated. Haiku~5.5 at low effort also updates (slope $0.93$), with median ESS $2.7$ at the stated $p=0.5$, against $5.5$ under the open-alternative design's warning. Opus~4.6 without reasoning does not (slope $-0.18$), with median ESS $3.8$ under both warnings.

\begin{table}[!htb]
\centering\small
\setlength{\tabcolsep}{3pt}
\caption{\textbf{Stating the baseline prior changes description weights.} Fitted coefficients on the $c=36$ prior mean for the first $200$ tasks, as in Table~\ref{tab:hedge}. ``Unstated'' denotes the open-alternative design's warning, ``the mean is unknown''; ``stated'' specifies that the mean is otherwise uniform over $100$ to $999$, as the reference assumes. GPT-6 Sol uses low effort, Haiku~5.5 medium effort, and Opus~4.6 high effort.}
\label{tab:fallback}
\vspace{3pt}
\begin{tabular}{lcccccc}
\toprule
& \multicolumn{3}{c}{Stated $p=0.5$} & \multicolumn{3}{c}{Stated $p=0.9$} \\
\cmidrule(lr){2-4} \cmidrule(lr){5-7}
& $n=1$ & $2$ & $4$ & $1$ & $2$ & $4$ \\
\midrule
Reference & $0.58$ & $0.61$ & $0.64$ & $0.88$ & $0.87$ & $0.84$ \\
Fixed weights &$0.49$ & $0.47$ & $0.45$ & $0.88$ & $0.85$ & $0.81$ \\
GPT-6 Sol, unstated & $0.49$ & $0.46$ & $0.40$ & $0.90$ & $0.86$ & $0.71$ \\
GPT-6 Sol, stated & $0.58$ & $0.59$ & $0.59$ & $0.88$ & $0.83$ & $0.80$ \\
Haiku~5.5, unstated & $0.55$ & $0.56$ & $0.46$ & $0.79$ & $0.93$ & $0.89$ \\
Haiku~5.5, stated & $0.62$ & $0.60$ & $0.63$ & $0.89$ & $0.84$ & $0.80$ \\
Opus~4.6, unstated & $0.56$ & $0.43$ & $0.40$ & $0.93$ & $0.86$ & $0.77$ \\
Opus~4.6, stated & $0.58$ & $0.61$ & $0.64$ & $0.88$ & $0.87$ & $0.83$ \\
\bottomrule
\end{tabular}
\end{table}

\subsection{Wording with Reasoning Enabled}\label{app:wording-reasoning}

Because reasoning brings description weights close to the reference under the original wording, we test whether it does so under the alternative wordings as well. We evaluate the three alternative wordings with Haiku~5.5 at medium effort, for the three relevant descriptions at $n = 1$, $2$, and $4$ on the first $200$ tasks; all $5{,}400$ replies contain a number. With ``An earlier estimate put its mean at $m$, with a standard error of $\tau$'', at least $71\%$ of replies in each cell include reasoning output; $99\%$ of these lie within $2$ of the reference prediction, the tolerance of Table~\ref{tab:matched}. Under ``Its mean is about $m$, give or take $\tau$'' and ``Mean: $m \pm \tau$'', reasoning output is absent from $4$ to $36\%$ and $2$ to $74\%$ of replies per cell, respectively. Even among replies with reasoning output, $71\%$ and $91\%$ repeat the prior mean, and $43\%$ and $32\%$ lie within $2$ of the reference prediction; the two groups overlap, because repeating the prior mean can lie within $2$ of the reference when the description is specific. At $c = 1$ and $n = 4$, its description weights are $0.72$ and $0.93$, compared with reference weight $0.20$.

All wordings provide a value $m$ and a scale $\tau$, but ``give or take'' and ``$\pm$'' do not identify $\tau$ as a prior standard deviation or specify a distribution for the mean. The comparison varies the explicitness of the probabilistic specification as well as its wording. A model that combines a description with demonstrations correctly when the prompt states a distribution therefore need not do so when the prompt does not.

\subsection{Descriptions in Words: Sorting Messages}\label{app:semantic}

The numerical descriptions state a distribution over the task, whereas descriptions in practice usually state constraints in words that the model must interpret. We therefore design a message-sorting experiment with such descriptions that still admits an exact Bayes-optimal predictor. As in Bayesian models of rule-based concept learning \citep{goodman2008}, the latent task is one of a finite set of rules, here $60$ sorting rules under a uniform prior. A note constrains the rule in words: a broad note leaves $20$ compatible rules, and a specific note leaves $2$. With reasoning, Haiku~5.5 and GPT-6 Sol obtain ESS $9.9$ and $9.5$ for the specific note at $n=2$, compared with $11.4$ for the reference, and when the note may be irrelevant, most of their predictions are closer to the Bayesian reference than to fixed-weight averaging.

\paragraph{Task.}
A message has six binary properties: its sender is outside the organization; it requests a payment; it concerns travel; it requests computer access; it has a deadline within a day; and it contains personal contact information. Each message states every property as present or absent, using one of three phrasings for each value, such as ``I'm a contractor from another company.'' or ``I'm on the sales team here.'' Properties are sampled independently with probability one half, phrasings are sampled uniformly, and sentences appear in random order.

A rule uses exactly two properties and requires each to be present or absent. A message goes into folder A when both requirements hold, and into B otherwise. There are $\binom{6}{2}2^2=60$ such rules. The prompt describes this family, calling the rule the company's filing policy, and states that it was chosen uniformly from the family. It then provides a note, $n$ previously sorted messages with their folders, and four new query messages. The model is asked for each new message's probability of going into A, as a number between $0$ and $1$, on four lines from ``1: \textless probability\textgreater'' to ``4: \textless probability\textgreater''.

\paragraph{Notes.}
A broad note reads, for example, ``Policy note: whether it concerns travel is one of the two properties the policy uses,'' leaving $20$ possible rules. A specific note reads ``Policy note: to go into A, a message must request a payment. The other property the policy uses is whether it concerns travel,'' leaving $2$. These notes state constraints in words; the model must determine which rules satisfy them.

The unreliable note is a specific note with an added sentence: ``This note has a 50\% chance of being wrong, in which case it says nothing about the policy, which is then any of the 60, equally likely.'' In the generator, the note is relevant with probability one half and otherwise describes an independently drawn rule. The note is irrelevant in $92$ of the $200$ tasks. An independently drawn note can agree with the actual rule by chance, so we distinguish \emph{relevance}, the latent component that generated the note, from \emph{compatibility}, whether a rule satisfies it.

\paragraph{Reference and ESS.}
For a note asserted to be relevant, the reference predicts the fraction of rules compatible with the note and consistent with the sorted messages that assign the new message to A. For an unreliable note, it averages this prediction with the prediction from rules consistent with the sorted messages alone, weighting the two by the posterior relevance probability. The prior relevance probability is $0.5$; the posterior probability is obtained using the likelihood of the demonstrations under each component. As a check on our implementation, on $3{,}000$ simulated tasks the reference's predictive probabilities are calibrated in every condition, and its posterior probability that the unreliable note is relevant matches the frequency of relevance: where it exceeds $0.7$, it averages $0.80$, and the note is relevant in $80\%$ of cases.

The loss is the Brier score, the squared error between the predicted probability of A and the true folder indicator, averaged over the four query messages. Without demonstrations, a broad note leaves the prediction unchanged: for any message, exactly a quarter of the $20$ rules compatible with the note send it to A, the same fraction as among all $60$, so the broad note's ESS at $n=0$ is zero. ESS follows Definition~\ref{def:ess}, using each model's baseline curve, its loss without a note, at $n=0$, $2$, $4$, $6$, $8$, $12$, $16$, $20$, $24$, and $32$, with linear interpolation in $n$. On simulated tasks, this grid changes the reference ESS by at most $0.6$ relative to evaluation at every integer demonstration count.

\paragraph{Runs.}
We use $200$ tasks, each defined by a sampled rule, and $19$ conditions: no note at the ten demonstration counts above, and each note at $n=0$, $2$, and $4$. This gives $3{,}800$ prompts per configuration. Haiku~5.5 runs at medium effort; GPT-6 Sol runs without reasoning and at low effort. We leave out Opus~4.6, because in a pilot it still wrote out its working on this task with reasoning disabled, so its two settings would not differ only in reasoning. The command-line agents are set up as in Appendix~\ref{app:reasoning-runs}, with no tools, and no reply calls a tool. To check that GPT-6 Sol without reasoning reads the messages correctly, we asked it to state each property of $96$ messages; it was correct in $97$ to $100\%$ of cases for each property; all three errors concern one phrasing, ``I'm not traveling for this.'' Fourteen replies from GPT-6 Sol at low effort contain no probability and are excluded; one Haiku~5.5 reply omits one of its four probabilities, and we score its other three. Every Haiku~5.5 reply includes reasoning output, with a median of $3{,}008$ reasoning tokens. For GPT-6 Sol at low effort, $91\%$ of replies include reasoning output, with a median of $142$ reasoning tokens.

\paragraph{Results.}
Table~\ref{tab:semantic} reports each note's ESS. With reasoning, both models are within $2$ demonstrations of the reference in every condition except GPT-6 Sol's specific note at $n=4$: its ESS is $26.1$, compared with $14.8$ for the reference. Its loss with this note, $0.008$, is close to the reference's $0.007$, but its baseline curve flattens beyond $n=16$, with loss $0.022$ at $n=20$, compared with $0.004$ for the reference. It consequently needs more demonstrations on its own baseline curve to match the note's error reduction.

For the unreliable note, we compare predictions with the Bayesian reference and with fixed-weight averaging, which averages the two component predictions using their prior weights. Here it retains the stated weight $p=0.5$ unless the sorted messages rule out every rule compatible with the note; in that case, it assigns zero weight to the relevant component. We use query messages on which the two predictions differ by at least $0.05$: about $150$ at $n=2$ and $200$ at $n=4$, out of $800$. The share of predictions closer to the reference is $90\%$ at $n=2$ and $99\%$ at $n=4$ for Haiku~5.5; $77\%$ and $87\%$ for GPT-6 Sol at low effort; and $45\%$ at both counts for GPT-6 Sol without reasoning.

Without reasoning, GPT-6 Sol learns slowly from demonstrations. Its loss without a note is still $0.078$ at $n=32$, compared with less than $0.0005$ for the reference. Measured against this weaker baseline, a note can have an ESS as large as the reference's or larger. For example, the specific note has ESS $13.5$ at $n=2$, although the model's loss with that note is $0.130$, compared with $0.027$ for the reference. This illustrates why ESS on a model's own baseline curve must be interpreted alongside absolute prediction error.

\begin{table}[!htb]
\centering\small
\setlength{\tabcolsep}{4pt}
\caption{\textbf{ESS of notes in the message-sorting experiment.} Results use $200$ tasks and each model's own baseline curve, measured without a note. ``$\leq0$'' means that the note did not lower loss; ``$>28$'' means that the baseline curve does not reach the loss with the note by $n=32$. On $6{,}000$ simulated tasks, the reference ESS is $9.6$ and $13.2$ for the specific note at $n=2$ and $4$, and $2.5$ and $2.9$ for the unreliable note. Models are compared with the reference on the same $200$ tasks. Paired $95\%$ bootstrap intervals over tasks ($1{,}000$ replicates) for the specific note at $n=2$ are: reference [$9.7$, $12.9$], Haiku~5.5 [$8.7$, $11.5$], GPT-6 Sol at low effort [$7.8$, $12.7$], and GPT-6 Sol without reasoning [$3.7$, $21.9$].}
\label{tab:semantic}
\vspace{3pt}
\begin{tabular}{llrrrr}
\toprule
Note & $n$ & Reference & \shortstack{Haiku 5.5\\medium} & \shortstack{GPT-6 Sol\\low} & \shortstack{GPT-6 Sol\\no reasoning} \\
\midrule
Broad & 0 & 0.0 & 0.0 & $\leq0$ & $\leq0$ \\
& 2 & 1.4 & 1.3 & 1.3 & 0.7 \\
& 4 & 1.8 & 1.6 & 1.7 & 8.6 \\
\addlinespace
Specific & 0 & 3.7 & 3.6 & 3.3 & 3.7 \\
& 2 & 11.4 & 9.9 & 9.5 & 13.5 \\
& 4 & 14.8 & 14.5 & 26.1 & $>28$ \\
\addlinespace
Unreliable & 0 & 1.3 & 1.3 & 1.6 & 1.7 \\
& 2 & 3.0 & 2.6 & 3.2 & 1.7 \\
& 4 & 5.0 & 4.9 & 5.1 & 2.7 \\
\bottomrule
\end{tabular}
\end{table}

\subsection{Complete Tables}
Tables~\ref{tab:llm-worth}--\ref{tab:llm-full} report ESS and description weights at every evaluated demonstration count for the $2{,}000$-task run without reasoning.

\begin{table}[p]
\centering\scriptsize
\setlength{\tabcolsep}{3pt}
\caption{\textbf{ESS by description condition and demonstration count: distribution readout.} Results cover the Bayesian reference and the three distribution-readout models in the $2{,}000$-task run without reasoning. ``Number'' is the numeric response rate averaged over $n$. ``Stated $p$'' is the reliability that the prompt states. The censoring marker $\leq$ marks error above the first baseline point; $>64-n$ means that the error with the description lies below the entire baseline curve through $64$ demonstrations.}
\label{tab:llm-worth}
\vspace{3pt}
\begin{tabular}{llrrrrrrrr}
\toprule
Predictor & Description & number & $n = 1$ & $n = 2$ & $n = 4$ & $n = 8$ & $n = 16$ & $n = 32$ & $n = 64$ \\
\midrule
Reference & $\sigma = 60$, $c = 1$, relevant &  & 1.2 & 1.3 & 1.3 & 1.3 & 1.2 & 1.5 & $> 0$ \\
 & $\sigma = 60$, $c = 1$, irrelevant &  & $\leq$ 0.0 & $-$0.5 & $-$0.8 & $-$1.2 & $-$1.7 & $-$1.5 & $-$1.7 \\
 & $\sigma = 60$, $c = 4$, relevant &  & 4.5 & 4.4 & 3.9 & 4.6 & 4.6 & 5.2 & $> 0$ \\
 & $\sigma = 60$, $c = 4$, irrelevant &  & $\leq$ 0.0 & $\leq$ $-$1.0 & $\leq$ $-$3.0 & $-$6.2 & $-$12 & $-$20 & $-$32 \\
 & $\sigma = 60$, $c = 36$, relevant &  & 36 & 36 & 37 & 38 & 39 & $> 32$ & $> 0$ \\
 & $\sigma = 60$, $c = 36$, irrelevant &  & $\leq$ 0.0 & $\leq$ $-$1.0 & $\leq$ $-$3.0 & $\leq$ $-$7.0 & $\leq$ $-$15 & $\leq$ $-$31 & $-$62 \\
 & $\sigma = 30$, $c = 1$, relevant &  & 1.0 & 1.2 & 1.4 & 1.6 & 1.4 & 1.5 & $> 0$ \\
 & $\sigma = 30$, $c = 4$, relevant &  & 4.5 & 4.7 & 5.0 & 5.1 & 4.9 & 5.6 & $> 0$ \\
 & $\sigma = 30$, $c = 36$, relevant &  & 42 & 43 & 43 & 43 & 42 & $> 32$ & $> 0$ \\
 & $\sigma = 60$, $c = 4$, relevant, stated $p = 0.9$ &  & 3.8 & 3.9 & 3.6 & 4.4 & 4.5 & 5.1 & $> 0$ \\
 & $\sigma = 60$, $c = 4$, irrelevant, stated $p = 0.9$ &  & $\leq$ 0.0 & $-$0.7 & $-$1.2 & $-$1.6 & $-$2.2 & $-$1.9 & $-$2.2 \\
 & $\sigma = 60$, $c = 36$, relevant, stated $p = 0.9$ &  & 13 & 22 & 26 & 35 & 38 & $> 32$ & $> 0$ \\
 & $\sigma = 60$, $c = 36$, irrelevant, stated $p = 0.9$ &  & $\leq$ 0.0 & $-$0.9 & $-$1.7 & $-$2.8 & $-$5.4 & $-$6.9 & $-$10 \\
 & $\sigma = 60$, $c = 4$, relevant, stated $p = 0.5$ &  & 1.9 & 2.2 & 2.9 & 3.7 & 4.1 & 4.5 & $> 0$ \\
 & $\sigma = 60$, $c = 4$, irrelevant, stated $p = 0.5$ &  & $\leq$ 0.0 & $-$0.2 & $-$0.4 & $-$0.5 & $-$0.8 & $-$0.6 & $-$0.7 \\
 & $\sigma = 60$, $c = 36$, relevant, stated $p = 0.5$ &  & 3.1 & 5.8 & 12 & 21 & 31 & $> 32$ & $> 0$ \\
 & $\sigma = 60$, $c = 36$, irrelevant, stated $p = 0.5$ &  & $\leq$ 0.0 & $-$0.3 & $-$0.7 & $-$1.2 & $-$2.4 & $-$2.8 & $-$4.5 \\
\midrule
gpt-oss-20b & $\sigma = 60$, $c = 1$, relevant & 0.98 & 1.0 & 0.9 & $-$0.1 & $-$1.5 & $-$9.2 & $-$25 & $-$57 \\
 & $\sigma = 60$, $c = 1$, irrelevant & 0.98 & $\leq$ 0.0 & $-$0.3 & $-$1.5 & $-$3.6 & $-$10 & $-$25 & $-$59 \\
 & $\sigma = 60$, $c = 4$, relevant & 0.98 & 2.3 & 1.7 & 2.0 & 2.9 & $-$7.5 & $-$24 & $-$48 \\
 & $\sigma = 60$, $c = 4$, irrelevant & 0.97 & $\leq$ 0.0 & $-$0.3 & $-$1.4 & $-$3.7 & $-$11 & $-$26 & $-$59 \\
 & $\sigma = 60$, $c = 36$, relevant & 0.98 & 2.8 & 2.3 & 3.8 & 17 & 3.7 & $-$12 & $-$17 \\
 & $\sigma = 60$, $c = 36$, irrelevant & 0.98 & $\leq$ 0.0 & $-$0.5 & $-$1.5 & $-$3.7 & $-$11 & $-$26 & $-$60 \\
 & $\sigma = 30$, $c = 1$, relevant & 0.99 & 0.8 & 0.4 & $-$0.3 & $-$1.8 & $-$9.9 & $-$26 & $-$59 \\
 & $\sigma = 30$, $c = 4$, relevant & 0.98 & 1.2 & 1.3 & 1.9 & 3.4 & $-$8.1 & $-$24 & $-$44 \\
 & $\sigma = 30$, $c = 36$, relevant & 0.99 & 1.6 & 2.3 & 4.7 & 31 & 3.0 & $-$13 & $> 0$ \\
 & $\sigma = 60$, $c = 4$, relevant, stated $p = 0.9$ & 0.96 & 2.2 & 1.8 & 0.0 & $-$0.2 & $-$8.1 & $-$24 & $-$52 \\
 & $\sigma = 60$, $c = 4$, irrelevant, stated $p = 0.9$ & 0.96 & $\leq$ 0.0 & $-$0.6 & $-$1.1 & $-$2.2 & $-$9.6 & $-$25 & $-$58 \\
 & $\sigma = 60$, $c = 36$, relevant, stated $p = 0.9$ & 0.97 & 2.4 & 2.4 & 0.5 & 2.8 & $-$3.6 & $-$20 & $-$33 \\
 & $\sigma = 60$, $c = 36$, irrelevant, stated $p = 0.9$ & 0.96 & $\leq$ 0.0 & $-$0.7 & $-$1.1 & $-$2.3 & $-$9.7 & $-$25 & $-$59 \\
 & $\sigma = 60$, $c = 4$, relevant, stated $p = 0.5$ & 0.95 & 2.3 & 1.7 & 0.1 & $-$0.5 & $-$8.3 & $-$24 & $-$53 \\
 & $\sigma = 60$, $c = 4$, irrelevant, stated $p = 0.5$ & 0.94 & $\leq$ 0.0 & $-$0.5 & $-$1.1 & $-$2.3 & $-$9.7 & $-$25 & $-$58 \\
 & $\sigma = 60$, $c = 36$, relevant, stated $p = 0.5$ & 0.95 & 2.8 & 2.3 & 0.7 & 1.2 & $-$4.5 & $-$21 & $-$37 \\
 & $\sigma = 60$, $c = 36$, irrelevant, stated $p = 0.5$ & 0.95 & $\leq$ 0.0 & $-$0.6 & $-$1.1 & $-$2.3 & $-$9.7 & $-$25 & $-$59 \\
\midrule
Phi-4 & $\sigma = 60$, $c = 1$, relevant & 0.87 & 0.8 & 0.7 & $-$0.2 & $-$0.4 & $-$1.0 & $-$3.1 & $-$35 \\
 & $\sigma = 60$, $c = 1$, irrelevant & 0.86 & $\leq$ 0.0 & $\leq$ $-$1.0 & $\leq$ $-$3.0 & $-$4.9 & $-$6.1 & $-$8.8 & $-$40 \\
 & $\sigma = 60$, $c = 4$, relevant & 0.85 & 5.8 & 4.1 & 2.4 & 2.1 & 0.7 & $-$4.6 & $-$33 \\
 & $\sigma = 60$, $c = 4$, irrelevant & 0.84 & $\leq$ 0.0 & $\leq$ $-$1.0 & $\leq$ $-$3.0 & $-$4.7 & $-$6.2 & $-$13 & $-$42 \\
 & $\sigma = 60$, $c = 36$, relevant & 0.86 & 15 & 7.6 & 4.7 & 5.6 & 8.5 & $> 32$ & $> 0$ \\
 & $\sigma = 60$, $c = 36$, irrelevant & 0.84 & $\leq$ 0.0 & $\leq$ $-$1.0 & $\leq$ $-$3.0 & $-$4.5 & $-$6.4 & $-$12 & $-$42 \\
 & $\sigma = 30$, $c = 1$, relevant & 0.87 & 1.1 & 0.8 & $-$0.3 & $-$0.3 & $-$1.0 & $-$1.8 & $-$15 \\
 & $\sigma = 30$, $c = 4$, relevant & 0.86 & 4.8 & 3.5 & 2.0 & 1.6 & 0.6 & $-$1.9 & $-$2.2 \\
 & $\sigma = 30$, $c = 36$, relevant & 0.88 & 9.7 & 6.0 & 3.8 & 4.4 & 8.4 & 17 & $> 0$ \\
 & $\sigma = 60$, $c = 4$, relevant, stated $p = 0.9$ & 0.87 & 6.0 & 5.5 & 5.3 & 8.3 & 15 & $> 32$ & $> 0$ \\
 & $\sigma = 60$, $c = 4$, irrelevant, stated $p = 0.9$ & 0.85 & $\leq$ 0.0 & $\leq$ $-$1.0 & $-$2.1 & $-$1.2 & 2.9 & $-$3.2 & $> 0$ \\
 & $\sigma = 60$, $c = 36$, relevant, stated $p = 0.9$ & 0.87 & 10 & 8.7 & 8.2 & 17 & $> 48$ & $> 32$ & $> 0$ \\
 & $\sigma = 60$, $c = 36$, irrelevant, stated $p = 0.9$ & 0.86 & $\leq$ 0.0 & $\leq$ $-$1.0 & $-$2.4 & $-$1.7 & 1.3 & $-$4.7 & $> 0$ \\
 & $\sigma = 60$, $c = 4$, relevant, stated $p = 0.5$ & 0.87 & 6.0 & 6.4 & 7.7 & 17 & $> 48$ & $> 32$ & $> 0$ \\
 & $\sigma = 60$, $c = 4$, irrelevant, stated $p = 0.5$ & 0.85 & $\leq$ 0.0 & $\leq$ $-$1.0 & $-$1.3 & 1.5 & 16 & $> 32$ & $> 0$ \\
 & $\sigma = 60$, $c = 36$, relevant, stated $p = 0.5$ & 0.88 & 9.2 & 9.7 & 11 & $> 56$ & $> 48$ & $> 32$ & $> 0$ \\
 & $\sigma = 60$, $c = 36$, irrelevant, stated $p = 0.5$ & 0.86 & $\leq$ 0.0 & $\leq$ $-$1.0 & $-$1.8 & 0.6 & 13 & $> 32$ & $> 0$ \\
\midrule
OLMo 2 32B & $\sigma = 60$, $c = 1$, relevant & 0.90 & 1.9 & 1.8 & 2.0 & 2.3 & 5.8 & $> 32$ & $> 0$ \\
 & $\sigma = 60$, $c = 1$, irrelevant & 0.89 & $\leq$ 0.0 & $-$0.5 & $-$0.2 & $-$0.2 & 1.7 & 25 & $> 0$ \\
 & $\sigma = 60$, $c = 4$, relevant & 0.91 & 5.8 & 3.4 & 3.4 & 4.5 & 9.3 & $> 32$ & $> 0$ \\
 & $\sigma = 60$, $c = 4$, irrelevant & 0.90 & $\leq$ 0.0 & $-$0.7 & $-$0.3 & $-$0.5 & 0.3 & 17 & $> 0$ \\
 & $\sigma = 60$, $c = 36$, relevant & 0.91 & 11 & 4.6 & 5.3 & 6.8 & 15 & $> 32$ & $> 0$ \\
 & $\sigma = 60$, $c = 36$, irrelevant & 0.90 & $\leq$ 0.0 & $\leq$ $-$1.0 & $-$0.6 & $-$0.5 & $-$0.2 & 15 & $> 0$ \\
 & $\sigma = 30$, $c = 1$, relevant & 0.92 & 1.8 & 1.9 & 1.7 & 1.7 & 4.4 & 24 & $> 0$ \\
 & $\sigma = 30$, $c = 4$, relevant & 0.92 & 4.8 & 3.3 & 3.1 & 4.0 & 9.5 & $> 32$ & $> 0$ \\
 & $\sigma = 30$, $c = 36$, relevant & 0.93 & 7.0 & 4.4 & 4.0 & 5.9 & 14 & $> 32$ & $> 0$ \\
 & $\sigma = 60$, $c = 4$, relevant, stated $p = 0.9$ & 0.96 & 6.0 & 4.6 & 3.8 & 4.5 & 5.8 & 23 & $> 0$ \\
 & $\sigma = 60$, $c = 4$, irrelevant, stated $p = 0.9$ & 0.96 & $\leq$ 0.0 & $-$0.9 & 0.0 & 0.0 & $-$0.4 & 4.1 & $> 0$ \\
 & $\sigma = 60$, $c = 36$, relevant, stated $p = 0.9$ & 0.96 & 12 & 5.9 & 6.0 & 6.3 & 8.9 & 31 & $> 0$ \\
 & $\sigma = 60$, $c = 36$, irrelevant, stated $p = 0.9$ & 0.96 & $\leq$ 0.0 & $\leq$ $-$1.0 & $-$0.3 & $-$0.1 & $-$0.8 & 0.3 & $> 0$ \\
 & $\sigma = 60$, $c = 4$, relevant, stated $p = 0.5$ & 0.93 & 5.3 & 4.2 & 3.5 & 4.0 & 5.4 & 27 & $> 0$ \\
 & $\sigma = 60$, $c = 4$, irrelevant, stated $p = 0.5$ & 0.92 & $\leq$ 0.0 & $-$0.5 & 0.1 & 0.1 & 0.0 & 11 & $> 0$ \\
 & $\sigma = 60$, $c = 36$, relevant, stated $p = 0.5$ & 0.93 & 7.8 & 5.4 & 4.9 & 5.6 & 7.9 & $> 32$ & $> 0$ \\
 & $\sigma = 60$, $c = 36$, irrelevant, stated $p = 0.5$ & 0.92 & $\leq$ 0.0 & $-$0.8 & $-$0.1 & 0.0 & $-$0.3 & 7.8 & $> 0$ \\
\bottomrule
\end{tabular}

\end{table}

\begin{table}[p]
\centering\scriptsize
\setlength{\tabcolsep}{3pt}
\caption{\textbf{ESS by description condition and demonstration count: written readout.} Results cover the three written-readout models in the $2{,}000$-task run without reasoning. Replies without a qualifying number are excluded. Columns and reference values follow Table~\ref{tab:llm-worth}.}
\label{tab:llm-worth-written}
\vspace{3pt}
\begin{tabular}{llrrrrrrrr}
\toprule
Predictor & Description & number & $n = 1$ & $n = 2$ & $n = 4$ & $n = 8$ & $n = 16$ & $n = 32$ & $n = 64$ \\
\midrule
Qwen3.5 9B (written) & $\sigma = 60$, $c = 1$, relevant & 1.00 & 0.2 & 0.3 & $-$0.9 & $-$2.8 & $-$9.3 & $-$24 & $-$51 \\
 & $\sigma = 60$, $c = 1$, irrelevant & 1.00 & $\leq$ 0.0 & $\leq$ $-$1.0 & $-$2.7 & $-$4.8 & $-$9.7 & $-$23 & $-$49 \\
 & $\sigma = 60$, $c = 4$, relevant & 1.00 & 3.9 & 2.7 & 1.3 & $-$0.7 & $-$6.2 & $-$18 & $-$44 \\
 & $\sigma = 60$, $c = 4$, irrelevant & 1.00 & $\leq$ 0.0 & $\leq$ $-$1.0 & $-$1.7 & $-$3.8 & $-$9.4 & $-$24 & $-$49 \\
 & $\sigma = 60$, $c = 36$, relevant & 1.00 & 6.5 & 4.0 & 2.8 & 6.9 & 17 & $> 32$ & $> 0$ \\
 & $\sigma = 60$, $c = 36$, irrelevant & 1.00 & $\leq$ 0.0 & $\leq$ $-$1.0 & $-$2.8 & $-$5.1 & $-$10 & $-$24 & $-$51 \\
 & $\sigma = 30$, $c = 1$, relevant & 1.00 & $\leq$ 0.0 & 0.2 & $-$0.7 & $-$2.5 & $-$8.7 & $-$22 & $-$49 \\
 & $\sigma = 30$, $c = 4$, relevant & 1.00 & 3.4 & 2.7 & 1.5 & $-$0.3 & $-$4.5 & $-$14 & $-$40 \\
 & $\sigma = 30$, $c = 36$, relevant & 1.00 & 4.7 & 4.3 & 3.0 & 9.2 & 14 & $> 32$ & $> 0$ \\
 & $\sigma = 60$, $c = 4$, relevant, stated $p = 0.9$ & 1.00 & 3.8 & $-$0.4 & $-$0.7 & $-$1.8 & $-$8.5 & $-$21 & $-$47 \\
 & $\sigma = 60$, $c = 4$, irrelevant, stated $p = 0.9$ & 1.00 & $\leq$ 0.0 & $\leq$ $-$1.0 & $\leq$ $-$3.0 & $-$4.9 & $-$10 & $-$24 & $-$50 \\
 & $\sigma = 60$, $c = 36$, relevant, stated $p = 0.9$ & 1.00 & 5.3 & 0.4 & 0.6 & $-$0.3 & $-$3.3 & $-$4.2 & $-$6.9 \\
 & $\sigma = 60$, $c = 36$, irrelevant, stated $p = 0.9$ & 1.00 & $\leq$ 0.0 & $\leq$ $-$1.0 & $\leq$ $-$3.0 & $-$5.2 & $-$10 & $-$25 & $-$53 \\
 & $\sigma = 60$, $c = 4$, relevant, stated $p = 0.5$ & 1.00 & 4.5 & $-$0.1 & $-$0.4 & $-$1.9 & $-$9.0 & $-$23 & $-$50 \\
 & $\sigma = 60$, $c = 4$, irrelevant, stated $p = 0.5$ & 1.00 & $\leq$ 0.0 & $\leq$ $-$1.0 & $\leq$ $-$3.0 & $-$6.2 & $-$12 & $-$26 & $-$54 \\
 & $\sigma = 60$, $c = 36$, relevant, stated $p = 0.5$ & 1.00 & 7.3 & 0.8 & 1.0 & $-$0.1 & $-$4.0 & $-$4.2 & $-$12 \\
 & $\sigma = 60$, $c = 36$, irrelevant, stated $p = 0.5$ & 1.00 & $\leq$ 0.0 & $\leq$ $-$1.0 & $\leq$ $-$3.0 & $-$7.0 & $-$13 & $-$27 & $-$56 \\
\midrule
Gemma 4 12B (written) & $\sigma = 60$, $c = 1$, relevant & 1.00 & 0.0 & 0.0 & $-$0.1 & $-$0.6 & $-$3.2 & $-$5.4 & $-$21 \\
 & $\sigma = 60$, $c = 1$, irrelevant & 1.00 & $\leq$ 0.0 & $-$0.4 & $-$1.3 & $-$1.3 & $-$4.4 & $-$6.8 & $-$24 \\
 & $\sigma = 60$, $c = 4$, relevant & 1.00 & 0.3 & 0.5 & 1.4 & 1.1 & $-$0.6 & $-$2.7 & $-$13 \\
 & $\sigma = 60$, $c = 4$, irrelevant & 1.00 & $\leq$ 0.0 & $-$0.4 & $-$1.1 & $-$1.2 & $-$4.1 & $-$6.9 & $-$23 \\
 & $\sigma = 60$, $c = 36$, relevant & 1.00 & 1.8 & 1.2 & 2.7 & 3.6 & 3.6 & 4.9 & $> 0$ \\
 & $\sigma = 60$, $c = 36$, irrelevant & 1.00 & $\leq$ 0.0 & $-$0.8 & $-$1.3 & $-$1.3 & $-$4.9 & $-$7.9 & $-$26 \\
 & $\sigma = 30$, $c = 1$, relevant & 1.00 & $\leq$ 0.0 & $-$0.2 & $-$0.1 & $-$0.5 & $-$3.2 & $-$7.9 & $-$27 \\
 & $\sigma = 30$, $c = 4$, relevant & 1.00 & 0.7 & 0.8 & 2.1 & 1.5 & $-$0.2 & $-$4.6 & $-$18 \\
 & $\sigma = 30$, $c = 36$, relevant & 1.00 & 4.0 & 1.7 & 3.4 & 3.7 & 2.4 & $-$1.8 & $-$8.9 \\
 & $\sigma = 60$, $c = 4$, relevant, stated $p = 0.9$ & 1.00 & 0.1 & 0.7 & 1.4 & 0.4 & $-$1.0 & $-$3.1 & $-$7.3 \\
 & $\sigma = 60$, $c = 4$, irrelevant, stated $p = 0.9$ & 1.00 & $\leq$ 0.0 & $-$0.2 & $-$0.4 & $-$0.5 & $-$2.2 & $-$5.4 & $-$13 \\
 & $\sigma = 60$, $c = 36$, relevant, stated $p = 0.9$ & 1.00 & 0.6 & 1.3 & 2.3 & 1.6 & 0.7 & $-$0.6 & $-$1.3 \\
 & $\sigma = 60$, $c = 36$, irrelevant, stated $p = 0.9$ & 1.00 & $\leq$ 0.0 & $-$0.2 & $-$0.5 & $-$0.5 & $-$2.5 & $-$5.8 & $-$15 \\
 & $\sigma = 60$, $c = 4$, relevant, stated $p = 0.5$ & 1.00 & 1.0 & 0.6 & 1.7 & 1.0 & $-$0.1 & $-$1.7 & $-$4.5 \\
 & $\sigma = 60$, $c = 4$, irrelevant, stated $p = 0.5$ & 1.00 & $\leq$ 0.0 & $-$0.3 & $-$0.4 & $-$0.4 & $-$1.3 & $-$4.1 & $-$9.6 \\
 & $\sigma = 60$, $c = 36$, relevant, stated $p = 0.5$ & 1.00 & 5.1 & 1.0 & 2.7 & 2.3 & 2.4 & 1.1 & $> 0$ \\
 & $\sigma = 60$, $c = 36$, irrelevant, stated $p = 0.5$ & 1.00 & $\leq$ 0.0 & $-$0.3 & $-$0.6 & $-$0.4 & $-$1.9 & $-$4.8 & $-$13 \\
\midrule
Gemma 4 31B (written) & $\sigma = 60$, $c = 1$, relevant & 1.00 & 0.1 & 0.0 & $-$0.1 & $-$0.5 & $-$1.4 & $-$5.1 & $-$13 \\
 & $\sigma = 60$, $c = 1$, irrelevant & 1.00 & $\leq$ 0.0 & $-$0.1 & $-$0.1 & $-$0.5 & $-$1.0 & $-$4.5 & $-$11 \\
 & $\sigma = 60$, $c = 4$, relevant & 1.00 & 0.8 & 0.1 & 0.2 & $-$0.5 & $-$1.0 & $-$4.3 & $-$12 \\
 & $\sigma = 60$, $c = 4$, irrelevant & 1.00 & $\leq$ 0.0 & $-$0.2 & $-$0.2 & $-$0.4 & $-$1.0 & $-$4.4 & $-$10 \\
 & $\sigma = 60$, $c = 36$, relevant & 1.00 & 2.6 & 0.6 & 1.4 & 0.9 & 2.9 & 3.4 & $-$2.9 \\
 & $\sigma = 60$, $c = 36$, irrelevant & 1.00 & $\leq$ 0.0 & $\leq$ $-$1.0 & $-$0.8 & $-$0.6 & $-$1.1 & $-$2.7 & $-$4.8 \\
 & $\sigma = 30$, $c = 1$, relevant & 1.00 & $\leq$ 0.0 & 0.0 & $-$0.1 & $-$0.7 & $-$1.2 & $-$3.0 & $-$8.1 \\
 & $\sigma = 30$, $c = 4$, relevant & 1.00 & 0.5 & 0.0 & 0.1 & $-$0.5 & $-$1.3 & $-$3.4 & $-$8.3 \\
 & $\sigma = 30$, $c = 36$, relevant & 1.00 & 4.3 & 1.5 & 2.1 & 2.3 & 3.3 & 3.3 & $-$0.4 \\
 & $\sigma = 60$, $c = 4$, relevant, stated $p = 0.9$ & 1.00 & 0.1 & 0.0 & 0.0 & $-$0.7 & $-$1.9 & $-$5.0 & $-$15 \\
 & $\sigma = 60$, $c = 4$, irrelevant, stated $p = 0.9$ & 1.00 & $\leq$ 0.0 & 0.0 & $-$0.1 & $-$0.4 & $-$1.2 & $-$4.1 & $-$9.5 \\
 & $\sigma = 60$, $c = 36$, relevant, stated $p = 0.9$ & 1.00 & 1.0 & 0.0 & 0.6 & $-$0.3 & $-$1.4 & $-$4.8 & $-$12 \\
 & $\sigma = 60$, $c = 36$, irrelevant, stated $p = 0.9$ & 1.00 & $\leq$ 0.0 & 0.0 & $-$0.2 & $-$0.4 & $-$0.9 & $-$3.0 & $-$6.2 \\
 & $\sigma = 60$, $c = 4$, relevant, stated $p = 0.5$ & 1.00 & 0.2 & 0.0 & 0.0 & $-$0.7 & $-$1.7 & $-$4.6 & $-$17 \\
 & $\sigma = 60$, $c = 4$, irrelevant, stated $p = 0.5$ & 1.00 & $\leq$ 0.0 & 0.0 & $-$0.2 & $-$0.4 & $-$1.2 & $-$3.6 & $-$12 \\
 & $\sigma = 60$, $c = 36$, relevant, stated $p = 0.5$ & 1.00 & 1.0 & 0.0 & 0.5 & $-$0.3 & $-$1.5 & $-$4.5 & $-$15 \\
 & $\sigma = 60$, $c = 36$, irrelevant, stated $p = 0.5$ & 1.00 & $\leq$ 0.0 & 0.0 & $-$0.1 & $-$0.4 & $-$1.0 & $-$2.9 & $-$8.5 \\
\bottomrule
\end{tabular}

\end{table}

\begin{table}[p]
\centering\scriptsize
\setlength{\tabcolsep}{3pt}
\caption{\textbf{Description weights by demonstration count: distribution readout.} Results cover the Bayesian reference and the three distribution-readout models in the $2{,}000$-task run without reasoning. ``Number'' is defined as in Table~\ref{tab:llm-worth}; cells are italicized when the numeric response rate is below three quarters at the corresponding $n$.}
\label{tab:llm-full}
\vspace{3pt}
\begin{tabular}{llrrrrrrrrr}
\toprule
Predictor & Description & number & $n = 0$ & $n = 1$ & $n = 2$ & $n = 4$ & $n = 8$ & $n = 16$ & $n = 32$ & $n = 64$ \\
\midrule
Reference & $\sigma = 60$, $c = 1$, relevant &  & 1.00 & 0.50 & 0.33 & 0.20 & 0.11 & 0.06 & 0.03 & 0.02 \\
 & $\sigma = 60$, $c = 1$, irrelevant &  & 1.00 & 0.50 & 0.33 & 0.20 & 0.11 & 0.06 & 0.03 & 0.02 \\
 & $\sigma = 60$, $c = 4$, relevant &  & 1.00 & 0.80 & 0.67 & 0.50 & 0.33 & 0.20 & 0.11 & 0.06 \\
 & $\sigma = 60$, $c = 4$, irrelevant &  & 1.00 & 0.80 & 0.67 & 0.50 & 0.33 & 0.20 & 0.11 & 0.06 \\
 & $\sigma = 60$, $c = 36$, relevant &  & 1.00 & 0.97 & 0.95 & 0.90 & 0.82 & 0.69 & 0.53 & 0.36 \\
 & $\sigma = 60$, $c = 36$, irrelevant &  & 1.00 & 0.97 & 0.95 & 0.90 & 0.82 & 0.69 & 0.53 & 0.36 \\
 & $\sigma = 30$, $c = 1$, relevant &  & 1.00 & 0.50 & 0.33 & 0.20 & 0.11 & 0.06 & 0.03 & 0.02 \\
 & $\sigma = 30$, $c = 4$, relevant &  & 1.00 & 0.80 & 0.67 & 0.50 & 0.33 & 0.20 & 0.11 & 0.06 \\
 & $\sigma = 30$, $c = 36$, relevant &  & 1.00 & 0.97 & 0.95 & 0.90 & 0.82 & 0.69 & 0.53 & 0.36 \\
 & $\sigma = 60$, $c = 4$, relevant, stated $p = 0.9$ &  & 0.90 & 0.71 & 0.59 & 0.45 & 0.30 & 0.19 & 0.10 & 0.05 \\
 & $\sigma = 60$, $c = 4$, irrelevant, stated $p = 0.9$ &  & 0.90 & 0.27 & 0.14 & 0.08 & 0.04 & 0.02 & 0.01 & 0.00 \\
 & $\sigma = 60$, $c = 36$, relevant, stated $p = 0.9$ &  & 0.90 & 0.84 & 0.84 & 0.82 & 0.77 & 0.65 & 0.51 & 0.34 \\
 & $\sigma = 60$, $c = 36$, irrelevant, stated $p = 0.9$ &  & 0.90 & 0.26 & 0.14 & 0.07 & 0.03 & 0.01 & 0.00 & 0.00 \\
 & $\sigma = 60$, $c = 4$, relevant, stated $p = 0.5$ &  & 0.50 & 0.46 & 0.39 & 0.32 & 0.22 & 0.14 & 0.08 & 0.04 \\
 & $\sigma = 60$, $c = 4$, irrelevant, stated $p = 0.5$ &  & 0.50 & 0.12 & 0.07 & 0.03 & 0.02 & 0.01 & 0.00 & 0.00 \\
 & $\sigma = 60$, $c = 36$, relevant, stated $p = 0.5$ &  & 0.50 & 0.55 & 0.58 & 0.60 & 0.59 & 0.53 & 0.42 & 0.29 \\
 & $\sigma = 60$, $c = 36$, irrelevant, stated $p = 0.5$ &  & 0.50 & 0.12 & 0.06 & 0.03 & 0.01 & 0.00 & 0.00 & 0.00 \\
\midrule
gpt-oss-20b & $\sigma = 60$, $c = 1$, relevant & 0.98 & 0.94 & 0.53 & 0.29 & 0.23 & 0.16 & 0.12 & 0.12 & 0.23 \\
 & $\sigma = 60$, $c = 1$, irrelevant & 0.98 & 0.93 & 0.59 & 0.25 & 0.18 & 0.09 & 0.05 & 0.04 & 0.10 \\
 & $\sigma = 60$, $c = 4$, relevant & 0.98 & 0.92 & 0.48 & 0.32 & 0.26 & 0.21 & 0.19 & 0.18 & 0.31 \\
 & $\sigma = 60$, $c = 4$, irrelevant & 0.97 & 0.92 & 0.57 & 0.25 & 0.16 & 0.08 & 0.04 & 0.04 & 0.09 \\
 & $\sigma = 60$, $c = 36$, relevant & 0.98 & 0.94 & 0.49 & 0.35 & 0.30 & 0.25 & 0.23 & 0.22 & 0.35 \\
 & $\sigma = 60$, $c = 36$, irrelevant & 0.98 & 0.95 & 0.60 & 0.25 & 0.15 & 0.07 & 0.04 & 0.04 & 0.09 \\
 & $\sigma = 30$, $c = 1$, relevant & 0.99 & 0.98 & 0.43 & 0.23 & 0.17 & 0.12 & 0.09 & 0.11 & 0.21 \\
 & $\sigma = 30$, $c = 4$, relevant & 0.98 & 0.98 & 0.38 & 0.28 & 0.23 & 0.19 & 0.18 & 0.19 & 0.31 \\
 & $\sigma = 30$, $c = 36$, relevant & 0.99 & 0.99 & 0.37 & 0.32 & 0.27 & 0.23 & 0.22 & 0.24 & 0.37 \\
 & $\sigma = 60$, $c = 4$, relevant, stated $p = 0.9$ & 0.96 & 0.91 & 0.45 & 0.33 & 0.15 & 0.13 & 0.12 & 0.13 & 0.22 \\
 & $\sigma = 60$, $c = 4$, irrelevant, stated $p = 0.9$ & 0.96 & 0.91 & 0.55 & 0.34 & 0.11 & 0.05 & 0.02 & 0.02 & 0.04 \\
 & $\sigma = 60$, $c = 36$, relevant, stated $p = 0.9$ & 0.97 & 0.92 & 0.44 & 0.34 & 0.17 & 0.15 & 0.15 & 0.16 & 0.25 \\
 & $\sigma = 60$, $c = 36$, irrelevant, stated $p = 0.9$ & 0.96 & 0.92 & 0.53 & 0.34 & 0.10 & 0.04 & 0.02 & 0.02 & 0.04 \\
 & $\sigma = 60$, $c = 4$, relevant, stated $p = 0.5$ & 0.95 & 0.79 & 0.46 & 0.32 & 0.16 & 0.12 & 0.12 & 0.12 & 0.20 \\
 & $\sigma = 60$, $c = 4$, irrelevant, stated $p = 0.5$ & 0.94 & 0.79 & 0.53 & 0.32 & 0.11 & 0.04 & 0.02 & 0.01 & 0.04 \\
 & $\sigma = 60$, $c = 36$, relevant, stated $p = 0.5$ & 0.95 & 0.83 & 0.46 & 0.32 & 0.17 & 0.14 & 0.15 & 0.15 & 0.23 \\
 & $\sigma = 60$, $c = 36$, irrelevant, stated $p = 0.5$ & 0.95 & 0.83 & 0.53 & 0.31 & 0.11 & 0.04 & 0.02 & 0.02 & 0.04 \\
\midrule
Phi-4 & $\sigma = 60$, $c = 1$, relevant & 0.87 & \textit{0.97} & 0.82 & 0.62 & 0.43 & 0.28 & 0.18 & 0.13 & 0.12 \\
 & $\sigma = 60$, $c = 1$, irrelevant & 0.86 & \textit{0.97} & 0.83 & 0.64 & 0.42 & 0.22 & 0.11 & 0.06 & 0.05 \\
 & $\sigma = 60$, $c = 4$, relevant & 0.85 & \textit{0.98} & 0.76 & 0.55 & 0.37 & 0.26 & 0.19 & 0.15 & 0.16 \\
 & $\sigma = 60$, $c = 4$, irrelevant & 0.84 & \textit{0.98} & 0.79 & 0.56 & 0.36 & 0.18 & 0.09 & 0.06 & 0.05 \\
 & $\sigma = 60$, $c = 36$, relevant & 0.86 & \textit{0.99} & 0.74 & 0.53 & 0.38 & 0.29 & 0.23 & 0.20 & 0.20 \\
 & $\sigma = 60$, $c = 36$, irrelevant & 0.84 & \textit{0.99} & 0.78 & 0.53 & 0.33 & 0.16 & 0.09 & 0.06 & 0.04 \\
 & $\sigma = 30$, $c = 1$, relevant & 0.87 & \textit{0.99} & 0.77 & 0.58 & 0.38 & 0.22 & 0.13 & 0.09 & 0.09 \\
 & $\sigma = 30$, $c = 4$, relevant & 0.86 & \textit{1.00} & 0.69 & 0.51 & 0.34 & 0.23 & 0.16 & 0.13 & 0.13 \\
 & $\sigma = 30$, $c = 36$, relevant & 0.88 & \textit{1.00} & 0.65 & 0.48 & 0.32 & 0.23 & 0.18 & 0.17 & 0.16 \\
 & $\sigma = 60$, $c = 4$, relevant, stated $p = 0.9$ & 0.87 & \textit{0.94} & 0.62 & 0.46 & 0.30 & 0.21 & 0.15 & 0.13 & 0.11 \\
 & $\sigma = 60$, $c = 4$, irrelevant, stated $p = 0.9$ & 0.85 & \textit{0.95} & 0.62 & 0.44 & 0.26 & 0.13 & 0.06 & 0.04 & 0.03 \\
 & $\sigma = 60$, $c = 36$, relevant, stated $p = 0.9$ & 0.87 & \textit{0.95} & 0.61 & 0.46 & 0.33 & 0.25 & 0.20 & 0.18 & 0.16 \\
 & $\sigma = 60$, $c = 36$, irrelevant, stated $p = 0.9$ & 0.86 & \textit{0.96} & 0.61 & 0.43 & 0.26 & 0.13 & 0.06 & 0.04 & 0.03 \\
 & $\sigma = 60$, $c = 4$, relevant, stated $p = 0.5$ & 0.87 & \textit{0.92} & 0.57 & 0.43 & 0.29 & 0.20 & 0.15 & 0.13 & 0.11 \\
 & $\sigma = 60$, $c = 4$, irrelevant, stated $p = 0.5$ & 0.85 & \textit{0.92} & 0.57 & 0.42 & 0.24 & 0.12 & 0.06 & 0.04 & 0.03 \\
 & $\sigma = 60$, $c = 36$, relevant, stated $p = 0.5$ & 0.88 & \textit{0.93} & 0.56 & 0.44 & 0.33 & 0.24 & 0.21 & 0.19 & 0.17 \\
 & $\sigma = 60$, $c = 36$, irrelevant, stated $p = 0.5$ & 0.86 & \textit{0.94} & 0.57 & 0.41 & 0.24 & 0.12 & 0.06 & 0.04 & 0.03 \\
\midrule
OLMo 2 32B & $\sigma = 60$, $c = 1$, relevant & 0.90 & \textit{0.99} & 0.72 & 0.33 & 0.25 & 0.16 & 0.09 & 0.05 & 0.05 \\
 & $\sigma = 60$, $c = 1$, irrelevant & 0.89 & \textit{0.99} & 0.77 & 0.29 & 0.16 & 0.09 & 0.05 & 0.02 & 0.02 \\
 & $\sigma = 60$, $c = 4$, relevant & 0.91 & \textit{1.00} & 0.70 & 0.37 & 0.30 & 0.21 & 0.14 & 0.08 & 0.07 \\
 & $\sigma = 60$, $c = 4$, irrelevant & 0.90 & \textit{1.00} & 0.75 & 0.29 & 0.15 & 0.08 & 0.04 & 0.02 & 0.01 \\
 & $\sigma = 60$, $c = 36$, relevant & 0.91 & \textit{1.00} & 0.70 & 0.40 & 0.34 & 0.26 & 0.19 & 0.12 & 0.10 \\
 & $\sigma = 60$, $c = 36$, irrelevant & 0.90 & \textit{1.00} & 0.76 & 0.29 & 0.17 & 0.08 & 0.04 & 0.02 & 0.02 \\
 & $\sigma = 30$, $c = 1$, relevant & 0.92 & \textit{0.99} & 0.69 & 0.32 & 0.23 & 0.14 & 0.09 & 0.04 & 0.04 \\
 & $\sigma = 30$, $c = 4$, relevant & 0.92 & \textit{1.00} & 0.65 & 0.34 & 0.27 & 0.19 & 0.13 & 0.06 & 0.06 \\
 & $\sigma = 30$, $c = 36$, relevant & 0.93 & \textit{1.00} & 0.64 & 0.36 & 0.32 & 0.23 & 0.16 & 0.08 & 0.09 \\
 & $\sigma = 60$, $c = 4$, relevant, stated $p = 0.9$ & 0.96 & 1.01 & 0.72 & 0.42 & 0.29 & 0.18 & 0.11 & 0.09 & 0.05 \\
 & $\sigma = 60$, $c = 4$, irrelevant, stated $p = 0.9$ & 0.96 & 1.01 & 0.81 & 0.32 & 0.14 & 0.06 & 0.03 & 0.02 & 0.01 \\
 & $\sigma = 60$, $c = 36$, relevant, stated $p = 0.9$ & 0.96 & 1.01 & 0.71 & 0.44 & 0.33 & 0.23 & 0.15 & 0.13 & 0.07 \\
 & $\sigma = 60$, $c = 36$, irrelevant, stated $p = 0.9$ & 0.96 & 1.01 & 0.82 & 0.33 & 0.15 & 0.07 & 0.03 & 0.02 & 0.01 \\
 & $\sigma = 60$, $c = 4$, relevant, stated $p = 0.5$ & 0.93 & 1.02 & 0.65 & 0.39 & 0.27 & 0.16 & 0.09 & 0.07 & 0.04 \\
 & $\sigma = 60$, $c = 4$, irrelevant, stated $p = 0.5$ & 0.92 & 1.01 & 0.71 & 0.29 & 0.13 & 0.06 & 0.02 & 0.01 & 0.01 \\
 & $\sigma = 60$, $c = 36$, relevant, stated $p = 0.5$ & 0.93 & 1.02 & 0.64 & 0.42 & 0.31 & 0.21 & 0.13 & 0.11 & 0.06 \\
 & $\sigma = 60$, $c = 36$, irrelevant, stated $p = 0.5$ & 0.92 & 1.02 & 0.72 & 0.30 & 0.14 & 0.06 & 0.03 & 0.02 & 0.01 \\
\bottomrule
\end{tabular}

\end{table}
\clearpage
\section{Additional Related Work}\label{app:related}

Concurrent work studies how reasoning and code execution affect language models' approximation of Bayesian posteriors \citep{filbry2026}; we vary only reasoning, with code execution disabled.

\paragraph{Task descriptions and demonstrations.}
\citet{weller2020} formalize learning from task descriptions by distinguishing the latent task, its natural-language description, and individual input--output examples. We use \emph{task description} for information about the task supplied in the prompt, including a numeric description token in the meta-trained setting and probabilistic statements or notes about a sorting rule in the language-model experiments. A \emph{demonstration} is an observed input--output example; in scalar mean estimation, it is an observation at the fixed input $x=1$. Summaries distilled from demonstrations can serve as descriptions \citep{honda2025}; our controlled experiments instead specify the joint distribution of task, description, and demonstrations.

\paragraph{Prior information and unreliable descriptions.}
Specificity and reliability correspond to the two ways in which a description's prior can fail to help a prediction: its variance can be large, or its mean can be independent of the task. The first is low specificity, and the second is irrelevance, which occurs with probability $1-p$. The two-component construction has the form of the robust mixture priors of Bayesian clinical trials, which combine informative and baseline distributions and update their component weights after observing data \citep{schmidli2014}. Our mixture is specified directly, rather than derived from a historical-data meta-analysis. Reliability $p$ is the prior probability of the relevant component; $\pi_n$ is its posterior probability. Averaging component predictions using $\pi_n$ is Bayesian model averaging \citep{hoeting1999}. In a robust mixture prior, the prior weight is fixed but the posterior weights still update; fixed-weight averaging, our experimental comparison, omits this update. Our use of \emph{reliability} follows Bayesian models of testimony, in which a source is reliable with some probability and the report of an unreliable source is independent of the truth \citep{bovens2003}. It differs from the reliability of a cue in models of cue combination, which is the cue's inverse variance \citep{ernst2002,ernst2006} and corresponds to our specificity.

In our model, an irrelevant description is independent of the task and may happen to agree with it. Irrelevance is therefore a property of how the description was generated, whereas prior--data conflict concerns disagreement between prior information and observed data.

\paragraph{Specificity and description weights.}
Specificity takes a concrete form in each setting. Section~\ref{sec:setting} defines it from the uncertainty that a relevant description leaves about the task, measured by the smallest expected loss of a prediction made from the description's prior alone, which \citet{grunwald2004} call the generalized entropy of that prior under the loss. Under squared error this is the prior's variance, which in the Gaussian model is $\tau^2$ in each coordinate that the description constrains, so specificity is the precision $c=\tau^{-2}$; a partial description constrains $k$ of the $d$ coordinates. A message-sorting note leaves a uniform prior over the rules compatible with it, and the fewer rules it leaves, the more specific it is: $20$ for a broad note and $2$ for a specific one. The logarithm of that number is the prior's entropy, the corresponding measure of uncertainty under log loss. These quantities do not share a scale across tasks; ESS expresses each in demonstrations. Specificity differs from a description's informativeness. A prior ESS quantifies how informative a prior is \citep{morita2008}, and under the mixture prior this depends on reliability as well as specificity: the zero-shot ESS of a description that may be irrelevant stays below a finite ceiling, however specific the description is (Section~\ref{sec:reliability}). Our description weights follow Gaussian mean estimation, in which the posterior combines the prior mean and the sample mean with weights proportional to their precisions \citep{neuenschwander2020}. Fitted and per-reply weights summarize model predictions; their normalized values can be compared with posterior component probabilities but need not be probabilities.

\paragraph{Effective sample size and predictive comparisons.}
Prior ESS has several definitions, including matching local information \citep{morita2008}, matching expected posterior loss \citep{reimherr2021}, and quantifying sample savings at a target estimation accuracy \citep{wiesenfarth2020}. Matching information requires an explicit prior. A network or a language model has prior knowledge from training, but only implicitly, in its parameters; its prediction error, by contrast, can be measured with and without the description. Our definition therefore follows the error-matching approach, but it evaluates population prediction error, allows a non-Bayesian predictor, and depends on the demonstration count. A predictor without a description still has prior knowledge, from the baseline prior or from training, and ESS measures what the description adds to it, so we measure ESS against each predictor's own baseline curve. For a predictor whose baseline curve falls slowly, the same error reduction is worth more demonstrations, so a large ESS need not imply low absolute error (Table~\ref{tab:error}). A negative ESS means that the description raises error. The word \emph{predictive} specifies what is matched; it does not assert the additive predictive-consistency property studied by \citet{neuenschwander2020}.

Two predictive comparisons are especially close. \citet{lescao2021} average horizontal data advantages over an accuracy range in prompt-based fine-tuning; we evaluate a pointwise difference in demonstration counts. \citet{selby2025} compare priors elicited from language models with data-based predictors by the observations needed to match predictive loss; we study descriptions given in the prompt and how their benefit changes after demonstrations. Other work compares instructions or prompts with examples. Beyond ICL, an instruction can replace up to $100$ execution examples in the fine-tuning settings studied by \citet{cook2026}, and \citet{wenliang2026} compare optimized prompts with demonstrations in terms of their length and effect.

\paragraph{Demonstrations, text, and probabilistic prediction.}
Other work examines semantic priors and flipped labels \citep{wei2023}, weak dependence on demonstration labels \citep{min2022}, Bayesian scaling laws \citep{arora2025}, and Bayesian reference curves for in-context function learning \citep{akata2026}. Text supplied alongside numbers can change numerical predictions \citep{requeima2024}. Language-model ICL can violate the martingale property that Bayesian prediction implies \citep{falck2024}, and \citet{tong2026} study how sensitivity to the prompt template varies with the number of demonstrations. Related work studies how language-model outputs compare with Bayesian inference and how training can improve probabilistic reasoning \citep{qiu2026}; \citet{he2025} examine Bayesian consistency during iterative reasoning on open-ended questions. Our numerical experiments state uncertainty explicitly, allowing comparison with a specified Bayesian reference as precision and relevance probability vary. The message-sorting experiment extends this comparison to constraints stated in words on a finite family of classification rules.

\paragraph{Scope of the results.}
The $d+1$ term of Proposition~\ref{prop:reliable} arises from random regression inputs; it is absent when every input is the same, as in our mean-estimation task (Appendix~\ref{app:ess-est}). For language models, updating the relevance probability from a stated reliability carries over to a description that states a rule in words (Appendix~\ref{app:semantic}). Agreement with the reference does not carry over to informal wording: when the prompt gives the prior mean ``give or take'' an amount, most of Haiku~5.5's replies with reasoning repeat the prior mean (Appendix~\ref{app:wording-reasoning}).
\end{document}